# DrugTargetWorld: A Synthetic Biobank for Training and Benchmarking AI Scientists

**Samuel Margolis[1,2], Paul Schmiedmayer[3], Alan Huang[1,2], Ethan Chen[4], Ishan Bhattacharjee[1], Atman Shah[4], Ben Viggiano[1,2], Fang Cao[1,2], Shriya Reddy[1,2], Roger Xia[1,2], Jack O'Sullivan[1,2], Daniel Katz[2,3], Matthew Wheeler[1,2], Euan Ashley[1,2], Bruna Gomes[†1,2]**

[1]Department of Biomedical Data Science, Stanford University, Stanford, CA 94305, USA
[2]Department of Medicine, Stanford University, Stanford, CA 94305, USA
[3]Division of Computational Medicine, Department of Medicine, Stanford University, Stanford, CA 94305, USA
[4]Brown University, Providence, RI 02912, USA
†Corresponding author: Bruna Gomes.

**Abstract**

**Drug target discovery requires distinguishing molecules that causally drive disease from the many that are merely associated with it, and on determining the direction of modulation expected to improve disease. Artificial intelligence (AI) agents capable of writing and executing code may increasingly automate portions of this workflow; however, training and evaluating such agents to perform target discovery end-to-end requires access to known ground truth targets. Real world biobanks cannot provide such ground truth, as causal relationships remain incompletely characterized and nominated targets require experimental validation. Furthermore, access controls on participant-level data impede large-scale training. DrugTargetWorld addresses these challenges by procedurally generating simulated biobanks, or 'worlds,' each containing genotypes, proteins, health records, and outcomes for 54,000 participants, including 10,800 with synthetic magnetic resonance imaging (MRI) data. The present study instantiates this framework in cardiovascular disease while the underlying world generation framework is designed to support other disease domains. Each world is governed by a concealed causal model that specifies the ground truth, including causal driver proteins for a disease, and non-causal proteins that appear causal due to biases such as confounding. Agents are tasked with constructing a disease phenotype from raw images or other released data, identifying which proteins causally drive disease and inferring the beneficial direction of modulation for each causal driver with the option to conduct virtual 'wet lab' experiments. Across 540 episodes, Opus 5 and GPT-5.6 Sol led nine agents on a 100 point composite score spanning target identification, causal confidence, intervention direction, bias identification and disease measurement, scoring 39.98 and 35.38, respectively. Both recovered 64% of causal drivers on average, but no agent reliably distinguished misleading non-causal proteins; Opus 5's advantage over GPT-5.6 Sol arose mainly from how well it measured disease from the raw data (7.2 versus 3.2 of 10 points). These findings suggest that leading agents can perform most individual analyses required in biobank studies but do not yet consistently make the integrative judgments needed to connect these analyses, namely how to measure disease, distinguish causal drivers from non-causal proteins, and determine when evidence is sufficient to support a claim. By making the causal structure of every world known but hidden from the agent, DrugTargetWorld turns end to end drug target discovery into a scalable, training problem in which research strategies can be evaluated against causal truth and improved through verifiable reward.**

## 1. Introduction

Identifying drug targets remains a difficult part of drug development (Pun et al., 2026). Human genetic evidence can make this step more reliable, and population biobanks now link genetic data with health records and outcomes for hundreds of thousands of individuals (Bycroft et al., 2018). For example, family and population studies identified PCSK9 as a target for lowering low-density lipoprotein cholesterol, and variants that reduce its activity anticipated the cardiovascular benefit of its inhibition before outcome trials were reported (Abifadel et al., 2003; Cohen et al., 2006; Ference et al., 2016). More broadly, drug mechanisms with genetic support are 2.6 times as likely as those without to progress from phase I to launch (Minikel et al., 2024), and plasma proteomics now extends such evidence to thousands of circulating proteins (Sun et al., 2023).

Within biobank-based target discovery, recent studies have used Mendelian randomization, colocalization, and cross-phenotype analysis to move from protein-disease associations to genetically supported drug targets (Henry et al., 2022; Lind et al., 2024; Rasooly et al., 2023, 2025; Reddy et al., 2025; Zheng et al., 2020). These analyses test whether a circulating protein causally affects a disease phenotype, which must first be defined. Which measurements to extract from raw imaging, health records, and outcomes, and how to combine them into a participant-level phenotype, are scientific decisions that can change what genetic analyses discover (An et al., 2023; Gomes et al., 2024). We call this step phenotype construction, meaning the selection and combination of available measurements into a participant-level representation of disease; it is broader than assigning a disease label from health records. Like the causal analyses that follow, it still relies heavily on human investigators.

In parallel, artificial intelligence (AI) systems built on language models and trained with reinforcement learning have advanced rapidly in mathematical reasoning, formal proof, and software engineering (Guo et al., 2025; Hubert et al., 2026; Jimenez et al., 2024). AI agents (hereafter, agents), which are language-model systems that write and run code to carry out multi-step tasks, have begun to contribute to mathematical research, including a proof that the forced three-dimensional Navier–Stokes equations can develop a singularity in finite time, formalized in the Lean proof assistant but not yet peer-reviewed (OpenAI, 2026). Many of these advances rely on a verifiable reward, an automatic check of whether an output is correct, which can both evaluate and train agents (Wen et al., 2025), as training environments for software-engineering agents illustrate (Pan et al., 2025).

Agents are also increasingly applied to science, where recent benchmarks and agent systems address data analysis, genetic analysis, and the reproduction of published studies (Chen, 2025; Koch, 2026; Li & Ho, 2026; Mitchener et al., 2025; Xu et al., 2025). Kosmos, for example, runs extended cycles of data analysis and literature search (Mitchener et al., 2025b). In drug discovery, the Virtual Biotech organizes agents into the divisions of a drug-development company, and more than 37,000 of its agents annotated the outcomes of 55,984 clinical trials (Zhang et al., 2026). Biomni, an agent that executes diverse biomedical research tasks (Huang et al., 2026), now underpins a commercial tool for target prioritization (Phylo, 2026), and Aviary has shown that agents improve through repeated training in a scientific environment (Narayanan et al., 2024). However, these benchmarks typically specify the research question, and neither they nor such systems can be scored against causal ground truth on real data. Because many of these systems also draw on published literature and curated databases, it is difficult to determine whether their conclusions come from the data or from existing knowledge. New targets, by contrast, require primary data in which the answer is not yet published, and single-cell atlases, perturbation screens, and population biobanks each supply only part of that evidence. We focus on population biobanks, which link genetic variation to clinical outcomes in the same individuals.

In real biobanks, however, drug target discovery lacks such a verifiable reward. Open Targets and Perturb-seq screens provide substantial evidence for many proteins (Ochoa et al., 2023; Replogle et al., 2022), but for most proteins they leave open whether modulation affects disease in patients. The causal links between proteins and a disease phenotype therefore remain only partially known, and the link between a measurable phenotype and the disease mechanism is often even less certain. An agent's nomination of a causal driver and its intervention direction thus requires experimental validation, since even compelling observational evidence can reflect confounding, reverse causation, selection bias, pleiotropy, or measurement artifacts (Leek et al., 2010; Munafò et al., 2018; Sanderson et al., 2022). Access is a further obstacle, because participant-level biobank data are released under controlled-access agreements and analyzed within approved computing environments, which hinders the repeated, large-scale interaction that training an agent requires. Causal discovery benchmarks do supply known ground truth (Chen et al., 2026; Leban & Sun, 2026; J. Yang et al., 2026), but their variables are predefined, so agents neither construct a phenotype from raw data nor search among thousands of candidate proteins.

To address these gaps, we developed DrugTargetWorld, an environment that supplies this reward by procedurally generating population biobanks with known causal structure. Each simulated biobank, or world, comes from a hidden causal model and holds genotypes, circulating proteins, health records, and outcomes for 54,000 participants, with raw cardiac magnetic resonance imaging (MRI) for 10,800 of them (Figures 1a and 2). The model differs between worlds and specifies a cardiac disease archetype, zero to six causal driver proteins that alter latent disease severity, non-causal proteins that appear causal, and a harmful surrogate that improves imaging measures while shortening survival. The current evaluation focuses on cardiovascular disease archetypes as initial test cases; the same world-generation framework can be extended to other disease domains. Because latent severity is never observed, the agent must construct a disease phenotype as well as identify causal drivers and their intervention directions. Anonymous protein labels require it to derive every causal claim from the released data without drawing on published knowledge. Our primary aim was to determine how well current agents perform this work when they choose their own research strategy, including which of an unknown number of candidates to pursue and, given a budget, which virtual follow-up experiments to perform and how to use the returned results.

We evaluated nine agents in 540 episodes across 20 worlds and three experimental budgets. The leading agents, Opus 5 and GPT-5.6 Sol, built disease phenotypes from raw data and applied genetic instruments, yet no agent averaged more than 40 of 100 points. Their performance was limited mainly by the judgments that connect these analyses, namely how to measure the disease, which evidence separates a causal driver from a non-causal protein, and when a claim is sufficiently supported. Because new worlds can be generated as needed, DrugTargetWorld can also supply verifiable rewards for training agents across the entire workflow, and strategies learned in this way can then be tested in real cohorts.

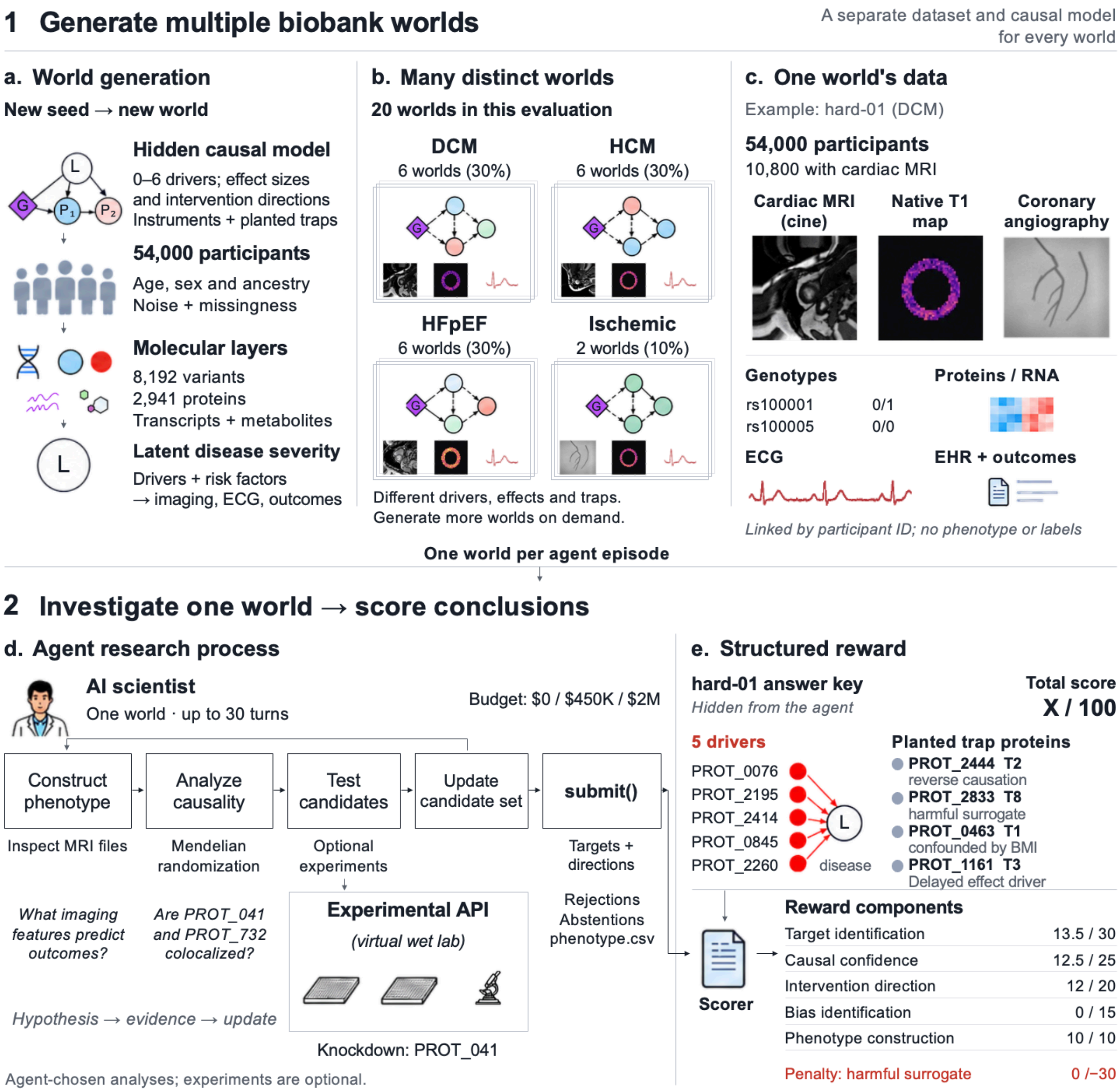

**Figure 1 | DrugTargetWorld generates multiple causal biobank worlds and evaluates complete agent research trajectories against hidden ground truth. a,** Each world is generated independently from a hidden causal model specifying the disease archetype, causal protein drivers, effect sizes and directions, genetic instruments, planted causal traps and participant-level variation. The model generates a synthetic population of 54,000 participants with linked molecular measurements, latent disease severity and multimodal clinical observations. **b,** Repeating this procedure with different seeds produces a collection of distinct biobank worlds rather than a single dataset. The version 1 evaluation panel contains 20 worlds: six dilated cardiomyopathy (DCM), six hypertrophic cardiomyopathy (HCM), six heart failure with preserved ejection fraction (HFpEF) and two ischemic worlds, each with its own causal structure and hidden answer key. **c,** Each world contains linked participant-level genotypes, proteins, transcripts, metabolites, cardiac MRI, ECG, coronary imaging, electronic health records and outcomes. No disease phenotype, causal labels or prescribed analysis pipeline are released. **d,** In each episode, an agent receives the released data from one world and has at most 30 turns to construct a phenotype, choose and execute

analyses, optionally purchase virtual experiments subject to an experimental budget, update its hypotheses and submit final claims. **e,** The submission is compared with that world's hidden causal ground truth to produce a verifiable reward based on target identification, causal confidence, intervention direction, bias identification and phenotype construction, with an additional penalty for advancing the harmful surrogate as beneficial.

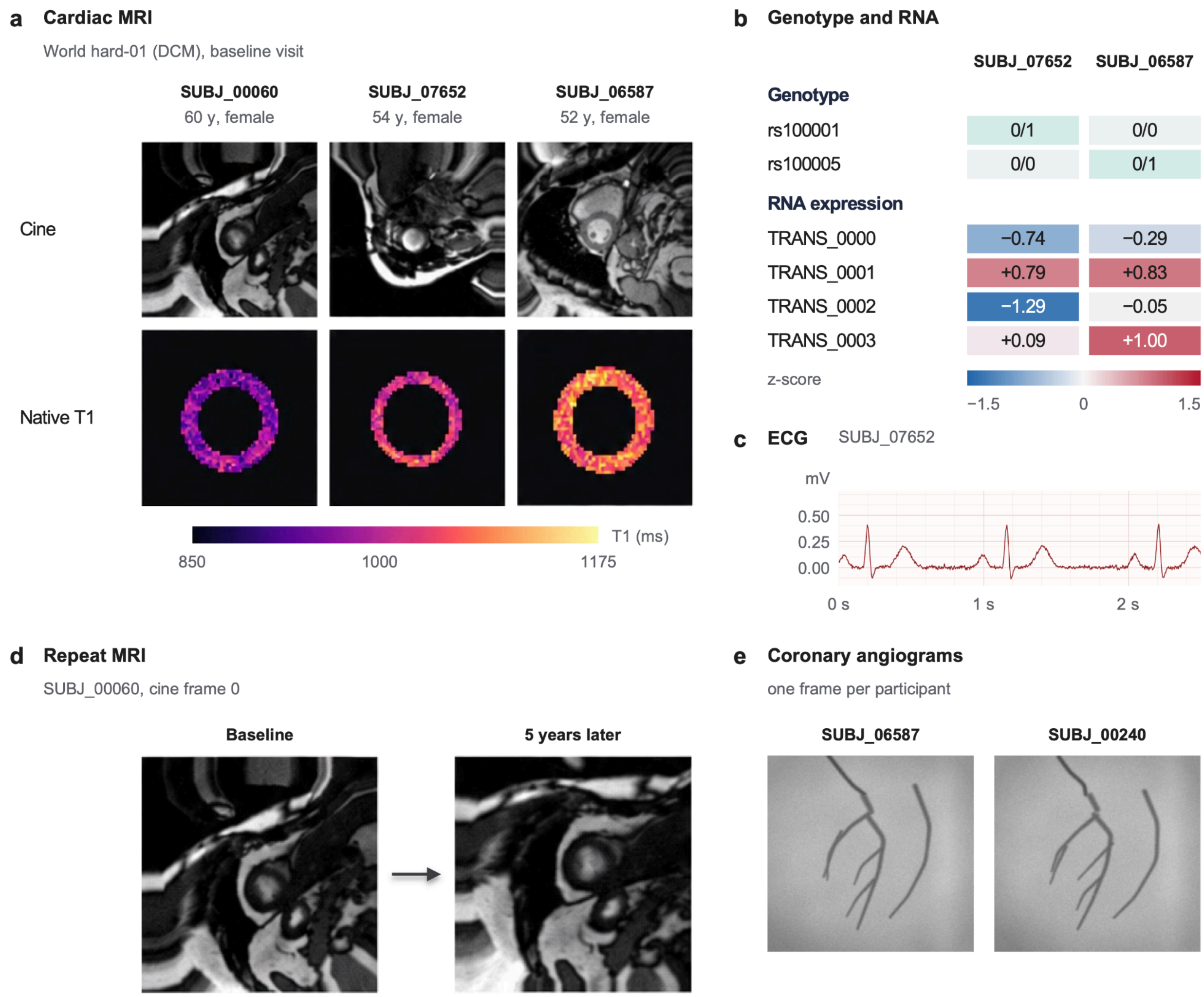

**Figure 2 | The simulated biobank: example data from world hard-01 (DCM). a,** Representative cardiac MRI showing a cine frame and a native T1 map for each of three participants. **b,** Linked genotype and RNA-expression measurements for two participants. **c,** Representative ECG trace. **d,** Repeat cardiac MRI from the same participant at baseline and 5 years. **e,** Representative coronary angiograms from two participants.

## 2. Related work

**Biobank-based drug target discovery**. Population biobanks have enabled the analysis of genetic, imaging, and longitudinal data for drug target discovery. Proteogenomic studies have mapped genetic determinants of circulating proteins and connected these measurements to cardiovascular disease phenotypes (Lind et al., 2024; Reddy et al., 2025; Sun et al., 2023). Building on these resources, instrumental-variable (IV) methods such as Mendelian randomization, along with colocalization and related genetic approaches, have been used to distinguish proteins associated with disease from those with stronger causal evidence (Henry et al., 2022; Rasooly et al., 2023, 2025). More broadly, human genetic evidence has been associated with an increased probability of success during drug development, which has motivated its wider use for target prioritization (Duffy et al., 2024; King et al., 2019; Minikel et al., 2024; Nelson et al., 2015). These studies establish a mature workflow for population-based drug target discovery. However, the analysis itself is still typically carried out by human investigators. In DrugTargetWorld, by contrast, this workflow becomes a sequential decision problem in which an agent must determine which analyses and experiments to perform under finite resources.

**Computational phenotyping in population biobanks**. Each of these analyses depends on how the disease phenotype is defined, and computational phenotyping is increasingly used to derive such phenotypes in population-scale biobanks. In the UK Biobank, automated processing and analysis of cardiac and brain imaging have enabled the derivation of quantitative phenotypes that would be impractical to obtain through manual measurement (Aung et al., 2022, 2023; Gomes et al., 2024; Smith et al., 2021). Other studies have constructed phenotypes from imaging and longitudinal tabular data (An et al., 2023; Gong et al., 2021, 2023; L. Yang et al., 2023; Z. Yang et al., 2026). These studies show that phenotype construction can change which biological associations are subsequently discovered. In DrugTargetWorld, phenotype construction is part of the task itself, because latent disease severity is never observed and the agent must derive its own participant-level disease phenotype from the released data.

**Scientific agents and benchmarks**. A separate line of work evaluates AI agents directly, and recent benchmarks such as ScienceAgentBench and DiscoveryBench have moved from evaluating static scientific knowledge toward testing whether agents can execute research workflows (Chen, 2025; Majumder et al., 2024). BixBench evaluates agents on real bioinformatics analyses, while BixBench3 extends this work to full study workflows in which agents start from raw biological data and reproduce outputs from published studies (Koch, 2026; Mitchener et al., 2025). In BixBench3, however, the research question and analysis methods are largely specified for the agent (Koch, 2026). In GeneBench-Pro, agents receive a biological dataset and a target estimand for a defined question and must then choose the appropriate statistical analyses across several dependent steps (Li & Ho, 2026). MRAgent automates Mendelian randomization analyses across existing genetic resources, thereby addressing one part of the broader drug target discovery process (Xu et al., 2025). Evaluating the whole workflow matters particularly for drug target discovery, where experimental validation is costly in both money and time, and agents must therefore prioritize which candidates merit testing. In contrast to these benchmarks, DrugTargetWorld leaves the research strategy across the entire workflow to the agent, which must construct the disease phenotype, determine which analyses and experimental evidence to acquire, and integrate potentially conflicting evidence. The agent then nominates causal drivers and their intervention directions for scoring against the hidden ground truth.

**Interactive environments and agent learning.** A parallel line of literature evaluates agents through their interaction with an environment rather than against a fixed task. Procedural environments have been used to test whether agents can generalize to new settings (Cobbe et al., 2020). ScienceWorld brings this idea to science by requiring agents to perform a sequence of actions and run experiments rather than answer static questions (Wang et al., 2022). AgentGym and SWE-Gym extend this approach by using executable environments for both training and evaluation of agents (Pan et al., 2025; Xi et al., 2025). Aviary applies this idea to scientific agents and shows that they can improve through training via repeated interactions with scientific environments (Narayanan et al., 2024). ResidencyRL trains clinical agents with reinforcement learning through simulated multi-turn patient encounters (Liévin et al., 2026, doi:10.48550/arXiv.2608.07418). DrugTargetWorld applies this framework to population-scale drug target discovery. Each world is generated from a different hidden causal model, and the agent chooses which evidence to collect and which experiments to pursue.

**Causal discovery benchmarks.** Causal benchmarks likewise generate synthetic data from known causal systems, so that model conclusions can be compared with the ground truth. Earlier work focused on causal direction and graph recovery, as well as on questions about associations, interventions, and counterfactuals (Jin et al., 2023; Mooij et al., 2016; Zhou et al., 2024). CausalDS generates structural causal models and tests whether agents can draw the correct causal conclusions about the resulting data (Leban & Sun, 2026). CausalGame asks agents to design experiments in settings with hidden confounding, selection bias, and measurement error, and thereby tests their actions when causal evidence is misleading (Chen et al., 2026). In CausaLab, agents intervene in a synthetic laboratory and are scored both on whether they solve a prediction task and on whether they recover the underlying causal mechanism (J. Yang et al., 2026). DrugTargetWorld builds on these ideas but treats causal reasoning as one part of drug target discovery, in which the agent must also construct the disease phenotype and separate causal drivers from non-causal proteins among thousands of candidate proteins.

**Table 1 | Comparison of DrugTargetWorld with scientific agent benchmarks, training environments, and causal discovery benchmarks.** Representative benchmarks are compared across design dimensions relevant to end-to-end evaluation of scientific agents. The table characterizes task structure rather than benchmark difficulty.

| Benchmark | Task | Method choice | Sequential | Simulated | Ground truth | Biobank | Trainable |
|---|---|---|---|---|---|---|---|
| ScienceAgentBench | Scientific analysis | Defined | Limited | No | No | No | No |
| DiscoveryBench | Scientific discovery | Defined | Limited | Variable | Variable | No | No |
| BixBench | Bioinformatics | Defined | Yes | No | No | No | No |
| BixBench3 | Full study execution | Guided | Yes | No | No | No | No |
| GeneBench-Pro | Genetic analysis | Open analysis | Yes | Yes | Yes | No | No |
| Aviary | Scientific tasks | Task specific | Yes | Variable | Variable | No | Yes |
| ResidencyRL | Clinical care | Scenario defined | Yes | Yes | Scenario | No | Yes |
| Causal benchmarks | Causal discovery | Variable | Variable | Yes | Yes | No | Variable |
| DrugTargetWorld | Drug target discovery | Open strategy | Yes | Yes | Yes | Yes | Yes |

**Notes.** "Method choice" describes how much of the analysis the agent selects. "Defined" means that the task specifies the analysis, "Guided" means that the research question and high-level methods are given, and "Open strategy" means that the agent chooses the phenotype, hypotheses, analyses, experiments, and stopping point. "Sequential" denotes multiple dependent research decisions within a task, and "Limited" denotes one or a few analysis steps. "Simulated" denotes synthetic data generation. "Ground truth" denotes known data-generating and intervention effects, and "Scenario" means that each simulated case defines the correct outcome. "Biobank" denotes linked participant-level population data. "Trainable" denotes an environment that can supply a verifiable reward for training. "Variable" means that the entry depends on the task.

## 3. Methods

DrugTargetWorld generates simulated population biobanks, which we call worlds, from hidden causal models whose causal drivers, non-causal proteins, and planted biases are known. Agents receive only the released data. From these data, an agent must extract relevant measurements, construct a disease phenotype, identify causal drivers and their intervention directions, and decide which analyses or experiments to perform within its experimental budget. An episode is one evaluation of one agent in one world under one experimental budget condition. At the end of each episode, the agent's submission is scored against the hidden causal ground truth, which is possible only because each world is generated with a known answer.

### 3.1 World generation

Each world is generated from a structural causal model that specifies genetic variation, molecular measurements, latent disease severity, observed disease phenotypes, longitudinal outcomes, and responses to experimental intervention. A procedural seed determines the world-level parameters, including the causal drivers, their effect sizes, the disease archetype, and the planted biases. These parameters are sampled once and held fixed for every agent evaluated in that world, so that score differences between agents within a world reflect their analyses rather than differences in the data.

Participant covariates are drawn from population distributions, and genetic variants are generated with allele-frequency and linkage-disequilibrium structure. Molecular measurements are then generated from cis and trans genetic effects, shared latent biological factors, participant covariates, and assay noise. To keep these distributions plausible, the evaluated panel was calibrated to aggregate distributions reported in the Multi-Ethnic Study of Atherosclerosis (MESA), including published MESA cohort summaries and TOPMed-linked MESA data (Heckbert et al., 2006; Liu et al., 2013; Marques et al., 2022). These sources informed demographic distributions, which participants received MRI, and the means, standard deviations, and selected correlations of cardiac MRI traits; TOPMed-linked MESA data additionally informed the allele-frequency and linkage-disequilibrium structure of the genetic variants and the protein missing-data rate. Only cohort-level summary statistics were used, and no MESA participant data were resampled into any world. The causal structure, molecular effects, and intervention effects were designed rather than estimated from MESA (Supplementary Methods S1.1). We first generate each molecular measurement as a combination of cis and trans genetic effects, shared latent biological factors, participant covariates, and assay noise. For participant i and molecular feature j:

$$M_{ij} = z[\ \alpha_j G_{i,cis}(j) + \beta_j G_{i,trans}(j) + F_i^T \lambda_j + \eta_j^T C_i + \varepsilon_{ij}\ ]$$

Transcript and protein levels are generated with the same structure, so they can share genetic and biological influences without a fixed transcript-to-protein causal relationship, and the agreement between them differs across worlds.

Disease is generated through a latent disease severity $L_i$, which is never released to the agent. For participant i, $L_i$ is generated from a hidden set of causal molecular drivers $\mathscr{D}$:

$$z[\ \sum_{j \in \mathscr{D}} w_j P_{ij} + \gamma^T C_i + \delta G_{i,direct} + \varepsilon_i\ ]$$

Here, $P_{ij}$ is the abundance of molecular driver j, $C_i$ contains participant covariates, $G_i$,direct captures genetic effects on disease not mediated through the measured molecular drivers, and $\varepsilon_i$ represents residual variation. The number and identity of causal drivers, their effect sizes, instrument strengths, and disease architecture vary across worlds, whereas every protein has a single cis instrument, so estimates cannot be compared across independent instruments. In the evaluated panel, one null world contained no causal driver, and each of the other 19 contained between two and six (Supplementary Table S3).

Latent disease severity is then expressed through multiple observable phenotypes, whose pattern depends on the disease archetype. The evaluated panel contained six worlds each of the dilated cardiomyopathy (DCM), hypertrophic cardiomyopathy (HCM), and heart failure with preserved ejection fraction (HFpEF) archetypes, and two of the ischemic archetype:

$$z(Y_{ik}) = \alpha_{a,k} L_i + \beta_k^T C_i + \delta_k P_{i,T8} + \varepsilon_{ik}$$

where $Y_{ik}$ is an observed phenotype k and α denotes the disease archetype. The archetype-specific coefficient $\alpha_{a,k}$ allows the same latent disease severity to produce different patterns across medical imaging, physiologic signals, clinical measurements, electronic health record (EHR) diagnoses, longitudinal visits, and survival outcomes. The final term represents the harmful surrogate (bias T8, described below), a protein that improves imaging phenotypes without being a causal driver, and one such protein was planted in each of the 20 evaluated worlds.

Beyond this core model, the generator can plant nine predefined biases, labeled T1 to T9 (Figure 3), which represent recognized ways in which observational biobank evidence can mislead, including the confounding, reverse causation, selection bias, pleiotropy, and measurement artifacts noted in the Introduction. These are confounding (T1), reverse causation (T2), selection bias (T3), causal non-identifiability (T4), imaging batch effects (T5), benign physiologic remodeling (T6), instrument pleiotropy (T7), surrogate–outcome discordance (T8), and assay unit mixing (T9). T4 and T6 are not biases in the strict statistical sense, but they also make observational evidence misleading. Each bias alters the causal or measurement process rather than adding random noise, so the proteins it affects can appear scientifically plausible. In the evaluated panel, T5, T6, and T8 were present in all 20 worlds, T1 in 18, T2 in 14, T3 in 12, T7 in 8, T4 in 6, and T9 in 5. In addition, 15 worlds contained a delayed-effect driver, a causal driver whose effect is expressed mainly over five years of follow-up and which is paired with a non-causal partner that tracks the later disease. T4 and the delayed-effect design both create a pair of proteins that observational data cannot separate, which we call an ambiguous pair. T4 creates a shared-instrument pair and the delayed-effect design a delayed-effect pair, and the causal-confidence score evaluates these ambiguous pairs (Section 3.5). The implementation of each bias is detailed in Supplementary Methods S1.4.

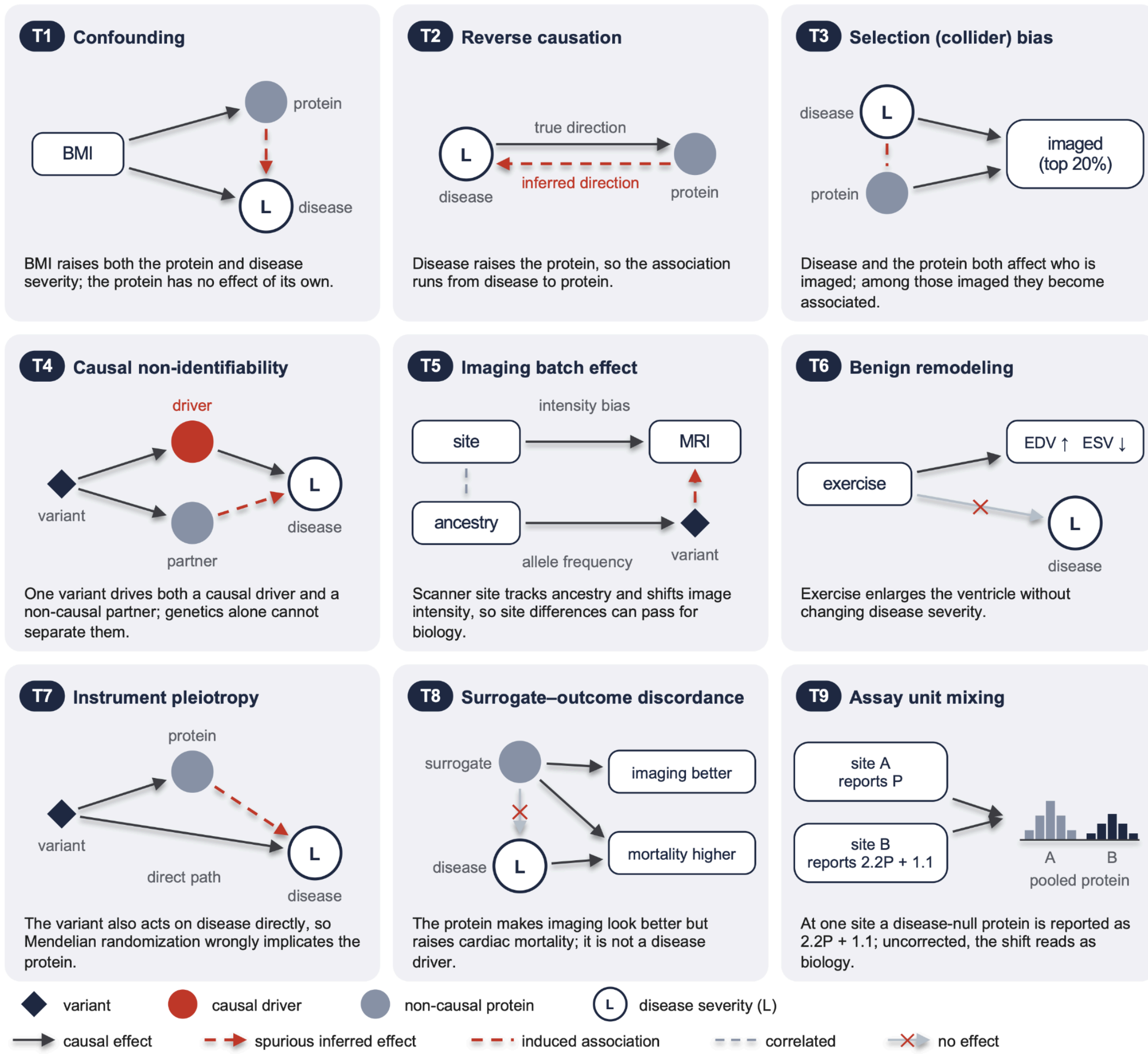

**Figure 3 | Planted biases in DrugTargetWorld.** Each card shows how a bias is built into a world's causal or measurement model and the misleading signal it creates. ◆, genetic variant; red circle, causal driver; gray circle, non-causal protein; L, latent disease severity; solid arrow, causal effect; red dashed arrow, spurious effect an agent could infer; red dashed line, induced association; gray dashed line, correlation; ✕, no causal effect. **T1**, confounding: BMI raises both a non-causal protein and L. **T2**, reverse causation: L raises the protein. **T3**, selection (collider) bias: L and the protein both raise the chance of being imaged (top 20% of a selection score), inducing an association among imaged participants. **T4**, causal non-identifiability: one variant drives a causal driver and a non-causal partner, so genetic evidence alone cannot separate them. **T5**, imaging batch effect: scanner site is correlated with ancestry and adds a site-specific intensity bias to images, so ancestry-related variants can appear to affect imaging. **T6**, benign remodeling: exercise increases end-diastolic and decreases end-systolic volume without changing L. **T7**, instrument pleiotropy: the protein's cis variant also affects L directly, violating the exclusion restriction. **T8**, surrogate–outcome discordance: the protein improves imaging phenotypes but increases cardiac mortality and is not a driver. **T9**, assay unit mixing: at one site a disease-null protein is reported as 2.2P + 1.1. Each world contains a subset of these biases; implementations are in Supplementary Methods S1.4.

### 3.2 Synthetic biobank

For each world, the causal model generates a participant-level biobank (Figure 2) that links genetic, proteomic, transcriptomic, and metabolomic data with raw cardiac MRI, electrocardiogram (ECG) features, coronary computed tomography (CT) features, EHR diagnoses and medications, mortality, and major adverse cardiovascular events (MACE). Each evaluated world contains 54,000 participants, 2,941 proteins, and 8,192 genetic variants, with cardiac MRI for 10,800 participants (20%). These dimensions are benchmark design choices inspired by the scale of UK Biobank rather than estimates from its data (Supplementary Methods S1.1). As in MESA, the imaged participants are not a random sample, because imaging participation depends on age, body mass index, ancestry group, and study site, and in the 12 worlds carrying T3 also on latent disease severity and a selected protein. A random 20% of participants attended a second visit about five years later, at which covariates and proteomics were measured again, with repeat MRI for those who had been imaged. A per-protein targetability table is also released, describing each protein's subcellular localization, binding pocket, paralog redundancy, genetic constraint, and tissue specificity, but it does not contribute to the score. Every protein and transcript carries an anonymized identifier with no mapping to a real gene, so an agent cannot draw on published knowledge about any protein and must reach its conclusions from the released data alone.

Neither latent disease severity nor any imaging-derived measurement is released. Phenotype construction, the choice and combination of released measurements into a participant-level disease phenotype, is therefore part of the drug target discovery task. An agent may build its phenotype from any released data, but an agent that uses cardiac MRI must first extract quantitative features from the raw cine and native T1-mapping images. These images are rendered on anatomical templates from the public ACDC cardiac MRI dataset (Bernard et al., 2018, doi:10.1109/TMI.2018.2837502) and deformed to each participant's simulated morphology. The data also contain missing modalities, measurement noise, genetic instruments of varying strength, and variable agreement between transcript and protein levels, imperfections that agents would also encounter in a real biobank.

### 3.3 Agent environment and harness

Within each episode, the agent works in a computational sandbox that contains the released biobank files and a writable working directory. It analyzes the data by writing Python programs that the harness executes, so that every analysis is recorded as an explicit computation.

The harness is the software interface between the agent and the world. We used a deliberately minimal harness so that differences in score reflect the agents' own research strategies rather than scaffolding that differs between systems. It supplies the task instructions, the two purchasable virtual experiments, the output of each program executed within the current episode, and the required submission format (Supplementary Methods S3.1). The instructions state that a small number of proteins causally drive cardiac disease and name the five sources of bias by which other proteins can appear associated, namely confounding, reverse causation, selection into the imaged subcohort, pleiotropy, and measurement artifact, but they do not identify any protein. Apart from a helper function for reading genotypes, the harness provides no analysis tools or analysis plan, and it gives no feedback on scientific correctness.

The released data are formatted to resemble a population biobank. Proteomics, transcriptomics, metabolomics, diagnoses, medications, and other structured measurements are provided as tabular files, with covariates documented in a data dictionary. Genotypes are provided in variant call format (VCF), and each imaged participant has one file containing cine images and native T1 maps. The data are linked by participant identifiers, and the agent must join them itself. Proteins, transcripts, and metabolites are identified only by anonymous labels (for example, PROT_0412), so published knowledge about named molecules cannot point an agent to a causal driver.

Each episode allows at most 30 turns, and each turn may contain one Python program. This limit was raised from an initial 15 turns after pilot episodes showed that the shorter limit, rather than the agents' own strategies, determined when most episodes ended. After each turn, the harness returns the program's output or error, the contents of the working directory, and the remaining experimental budget and turns. Files persist across turns. An episode ends when the agent writes a parseable submission or reaches the turn limit, and whatever submission and phenotype files are present at that point are scored (Supplementary Methods S2.4). Because the instructions invited agents to revise a provisional submission, this rule could end an episode earlier than an agent intended. The complete interaction history and executed code are recorded (Supplementary Methods S3.2), and Figure 1d,e follows one such episode from the agent's turn-by-turn activity to the scoring of its claims.

### 3.4 Action space and experiments

Laboratory experiments are expensive, so a useful agent must judge when an experiment is worth its cost. We therefore evaluated each agent under three experimental budget conditions. The observational-only condition ($0) allows no experiments, the limited condition ($450,000) pays for one knockdown or three cell perturbations, and the expanded condition ($2 million) pays for up to five knockdowns. These amounts are intended as an ordinal scale of experimental access rather than as estimates of real experimental costs. The limited budget thus forces a choice between one knockdown and three cell perturbations, whereas the expanded budget allows several candidates to be tested by knockdown. The observational-only condition measures what an agent can conclude from the biobank alone. Both types of experiment are described below.

In all three conditions, analyses of the released biobank are free and unrestricted, so an agent can apply any statistical method that it can implement in Python.

Agents with an experimental budget could purchase two types of simulated intervention, each applicable to any protein. A cell perturbation cost $150,000 and measured the effect at the first visit, whereas an in vivo knockdown cost $400,000 and measured the effect over five years of simulated follow-up (Supplementary Methods S1.5). Because the cheaper experiment cannot detect a delayed effect, choosing between the two is part of the task. For either experiment, perturbing protein j returned the change in four outcomes relative to the unperturbed state:

$$\Delta_j = (\Delta\ \text{cavity}_r, \Delta\ \text{wall}_t, \Delta\ \text{EF}, \Delta\ \text{surv}_{5y})$$

where each Δ is the mean outcome under perturbation minus the mean outcome without perturbation. Cell perturbation was based on a smaller virtual experiment and included measurement noise, whereas knockdown used a larger virtual experiment and returned the longer-term intervention effect without additional measurement noise.

An agent's work can therefore be viewed as a sequence of decisions, each chosen from the evidence gathered so far and the budget that remains, a formulation that also defines the policy that training in this environment would optimize:

$$a_t \sim \pi(a \mid s_t, B_t)$$

where $s_t$ represents the evidence accumulated by turn t, $B_t$ is the remaining experimental budget, and $a_t$ is the next analysis or experiment selected by the agent. Experimental actions update the remaining budget according to

$$B_{t+1} = B_t - c(a_t)$$

with $c(a_t) = 0$ for analyses of released biobank data.

At the end of each episode, the agent submits its claims. It nominates candidate causal drivers, each with an intervention direction (inhibit, activate, or unknown) and an outcome-alignment label. The label predicts whether a treatment that improves the disease phenotype through that protein would also improve survival (aligned), worsen survival (misaligned), or have an uncertain effect (unknown). The agent also lists rejected proteins, each with the source of bias that makes it appear causal, and records abstentions that name small sets of proteins that it cannot distinguish. Finally, it submits a participant-level disease phenotype, and Section 3.5 describes how each element of the submission is scored.

### 3.5 Scoring

Each part of the final submission is scored against the hidden ground truth. More than half of the 100 points reward finding the causal drivers with confidence matched to the evidence, through target identification (30 points) and causal confidence (25 points). The remaining points reward the intervention direction for each driver (20 points), bias identification (15 points), and phenotype construction (10 points). Safety adds no points, but advancing the harmful surrogate as beneficial incurs a 30-point penalty, equal to the full weight of target identification. These components correspond to the judgments that the task requires. Phenotype construction scores how well the agent measures the disease, target identification and bias identification score whether it separates causal drivers from non-causal proteins, causal confidence scores whether its claims are as strong as its evidence allows, and intervention direction scores the actionable conclusion for each driver. The final score is the sum of the five component scores and the safety penalty, bounded between 0 and 100:

$$S_{total} = \text{clip}(S_{target} + S_{confidence} + S_{mechanism} + S_{direction} + S_{phenotype} + P_{safety}, 0, 100),$$

where $S_{mechanism}$ denotes the bias-identification score. Each component is defined below. Sensitivity of the model ranking to removing phenotype credit is reported in Supplementary Methods S2.6.

Target identification (30 points) rewards finding the causal drivers without nominating proteins that are not drivers.

$$S_{target} = 30 \times \text{Recall} \times \text{Precision}.$$

Recall is the fraction of causal drivers that the agent nominated, and precision is the fraction of its nominations that are causal drivers. Because the two are multiplied, nominating many proteins to raise recall lowers precision, and full credit requires the exact set of drivers. Planted non-causal proteins and the harmful surrogate count as false positives.

Causal confidence (25 points) measures whether the agent's causal claims match the strength of its evidence. It is informative mainly in the 17 worlds that contain at least one ambiguous pair created by T4 or by the delayed-effect design, in which observational data cannot distinguish the causal member from its partner. In the two non-null worlds without such a pair, the score is a rescaling of target identification:

$$S_{confidence} = 25 \times S_{target} / 30.$$

In worlds containing K planted ambiguous target pairs,

$$S_{confidence} = (1/K) \times \Sigma_{k=1}^{K} c_k,$$

where $c_k$ is the credit for ambiguity k. A pair receives 25 points when the agent correctly identifies the causal protein and has experimentally tested that same protein. Recognizing that the pair cannot yet be resolved receives partial credit of up to 17.5 points, whereas an unsupported causal choice receives no credit. This ordering credits an explicit statement of unresolved ambiguity above an unsupported causal choice, because observational data alone cannot support a choice between the members of such a pair.

Full credit for a pair therefore requires an experiment on the protein that the agent names as causal. In the observational-only condition, the maximum credit per pair is consequently 17.5 points.
Bias identification (15 points) measures whether the agent rejects the planted non-causal proteins, which only appear causal, and names the source of bias that produced each one:

$$S_{mechanism} = 15 \times Recall_{rejection} \times Precision_{rejection}.$$

A rejection counts as correct only when it names a planted non-causal protein together with its source of bias, chosen from confounding, reverse causation, selection bias, pleiotropy, and measurement artifact. Scored non-causal proteins are those planted by T1, T2, T3, T7, and T9, whereas the harmful surrogate is handled by the safety penalty. As in target identification, recall and precision are multiplied, so rejecting many proteins indiscriminately earns little credit.

Intervention direction (20 points) measures whether the agent correctly predicts whether a drug should inhibit or activate each causal driver. The score is averaged over all causal drivers in the world, excluding the causal member of an ambiguous pair on which the agent abstained. For each causal driver,

$$d_j = \begin{cases} 1 & \text{correct direction} \\ 0.5 & \text{explicitly unknown} \\ 0 & \text{incorrect or missing} \end{cases}$$

The direction score is then

$$S_{direction} = 20 \times \text{mean}(d_j).$$

A correct direction receives full credit, and an incorrect direction or a driver that was not nominated receives none. An explicit unknown receives half credit, and a nominated driver with no valid direction is treated as unknown. Half credit equals the expected value of guessing, so declaring uncertainty is neither rewarded nor penalized relative to a guess.

Phenotype construction (10 points) measures how closely the agent's submitted participant-level phenotype tracks latent disease severity. Let $|r|$ denote the absolute correlation between the submitted phenotype and latent disease severity. With b the correlation achieved in that world by the environment's own MRI-derived baseline phenotype (Supplementary Table S3 and Supplementary Methods S2.5), $S_{phenotype} = 10 \times \text{clip}((|r| - b)/(0.85 - b), 0, 1)$; in a world without a valid recorded baseline,

$$S_{phenotype} = 10 \times \min(|r|/0.85, 1).$$

This fallback was never used, because a baseline was recorded for all 20 evaluated worlds. The baseline phenotype is computed by the environment from the released cine and native T1-mapping images, and agents cannot access it or its value. Across worlds, b ranged from 0.40 to 0.76, and the score rises linearly from zero at b to full credit at a fixed ceiling correlation of 0.85. Because every released measurement carries noise, no phenotype can correlate perfectly with latent disease severity, so full credit is given below a correlation of 1, and the best phenotypes in this evaluation approached the 0.85 ceiling (Supplementary Table S7B). An agent that merely reproduces a standard imaging summary therefore gains no benefit. Correlations are computed on a sealed random sample of 20% of all 54,000 participants, drawn without regard to imaging, using the participants in that sample for whom the submitted phenotype has a value. A phenotype derived only from cardiac MRI is therefore scored on the imaged members of the sample, between 2,059 and 2,278 participants depending on the world, and the baseline phenotype is scored in the same way.
Finally, the score penalizes the most consequential error: recommending a target whose benefit for the disease phenotype would come at the cost of survival. The harmful surrogate planted in every world improves imaging phenotypes but increases cardiac mortality and is not a causal driver. If the agent nominates the harmful surrogate and labels it aligned, predicting that treatment through it would also improve survival, 30 points are subtracted. Labeling it misaligned or unknown avoids the penalty, but the nomination still counts as a false positive in target identification:

$$P_{safety} = \begin{cases} -30 & \text{harmful surrogate claimed as outcome-aligned} \\ 0 & \text{otherwise.} \end{cases}$$

Special handling of null worlds, ambiguity abstentions, claim limits, and malformed outputs is described in Supplementary Methods S2.1–S2.4.

**3.6 Evaluation**

We evaluated nine language models as agents across 20 worlds and three experimental budget conditions, with one episode per model, world, and budget. The panel of 20 worlds was designed to cover all four disease archetypes and to include every bias in several worlds, while keeping the evaluation of nine agents under three experimental budgets tractable. This design yielded 540 episodes, 60 per model and 180 per budget condition, of which 360 could purchase experiments. The models were Opus 5, GPT-5.6 Sol, Sonnet 5, Haiku 4.5, GPT-OSS-20B, Qwen3-Coder-30B-A3B, GLM-4-32B, Qwen3-8B, and Devstral Small. Serving details are listed in Supplementary Table S1 and inference settings in Supplementary Methods S3.2. The open-weight models were chosen to be small enough to run on our local compute infrastructure while remaining capable enough to be candidates for future training on DrugTargetWorld. All models used the same harness, instructions, and sandbox.

The primary outcome was the total score. Secondary outcomes were the individual component scores, target precision and recall, experimental spending, model inference cost (Figure 4d), and completion of major research steps.

To characterize research workflows and failure modes, we also analyzed the complete recorded trajectories. A static audit of the executed code identified which data each program read and which calculations it performed, and it counted an analysis only when the calculation was executed, not when a library was imported or a method was named. In this exploratory analysis, the executed operations were grouped into six analysis families, each identified by a code signature, namely correlation or regression, covariate-adjusted regression, genotype-based computation, Mendelian randomization or instrumental-variable calculations, cross-molecular analysis, and repeat-visit or survival analysis. Separate indicators recorded phenotype creation, experiment purchases, nominations, submission, and references to returned experimental results. These measures describe what the agents computed and which data they accessed, not whether an analysis was statistically valid or correctly interpreted.

Statistical analysis. The world was the unit of analysis (n = 20), because a model's three budget episodes within a world share the same causal structure. For pairwise model comparisons, scores were first averaged across the three budgets within each world and model, and paired differences were then calculated across worlds. We report mean paired differences, 95% t-based confidence intervals (CIs), and exploratory two-sided paired t-tests, with exact sign tests and Wilcoxon signed-rank tests as sensitivity analyses. P values for these comparisons are not adjusted for multiplicity. Budget contrasts were likewise paired by world.

The trajectory analyses were exploratory. We compared the frequencies of the six analysis families between Opus 5 and GPT-5.6 Sol using paired world averages, 20,000 world-bootstrap resamples, and exact sign-flip tests with Holm correction. Each analysis family was then related separately to exact target recall and to the fraction of incorrect nominations, adjusting for model, budget, and world. Uncertainty was estimated with the CR2/Satterthwaite method, clustered over the 19 non-null worlds, and Holm correction was applied across these 12 exploratory tests (six analysis families and two outcomes). Incorrect-nomination fractions exclude episodes with no nominations. The pooled association between the number of core workflow milestones reached and score was summarized with Spearman's rank correlation and is descriptive. Finally, because there was one replicate per model, world, and budget, reported standard deviations describe variation across worlds and budgets rather than run-to-run stochastic variability.

## 4. Results

### 4.1 Leading agents recovered many causal drivers but remained unreliable across the full workflow

Opus 5 and GPT-5.6 Sol recovered similar fractions of causal drivers (target recall 0.64 for both) but achieved mean overall scores of only 39.98 and 35.38 of 100, reflecting failures elsewhere in the workflow.

Agents entered each simulated biobank without a prescribed analysis pipeline and had to develop their own strategy for identifying drug targets. Across all 540 episodes, the mean total score was 14.1 of 100, and fewer than half of the episodes (258, 47.8%) scored above 0. Sonnet 5 and Haiku 4.5 followed the two leading agents with means of 21.33 and 12.92, whereas the five open-weight models had mean scores of 0.81 to 7.41, each with a median of 0 (Figure 4a, Supplementary Table S4, and Supplementary Note S3).

For reference outside the model evaluation, two human-guided episodes, in which an investigator directed a language model, scored 37.5 and 81.67 and are included in the supplement as exploratory references. They were designed as qualitative workflow references rather than as a matched human baseline, and only the 37.5-point episode followed the standard evaluation protocol (Supplementary Notes S1 and S2).

Performance at the top of the range was confined to a few episodes. Six of the 540 model episodes (1.1%) scored above 80, five in heldout-hfpef-02 and one in heldout-hcm-01, and all six nominated exactly the two planted causal drivers with no false positives. Both worlds contain a T4 pair, in which a causal driver and a non-causal partner share a genetic instrument. The full 25 causal-confidence points therefore require a purchased experiment on the causal member, and none of the six episodes was observational. All six episodes also fell in the cells excluded from the sensitivity analysis for the follow-up MRI data-generation error described among the limitations, although excluding the affected reads did not change the order of the leading agents (Supplementary Table S11D).

Scores also varied with world design and were highest in the worlds with two causal drivers. They averaged 23.4 in the four two-driver worlds, compared with 11.2–12.2 in worlds with three to six causal drivers and 10.3 in the eight worlds designed to be difficult (Supplementary Note S3).

After averaging the three budgets within each world, Opus 5 exceeded GPT-5.6 Sol by a paired mean of 4.6 points across the 20 worlds (95% CI −0.26 to 9.45) and scored higher in 16 of them (Figure 4e). The paired t-test gave $P = 0.062$ and the exact sign test $P = 0.012$, so the difference is suggestive but not statistically conclusive. By contrast, every other pair among the four models differed when assessed through an application programming interface (API), with 95% CIs excluding zero (Supplementary Table S5). Because each model was run once per world and budget, these comparisons do not account for run-to-run variation.

The difference between the two leading agents lay mainly in phenotype construction, where Opus 5 averaged 7.2 of 10 points and GPT-5.6 Sol 3.2. It lay to a lesser extent in bias identification and was partly offset by a higher target-identification score for GPT-5.6 Sol (Figure 4c, Supplementary Table S4B). Without phenotype credit, the difference fell to 0.9 points (95% CI −3.64 to 5.46), and the ranking of all nine agents was unchanged (Supplementary Methods S2.6).

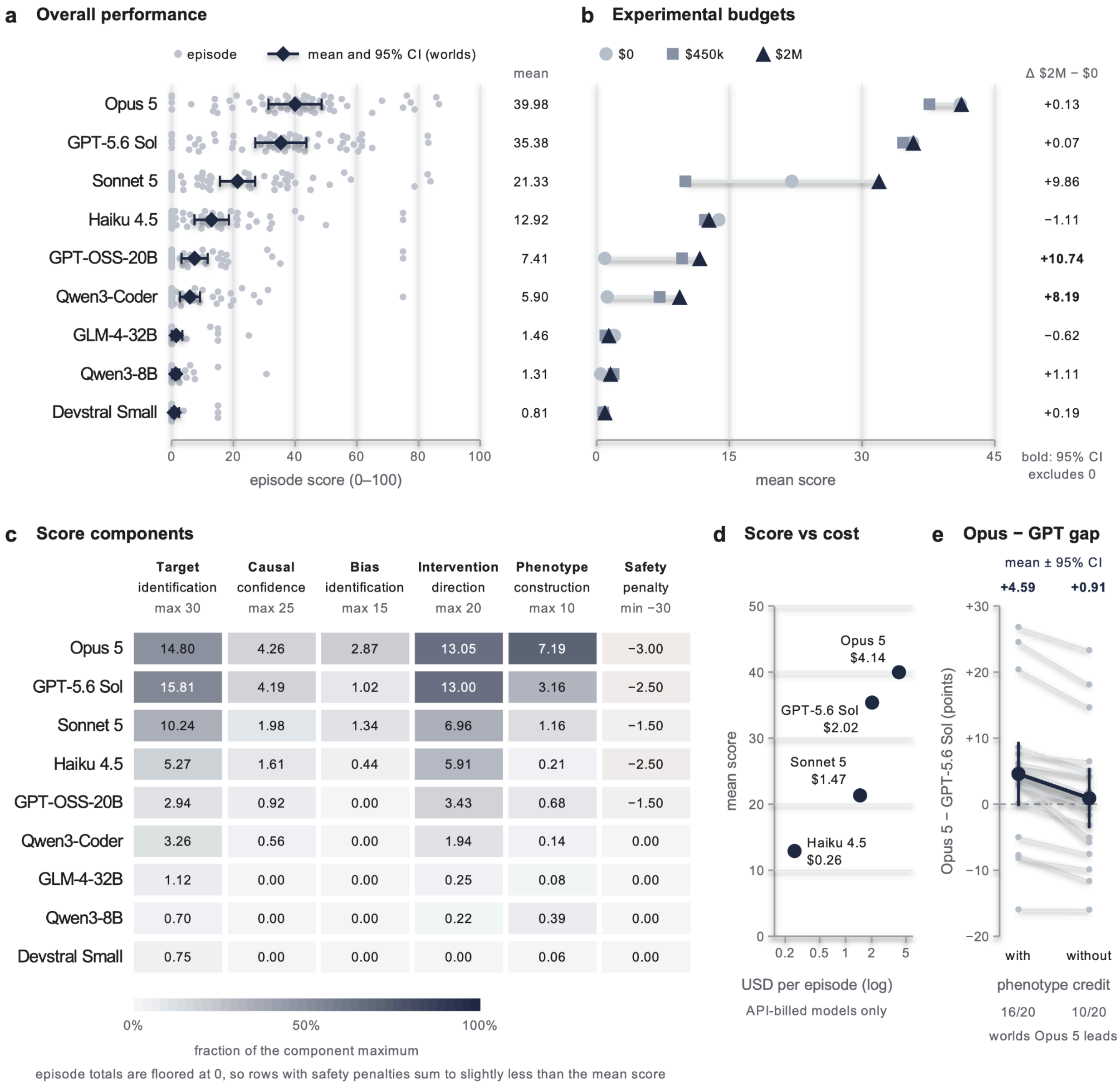

**Figure 4 | Benchmark performance. a,** Episode scores (dots; 60 per model) with the mean and a t-based 95% CI computed over worlds (scores averaged across the three budgets within each world; n = 20). **b,** Mean score under the observational-only ($0), limited ($450k) and expanded ($2M) experimental budgets; right, expanded minus observational, in bold where the paired world-level 95% CI excludes zero. **c,** Mean points per scoring component; cell shade is the fraction of the component maximum (navy) or of the maximum penalty (amber). Episode totals are floored at 0, so the components of models with safety penalties sum to slightly less than their mean score. **d,** Mean score versus API-billed cost per episode (log scale); self-hosted models are not shown. **e,** Opus 5 minus GPT-5.6 Sol in each of the 20 worlds, with and without phenotype-construction credit (gray lines); navy, mean and 95% CI (+4.60, −0.26 to 9.45; +0.91, −3.64 to 5.46). Counts give the number of worlds in which Opus 5 scored higher.

### 4.2 Phenotype construction separated the two leading agents

Opus 5 earned positive phenotype credit in 56 of 60 episodes, and GPT-5.6 Sol in 34 of 60. The phenotypes that correlated most strongly with latent disease severity were more often those that combined several imaging features and were checked against clinical endpoints before submission.

The first judgment the task required was how to measure the disease from the released data. Cardiac MRI was released only as raw images, without derived volumes, ejection fraction, or mass, so agents had to extract imaging features, assess their relevance, and combine them into a participant-level measure of disease before prioritizing targets. They could also draw on ECG, coronary CT, and outcome data (Supplementary Methods S3.1).

Performance on this step varied substantially across agents. Beyond the two leading agents, Sonnet 5 earned positive phenotype credit in 12 of 60 episodes and every other model in seven or fewer (Supplementary Note S3). Among the 349 phenotypes scored against latent disease severity, the mean correlations were 0.785 for Opus 5, 0.637 for GPT-5.6 Sol, and 0.502 for Sonnet 5 (Supplementary Table S7A). For comparison, the environment's MRI-derived baseline phenotype reached correlations of 0.40 to 0.76 across worlds, and full credit required 0.85 (Section 3.5).

These differences were associated with how agents constructed and validated their phenotypes. Phenotypes whose weights for several imaging features were learned from clinical outcomes and other observed disease signals correlated more strongly with latent disease severity than composites whose weights the agent chose by hand, and more strongly than single imaging measures (Supplementary Table S7B). The leading agents also checked the phenotype against other signs of disease already present in the dataset before submitting it. Among episodes with a scorable phenotype, those that performed this validation reached a mean correlation of 0.663, compared with 0.402 among those that did not (Supplementary Table S7C). Because these comparisons pool episodes across models, they are confounded with model identity.

The validation checks drew on mortality, coronary CT calcium or stenosis, MACE, ECG features, and EHR diagnoses, and in the source-linked traces they were used to choose among candidate phenotypes, not only to confirm the final one (Supplementary Table S7D). Because these endpoints are clinical or imaging measures rather than protein measurements, they keep the phenotype largely independent of the candidate proteins later tested against it. However, the T8 harmful surrogate affects cardiac mortality and MACE directly, so a phenotype fitted to these outcomes can partly incorporate the surrogate's signal. In the traces of the two leading agents, a few episodes compared candidate phenotypes by their strongest protein correlations, but none chose a phenotype by its number of protein associations.

The traces illustrate these measurement choices. For example, Opus 5 in heldout-hcm-01 derived a measure of contraction timing from the cine images and revised its extraction when the first version proved unstable. It then submitted cross-validated ridge predictions of a composite of observable cardiac indicators, which earned full phenotype credit (Supplementary Note S3).

### 4.3 The leading agents based nominations on genetic instruments, but no agent reliably rejected non-causal proteins

Opus 5 and GPT-5.6 Sol based most final nominations on cis Mendelian randomization and usually checked instrument strength, yet no agent averaged more than 2.87 of 15 points for bias identification, which requires rejecting non-causal proteins with the correct source of bias.

After constructing a disease phenotype, agents faced the second judgment, distinguishing proteins merely associated with the phenotype from those with evidence of a causal link. Executed IV analyses, including Mendelian randomization, were detected in 41 of 60 Opus 5 episodes and 29 of 60 GPT-5.6 Sol episodes, compared with 15 episodes across the other seven models.

A separate trajectory annotation, which applies a broader criterion than the static code audit (Section 3.6), showed that the leading and lower-ranked agents prioritized targets differently. Opus 5 based its final nominations primarily on cis Mendelian randomization in 57 of 60 episodes and GPT-5.6 Sol in 50, and the two agents checked instrument strength in 59 and 57 episodes, respectively (Supplementary Table S8A). In contrast, Haiku 4.5, GPT-OSS-20B, and Qwen3-Coder-30B-A3B relied predominantly on marginal association and correlation-based screening (Supplementary Note S3). Instrument strength does not, however, establish the exclusion restriction, the requirement that the genetic instrument affects disease only through the protein. Moreover, with a single cis instrument per protein (Section 3.1), agents could not test this requirement by comparing estimates across independent instruments.

Although the leading agents differed in which analysis methods they used, no analysis family was associated with outcomes after correction for multiple testing. For example, GPT-5.6 Sol used cross-molecular operations, which combine at least two of the proteomic, transcriptomic, and metabolomic layers, more often than Opus 5 (57 versus 38 of 60 episodes, Supplementary Note S3). Across all agents, however, none of the six analysis families defined by the code audit (Section 3.6) was associated with target recall or with the fraction of incorrect nominations after Holm correction across 12 exploratory tests.

The leading agents differed from the weaker agents in how they revised their hypotheses. Nearly all revisions by Opus 5 and GPT-5.6 Sol followed new evidence and narrowed the candidate set, whereas several smaller models formed an early hypothesis and did not revisit it. Consistent with this narrowing, the two leading agents nominated 3.7 and 3.1 proteins per episode, close to the two to six causal drivers planted in each non-null world, whereas in a few episodes weaker models submitted far longer lists (Supplementary Table S8B and Supplementary Note S3). One episode illustrates this narrowing. In heldout-hfpef-02, GPT-5.6 Sol knocked down both members of a pair that shared a genetic instrument. It then nominated only the member whose knockdown changed the outcomes, and the episode scored 83.01.

The low bias-identification scores reflected how rarely non-causal proteins were rejected with the correct source of bias. The reverse-causation (T2) protein, which agents rejected most often, was rejected without being nominated in 121 of the 378 episodes from the 14 worlds that contain this protein (32.0%), and 54 (14.3%) also named the correct source of bias, including 31 of 42 Opus 5 and 4 of 42 GPT-5.6 Sol episodes. Because the bias-identification score multiplies the recall of correct rejections by their precision, additional incorrect rejections further lowered it.

Agents also differed in which planted proteins they nominated, and the pattern varied by bias and by model (Figure 5a). All models nominated the confounding (T1) and selection (T3) proteins in fewer than 9% of the episodes from worlds that contain them, whereas the reverse-causation (T2) protein was nominated in half of such Haiku 4.5 and GPT-OSS-20B episodes. The harmful surrogate (T8) was nominated most often by the leading agents, in 44 of 60 Opus 5 and 25 of 60 GPT-5.6 Sol episodes. However, these agents usually labeled it as harmful to outcomes (30 of 44 and 16 of 25 nominations, respectively), which avoided the safety penalty although each nomination still counted as a false positive. By contrast, every surrogate nomination by Haiku 4.5 and GPT-OSS-20B claimed it was beneficial and therefore incurred the penalty (Figure 5b).

### 4.4 The leading agents tested pairs that shared a genetic instrument, but no agent tested a delayed-effect pair, and the expanded budget did not detectably change their scores

Where observational data could not separate two candidates because they shared a genetic instrument, the leading agents tested the causal member experimentally in 14 of 24 eligible episodes. However, no experiment targeted a delayed-effect pair, and the expanded ($2 million) budget did not detectably change the scores of the leading agents.

Experimental access bore on the third judgment, whether a claim was sufficiently supported, because it allowed agents to test directly the candidates and uncertainties that the observational analyses had left unresolved. Of the 360 episodes with an experimental budget, 143 received at least one experiment, and 413 experiments were delivered in total, most of them in vivo knockdowns (Supplementary Table S9 and Supplementary Note S3). Experimental use was concentrated in the three highest-scoring agents. Opus 5, GPT-5.6 Sol, and Sonnet 5 accounted for 316 of the deliveries (76.5%) but represented only one third of the budget-eligible episodes.

The targets of these experiments show which uncertainties agents chose to resolve. Of the 413 experiments, 171 (41.4%) targeted a causal driver, 52 (12.6%) the T8 harmful surrogate, 49 (11.9%) the reverse-causation (T2) protein, one each the confounding (T1) and selection (T3) proteins, and 139 (33.7%) other proteins (Figure 5c). Twenty-five experiments (6.1%) targeted a member of an ambiguous pair, all in the six worlds with a T4 pair, whose members share a genetic instrument. In these six worlds, Opus 5 and GPT-5.6 Sol tested the causal member in 14 of 24 eligible episodes (six worlds, two agents, and two budget conditions). Moreover, every causal driver that these two agents tested and nominated carried the intervention direction implied by its knockdown. In contrast, no experiment targeted either member of the 15 delayed-effect pairs, in which the causal member acts mainly over follow-up and which a cell perturbation cannot detect, and only one of 405 eligible episodes nominated such a causal driver.

When results were returned, agents usually read them. Among the 143 episodes that received at least one experiment, later code referred back to the returned result in 124 (86.7%), and 84 (58.7%) changed their conclusion. Of the 19 episodes that did not refer back, 16 had requested the experiment in the same turn as their final submission, so the episode ended before the result arrived, even though the instructions had invited revisions of a provisional submission (Section 3.3). Reading a result did not, however, guarantee that it was used correctly. Smaller models (Haiku 4.5, Qwen3-8B, and Devstral Small) nominated a total of 25 proteins whose knockdown had shown no effect. In addition, agents misread the format of the returned results in 16 of these 143 episodes, most of them from smaller models (Supplementary Note S3).

Budget effects were heterogeneous and not uniformly positive. In the 17 worlds containing an unidentifiable pair, observational-only totals are capped at 92.5. This cap, however, lies far above the mean score of every agent. Pooled mean scores were 13.11, 12.77, and 16.28 under the observational-only, limited, and expanded budgets (Figure 4b). Relative to the observational-only budget, the expanded budget increased the mean score of Opus 5 by 0.13 points and that of GPT-5.6 Sol by 0.07 points, with intervals spanning both gains and losses, whereas larger gains were observed in lower-ranked models (Supplementary Table S6). However, GPT-OSS-20B and Qwen3-Coder-30B-A3B received only one and four experiments, respectively, so their gains cannot be attributed to experimental evidence. For the leading agents, in turn, the small change did not reflect unused funds. Under the expanded budget, Opus 5 and GPT-5.6 Sol bought experiments in 19 of 20 episodes each and spent 93% and 95% of the available funds, and about half of these experiments targeted a causal driver. Their recall of causal drivers nevertheless stayed close to its observational-only level, at 0.65 and 0.60 compared with 0.64 for both.

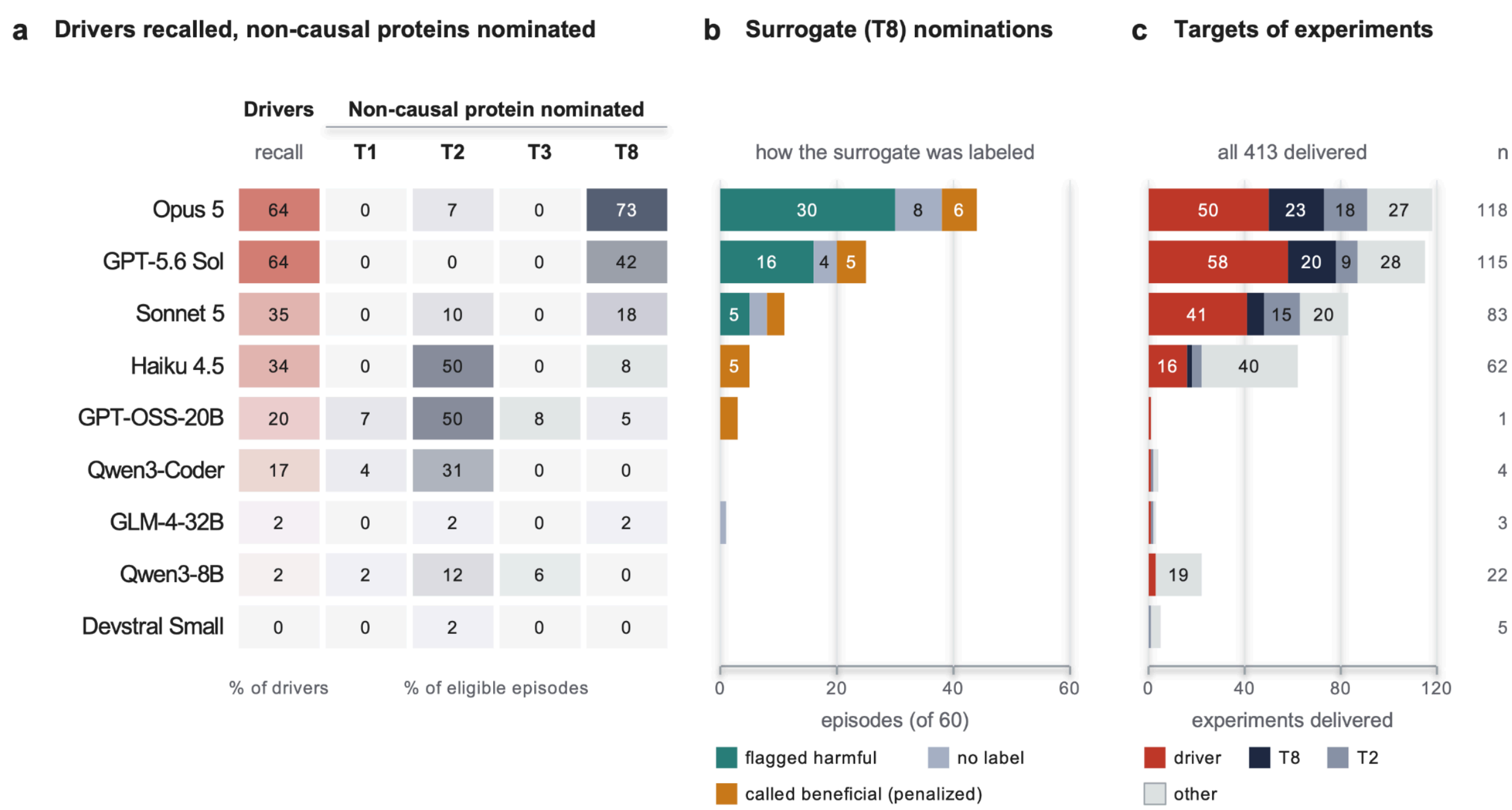

**Figure 5 | What agents did with the planted biases. a,** Mean driver recall and the percentage of episodes nominating the protein planted by each bias (non-causal protein or surrogate), among episodes in worlds carrying that bias (T1, 18 worlds; T2, 14; T3, 12; T8, 20; three episodes per model in each world). Biases for which the run export does not identify a planted protein are not shown. **b,** Episodes in which the T8 harmful surrogate was nominated, by the outcome-alignment label the agent assigned: flagged harmful (misaligned), no label, or claimed beneficial (aligned, which triggers the −30 safety penalty). **c,** Targets of the 413 delivered experiments: true drivers, the T8 surrogate, the reverse-causation (T2) protein, or other proteins (including one T1 and one T3 non-causal protein); n, experiments per model.

### 4.5 Failures arose at several workflow steps, and correct answers did not always reflect supporting analysis

Agents failed at several steps of the research workflow, from phenotype construction to evidence use and code execution, and the three 75-point episodes that recovered the exact set of causal drivers differed widely in the reasoning behind their answers.

The first failure occurred in phenotype construction, where a zero score can conflate several distinct problems. Phenotype construction received zero credit in 416 of 540 episodes (77.0%, Figure 6a). In 154 of them, the agent did not produce a phenotype at the required submission path, and in the other 262 it produced a phenotype that earned no credit, which can reflect validity failures, low correlation with latent disease severity, or failure to exceed the recorded baseline (Supplementary Note S3). However, causal confidence and bias identification were zero more often, and phenotype construction was not the largest source of lost points (Figure 4c).

A second failure was that agents generated evidence but did not use it. In the qualitative trajectory audit, some agents executed an analysis but neither cited its result in the final rationale nor revised the candidate set. For example, Opus 5 and GPT-5.6 Sol accessed genotypes in all 60 episodes (Figure 6b), but IV analyses were detected in only 41 and 29 of them, respectively (Section 4.3). Similarly, GLM-4-32B accessed genotypes in 20 episodes without any detected IV analysis. This failure mode was more frequent in smaller models. Haiku 4.5, for instance, purchased experiments in 18 of its 40 eligible episodes, and its code referred back to a returned result in 11 of those 18.

A third failure, particularly in smaller models, occurred in code execution. Several smaller models generated code that referenced nonexistent column names. Devstral Small reached the 30-turn limit in 53 of 60 episodes, repeatedly reloading and re-deriving information rather than maintaining its progress across turns (Figure 6d, Supplementary Note S3). These behaviors consumed turns that could otherwise have been used for scientific analysis.

Finally, correct final answers did not always reflect supporting analysis. Three selected episodes scored 75 and recovered the exact planted set of causal drivers. In the Haiku 4.5 episode, the final submission cited knockdown responses as evidence. By contrast, GPT-OSS-20B and Qwen3-Coder-30B-A3B recovered the same set after correlation-based screening, and both hard-coded "aligned" outcome-alignment labels, while GPT-OSS-20B also hard-coded intervention directions. Nevertheless, all three received full causal-confidence and intervention-direction credit, because the score evaluates final claims rather than the analyses behind them. More generally, across the 540 episodes, the number of core workflow milestones reached (of seven) correlated with the score (Spearman $\rho = 0.649$; Figure 6c), but this relationship is confounded by model identity and is partly mechanical.

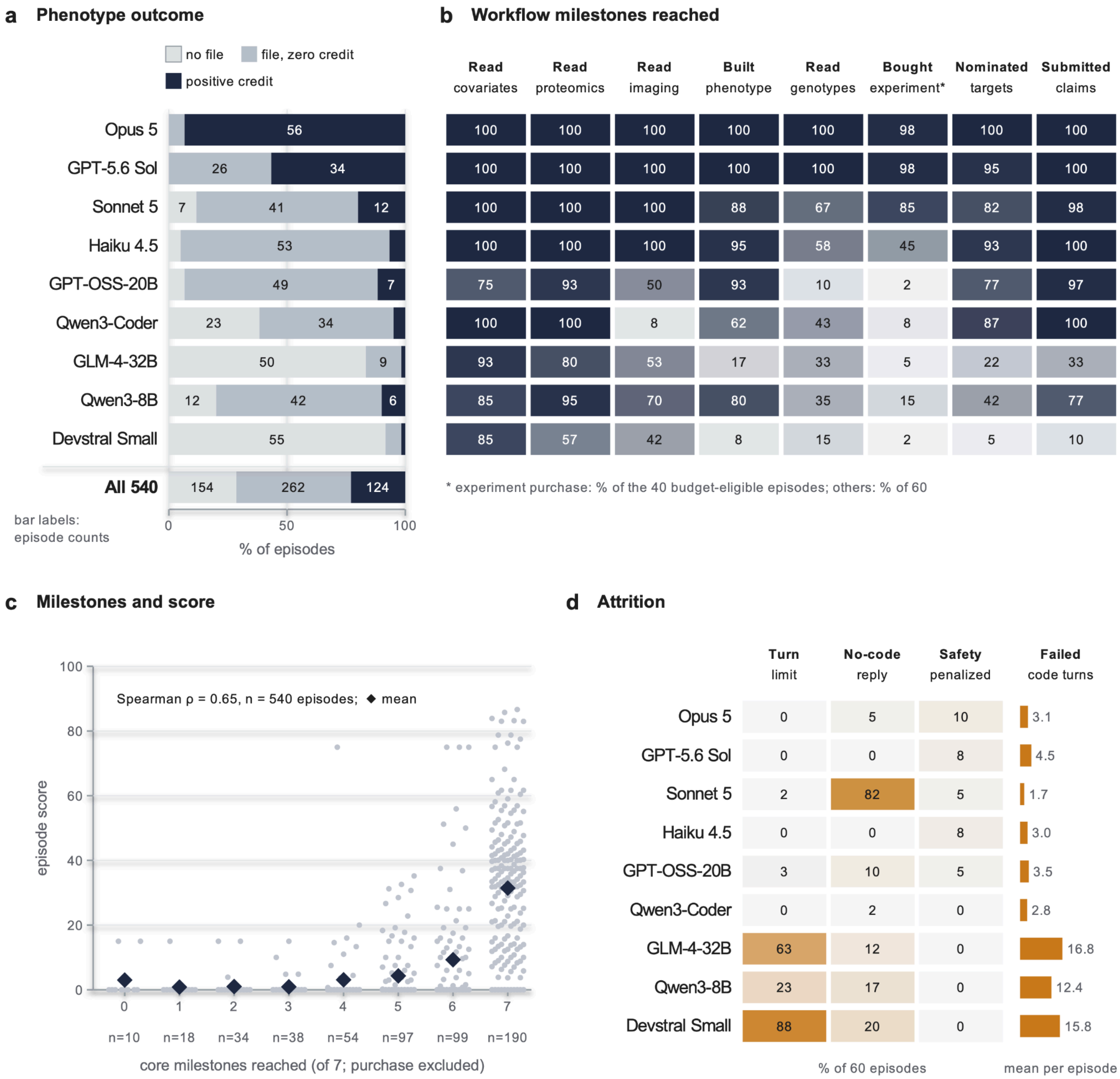

|  | Read covariates | Read proteomics | Read imaging | Built phenotype | Read genotypes | Bought experiment* | Nominated targets | Submitted claims |
|---|---|---|---|---|---|---|---|---|
| Opus 5 | 100 | 100 | 100 | 100 | 100 | 98 | 100 | 100 |
| GPT-5.6 Sol | 100 | 100 | 100 | 100 | 100 | 98 | 95 | 100 |
| Sonnet 5 | 100 | 100 | 100 | 88 | 67 | 85 | 82 | 98 |
| Haiku 4.5 | 100 | 100 | 100 | 95 | 58 | 45 | 93 | 100 |
| GPT-OSS-20B | 75 | 93 | 50 | 93 | 10 | 2 | 77 | 97 |
| Qwen3-Coder | 100 | 100 | 8 | 62 | 43 | 8 | 87 | 100 |
| GLM-4-32B | 93 | 80 | 53 | 17 | 33 | 5 | 22 | 33 |
| Qwen3-8B | 85 | 95 | 70 | 80 | 35 | 15 | 42 | 77 |
| Devstral Small | 85 | 57 | 42 | 8 | 15 | 2 | 5 | 10 |

|  | No file | File, zero credit | Positive credit |
|---|---|---|---|
| Opus 5 | | | 56 |
| GPT-5.6 Sol | | 26 | 34 |
| Sonnet 5 | 7 | 41 | 12 |
| Haiku 4.5 | | 53 | |
| GPT-OSS-20B | | 49 | 7 |
| Qwen3-Coder | 23 | 34 | |
| GLM-4-32B | 50 | 9 | |
| Qwen3-8B | 12 | 42 | 6 |
| Devstral Small | 55 | | |
| All 540 | 154 | 262 | 124 |

|  | Turn limit | No-code reply | Safety penalized | Failed code turns |
|---|---|---|---|---|
| Opus 5 | 0 | 5 | 10 | 3.1 |
| GPT-5.6 Sol | 0 | 0 | 8 | 4.5 |
| Sonnet 5 | 2 | 82 | 5 | 1.7 |
| Haiku 4.5 | 0 | 0 | 8 | 3.0 |
| GPT-OSS-20B | 3 | 10 | 5 | 3.5 |
| Qwen3-Coder | 0 | 2 | 0 | 2.8 |
| GLM-4-32B | 63 | 12 | 0 | 16.8 |
| Qwen3-8B | 23 | 17 | 0 | 12.4 |
| Devstral Small | 88 | 20 | 0 | 15.8 |

Figure 6 | Research workflow and failure modes of autonomous agents. a, Phenotype outcome per episode: no file at the required submission path, a file that received zero credit, or positive credit; bar labels are episode counts. b, Percentage of episodes reaching each harness-recorded workflow milestone; experiment purchase uses the 40 budget-eligible episodes per model, other milestones 60. c, Episode scores (dots) and means (diamonds) by the number of seven core milestones reached (purchase excluded); Spearman ρ = 0.65 across 540 episodes. This pooled relationship is descriptive and confounded with model identity. d, Percentage of episodes ending at the 30-turn limit, containing at least one reply without code, or receiving the safety penalty; bars, mean failed code turns per episode. Milestones are mechanical proxies and do not verify correct analysis or use of evidence.

## 5. Discussion

Here we introduce a framework for generating synthetic biomedical worlds in which end-to-end drug target discovery can be evaluated against a known causal ground truth. Each world is generated from a hidden causal model that jointly specifies genetic variation, molecular measurements, latent disease severity, multimodal clinical phenotypes, longitudinal outcomes and responses to intervention. This not only produces a synthetic dataset, but also an environment in which the causal drivers of disease, the processes that generate misleading evidence and the effects of intervention are known by construction but concealed from the agent. By varying these underlying causal structures across worlds, the framework can generate progressively more difficult discovery problems while retaining an exact answer against which each stage of the research process can be evaluated.

Using this environment, we find that the leading agents executed the individual analyses of a biobank-based target study without a prescribed pipeline. The leading agents constructed disease phenotypes from raw cardiac images, tested causal claims with genetic instruments, revised their hypotheses as evidence accumulated, used limited budgets to conduct simulated follow-up experiments, and recovered nearly two-thirds of the causal drivers. However, they rarely integrated these analyses into a correct and well-supported conclusion, and their performance was limited less by the analyses themselves than by the judgments that connect them, namely how to measure the disease, which evidence separates a causal driver from a non-causal protein, and when a claim is sufficiently supported. Because each of these judgments is scored separately against a known causal structure, DrugTargetWorld can attribute an agent's errors to specific stages of its analysis rather than only to its final answer.

Existing benchmarks (Chen, 2025; Koch, 2026; Li & Ho, 2026; Majumder et al., 2024; Mitchener et al., 2025; Xi et al., 2025) typically specify the research question and therefore rarely test these judgments, and agentic systems such as Kosmos (Mitchener et al., 2025b), the Virtual Biotech, and Biomni cannot yet be scored against the causal structure of the data they analyze. Such systems could be evaluated in the same worlds, where every final claim is scored against a known causal structure and anonymized protein identifiers (Section 3.2) confine the evaluation to what they infer from the data.

The first judgment, how to measure the disease, most clearly separated the two leading agents. Opus 5 and GPT-5.6 Sol had similar recall of causal drivers (0.64), and without phenotype credit the gap between their total scores narrowed from 4.6 to 0.9 points. The strongest phenotypes combined several imaging features, some weighted by clinical outcomes, and almost all were checked against other disease signals in the dataset. In the traces of the leading agents, none was chosen for its number of protein associations (Section 4.2). This pattern is consistent with biobank studies in which phenotype construction shapes statistical power and genetic discovery (An et al., 2023).

This result raises the prior question of how disease should be measured from multimodal biobank data. DrugTargetWorld answers it by design, scoring agreement with a single latent disease severity, but clinical medicine has no such reference point. Disease categories have been defined by presentation, organ, and consensus rather than by mechanism (Loscalzo et al., 2007), as the ejection-fraction thresholds that divide heart failure illustrate (Bozkurt et al., 2021). Nor do statistical criteria settle the matter, because broader definitions can strengthen genetic associations while capturing less of the intended disease (Cai et al., 2020), and prognostic phenotypes may reflect severity or non-causal markers rather than mechanism (Riley et al., 2013). Our worlds reproduce this tension, since a phenotype tuned to protein associations could lean on reverse causation (T2), and one tuned to clinical events could partly import the harmful surrogate. In real biobanks, moreover, diseases coexist and vary between patients (Shah et al., 2015), and data-driven phenotypes can reveal heritable structure that clinical labels miss (Yun et al., 2024). A phenotype designed by an agent that departs from a reference definition is therefore not necessarily wrong. In version 1, by contrast, a phenotype unrelated to the single latent severity carries little causal-driver signal, so such a departure is penalized by design. Version 2 will address this limitation by incorporating coexisting disease phenotypes, some outside the cardiovascular domain.

The second judgment, which evidence separates a causal driver from a non-causal protein, was only partly met. The leading agents based most nominations on cis genetic instruments and checked instrument strength, partly resembling established target prioritization, which combines Mendelian randomization, colocalization, cross-omic concordance, and perturbation data (Henry et al., 2022; Rasooly et al., 2023, 2025; Sanderson et al., 2022). Because the generator assigns each protein a single cis instrument, however, estimates could not be compared across independent instruments. Moreover, after correction for multiple testing, no analysis family that agents used was associated with target recall or with incorrect nominations. Bias-identification scores stayed low because agents rarely rejected non-causal proteins other than the reverse-causation (T2) protein and diluted correct rejections with incorrect ones. The harmful surrogate exposed the same gap, because the leading agents nominated it in 44 and 25 of 60 episodes, usually labeling it harmful to outcomes without concluding that it was not a causal driver.

The third judgment, when a claim is sufficiently supported, rested largely on experiments, because in this environment a single knockdown settles the causal status of any protein other than the harmful surrogate, which made experiments most valuable for candidates that observational data could not resolve. The leading agents used experiments effectively when a pair shared a genetic instrument, testing the causal member in 14 of 24 eligible episodes, and every driver they tested and nominated carried the intervention direction implied by its knockdown. By contrast, pairs whose causal member acts mainly over follow-up were never tested. Consistent with this, the expanded budget did not detectably change the scores of the leading agents, and their remaining errors appear to lie mostly in candidates never identified. The exceptions were two GPT-5.6 Sol episodes under the $2 million budget that ended on an empty provisional submission one turn after delivery, before the agent had seen its knockdown results. Smaller models, on the other hand, often bought experiments too late to read them or nominated proteins whose knockdown had shown no effect (Section 4.4).

These failure modes suggest harness changes that controlled ablations can test. Scaffolds could guide phenotype construction and track evidence, while prompts could require each experimental result to update or close an open question and submission could wait for pending results. Freshly generated worlds, each with its own ground truth, can also supply a verifiable reward for training agents across the entire workflow, an approach that has already improved agents in other interactive environments (Liévin et al., 2026; Narayanan et al., 2024). An outcome-based score can, however, credit answers that lack supporting analysis (Section 4.5), so a training reward should be paired with trajectory annotation and exploit audits. Smaller open-weight models, which scored zero in 78% of episodes, will likely need denser intermediate rewards or a curriculum of easier worlds.

This study has several limitations. The worlds are synthetic, calibrated to TOPMed MESA data only for selected population, imaging, genetic, protein missing-data, and event-rate summaries (Supplementary Methods S1.1), and their mostly linear mechanisms, effect sizes, and biases were designed rather than estimated from real disease. This limitation is inherent to any synthetic world with a known ground truth, whereas the remaining ones concern this evaluation and can be addressed in future versions. Each model ran once per world and budget, leaving run-to-run variation unmeasured, and the two human-guided episodes were exploratory references rather than a matched baseline. The minimal harness may also underestimate what optimized agent systems would achieve. Because the score evaluates final claims, hard-coded labels can earn credit (Section 4.5). The environment can also be exploited through the knockdown, which returns exactly zero for every non-causal protein except the harmful surrogate, unlike any real experiment, and through a phenotype tuned to the reverse-causation (T2) protein, which would earn credit without causal reasoning. Moreover, the prompt told agents in every world, including the null world, that a small number of proteins drive disease, and invited revisions of provisional submissions that the harness treated as final (Supplementary Methods S2.4). Phenotype scores were computed on a random 20% of participants, so a phenotype built only from cardiac MRI was scored on about 2,060 to 2,280 imaged participants, less precisely than one defined for the whole cohort. Finally, a data-generation error rendered follow-up MRI with DCM morphology in the 14 non-DCM worlds, including 9 of the 15 worlds whose delayed effects appear mainly on follow-up imaging. Excluding affected reads did not change the order of the leading agents (Supplementary Table S11D), although all six episodes above 80 points fell in excluded cells.

These limitations define the next iterations of the benchmark. Repeated episodes for each agent, world, and budget will measure run-to-run variation, and repeated episodes by independent human investigators will provide a baseline. Several human experts, together with agents using literature-search tools such as Paperclip, will annotate trajectories, distinguishing workflows a biostatistician would recognize from less conventional ones. Because the reward depends on the outcome, unconventional or more efficient pipelines that recover the ground truth still earn credit, so the benchmark may also support method discovery. Such pipelines will become candidate methods once their novelty is established and their frozen versions succeed on unseen worlds and in real cohorts.

The ultimate aim is to transfer what agents learn in simulation to real biobanks, which requires greater realism and a bridge to real data. Future worlds will include more causal drivers per disease, nonlinear and interacting effects, coexisting cardiovascular and non-cardiovascular conditions, and subtypes not captured by current guidelines, and each will be audited for exploitable cues. Pairing each biobank with simulated single-cell expression data and perturbation screens will require agents, like investigators, to combine evidence across incomplete sources. Following the injection-and-recovery design of the Kepler planet survey (Christiansen et al., 2015), known protein-to-phenotype effects can then be implanted into real biobank data, with planted non-causal proteins and permuted copies as known nulls. Agents can also be asked to recover causal relationships in MESA and UK Biobank that are supported by experimental evidence published after their training cutoff.

In summary, current AI agents can perform most of the individual analyses in biobank-based drug target discovery but do not yet reliably make the judgments that turn them into well-supported causal claims. By making these judgments measurable against a known ground truth, DrugTargetWorld offers both a benchmark for this capability and a verifiable reward with which agents can be trained to acquire it.

## Author contributions

B.G. conceived the study. S.M. and B.G. developed the methodology and benchmark design. S.M., B.G., A.H., E.C., I.B., A.S., and F.C. led software development and implementation. S.M. led the experimental evaluation, formal analysis, and visualization. B.G., P.S., and E.A. provided scientific and methodological guidance. S.M. wrote the original manuscript, and P.S., B.V., S.R., R.X., J.O., D.K., and M.W. contributed to its writing and editing. All authors reviewed and edited the manuscript. B.G. supervised the work.

## Acknowledgements

We thank Bailey Bova and Brittney Tong of Anthropic for their support through Anthropic's AI for Science program.

MESA data were obtained through the NIH database of Genotypes and Phenotypes (dbGaP; phs000209.v13.p3 and phs001416.v4.p1). Whole genome sequencing (WGS) for the Trans-Omics in Precision Medicine (TOPMed) program was supported by the National Heart, Lung and Blood Institute (NHLBI). WGS for "NHLBI TOPMed: Multi-Ethnic Study of Atherosclerosis (MESA)" (phs001416.v4.p1) was performed at the Broad Institute of MIT and Harvard (3U54HG003067-13S1). Centralized read mapping and genotype calling, along with variant quality metrics and filtering were provided by the TOPMed Informatics Research Center (3R01HL-117626-02S1, contract HHSN268201800002I) (Proteomics HHSN268201600034I). Phenotype harmonization, data management, sample-identity QC, and general study coordination, were provided by the TOPMed Data Coordinating Center (3R01HL-120393; U01HL-120393; contract HHSN268201800001I). MESA and the MESA SHARe projects are conducted and supported by the National Heart, Lung, and Blood Institute (NHLBI) in collaboration with MESA investigators. Support for MESA is provided by contracts 75N92020D00001, HHSN268201500003I, N01-HC-95159, 75N92020D00005, N01-HC-95160, 75N92020D00002, N01-HC-95161, 75N92020D00003, N01-HC-95162, 75N92020D00006, N01-HC-95163, 75N92020D00004, N01-HC-95164, 75N92020D00007, N01-HC-95165, N01-HC-95166, N01-HC-95167, N01-HC-95168, N01-HC-95169, UL1-TR-000040, UL1-TR-001079, and UL1-TR-001420. The authors thank the MESA participants and the MESA investigators and staff for their valuable contributions. A full list of participating MESA investigators and institutions can be found at http://www.mesa-nhlbi.org.

## Competing interests

E.A. reports advisory roles with Pacific Biosciences; ownership interests in Personalis, Deepcell, Svexa, Candela, Parameter Health, Saturnus Bio, and Swift Bio; and serves as a non-executive director of AstraZeneca and Dexcom. The remaining authors declare no competing interests.

## Funding

This work was supported in part by Anthropic's AI for Science program, which provided computational resources and model usage credits. We also acknowledge the following financial support: American Heart Association Career Development Award (26CDA1589055, to B.G.).

## Ethics statement

DrugTargetWorld generates synthetic participant-level worlds. Calibration used cohort-level summary statistics computed from MESA data obtained through dbGaP (phs000209 and phs001416), together with published MESA reports, and ACDC anatomical templates for MRI rendering (Supplementary Methods S1.1). These were group-level statistics only: counts, means, standard deviations, proportions, correlations, distribution quantiles, and simple model summaries such as covariate $R^2$. No MESA participant rows, identifiers, genotypes, or assay values were redistributed, and no real participant was resampled into any world. The two human-guided reference episodes, each conducted by an author, involved no research participants. PROT_ and TRANS_ identifiers are generated from integer indices and have no real gene or protein mapping; nominations therefore make no claim about an actual drug molecule. The scorer applies a 30-point penalty when the planted harmful surrogate (a clinically adverse target) is nominated as outcome-aligned. Releasing evaluation worlds and ground truth creates a risk of future training contamination; the generator permits evaluation on newly sampled seeds, and the separation of development and evaluation worlds will be documented for each subsequent panel.

## AI use statement

The authors (S.M.) wrote the first draft of the manuscript. Speech to text software Wispr was used for direct dictation of the manuscript. Both ChatGPT (OpenAI) and Claude (Anthropic) were used for minor language and formatting assistance. Grammarly was used for minor editing and grammar. All authors reviewed and approved the manuscript and take full responsibility for its content. No other writing assistance was provided or paid for.

The nine AI agents generated the analysis code and, where produced, submission files in the 540 canonical episodes. Supplementary Notes S1 and S2 separately report the two human-guided episodes, in which an investigator directed a language model, and only the episode in Supplementary Note S2 followed the standard evaluation protocol. AI assistants were also used to help edit the manuscript, inspect source code and archived trajectories, and develop the audit and sensitivity scripts reported here. The inspected records do not provide a complete history of AI assistance during the original generator and harness development. AI-assisted static-code classifications are treated as measurements with possible error, rather than independent expert judgments. The authors retain responsibility for the manuscript and its scientific claims.

## Code and data availability

DrugTargetWorld code is publicly available at https://github.com/sammargolis/DrugTargetWorld/. The project website is available at https://DrugTargetWorld.vercel.app/. Benchmark assets and released datasets are available at https://huggingface.co/datasets/sammargolis/DrugTargetWorld-assets.

# Supplementary Figures:

**Figure S1 | Model performance across DrugTargetWorld.** Mean score across the three budget regimes for each of nine agent systems in each of 20 worlds. Worlds are labeled by cardiac archetype and design class. Scores range from 0 to 100, with higher scores indicating better benchmark performance.

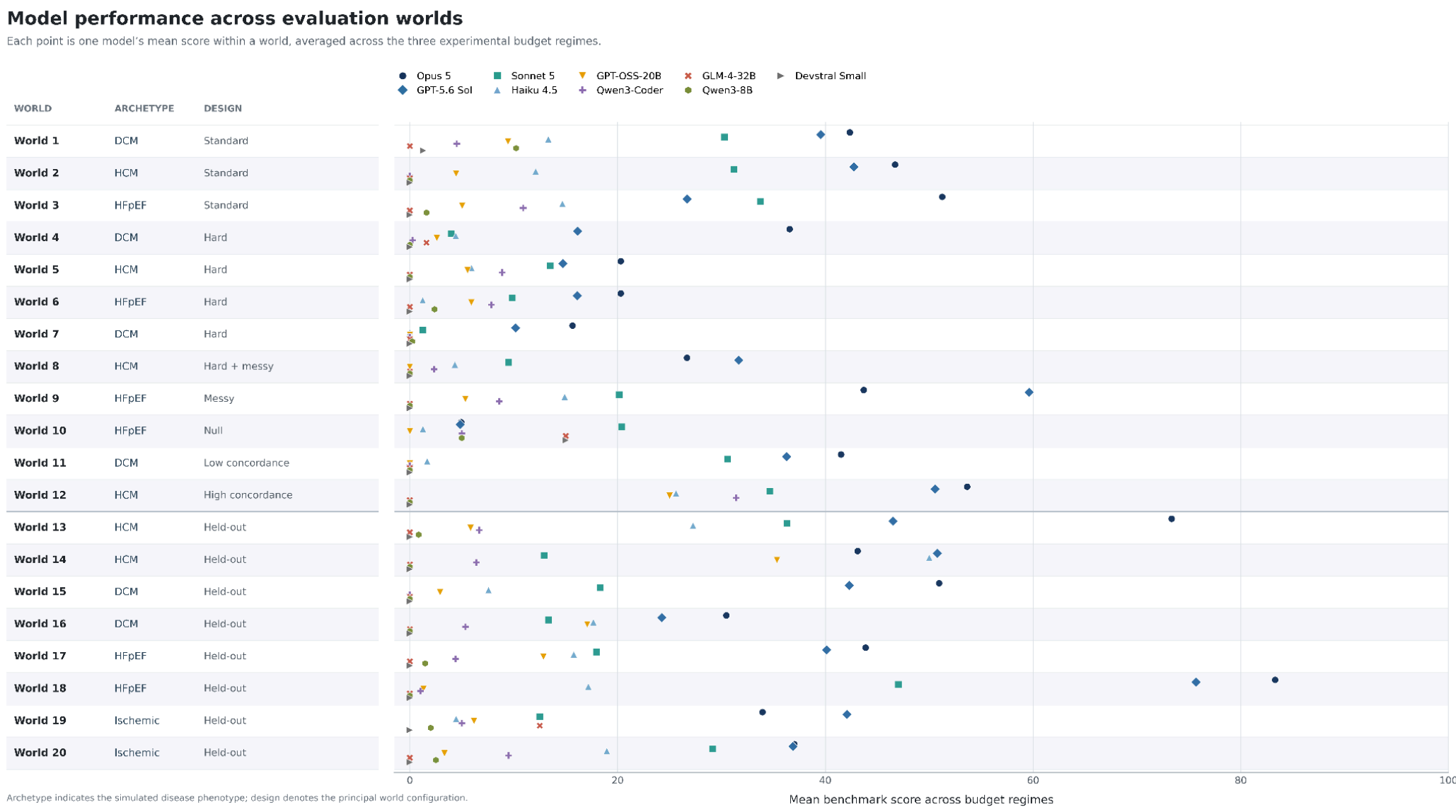

**Figure S2 | Cross-benchmark performance of evaluated models.** Mean DrugTargetWorld score compared with reported performance on Terminal-Bench and SWE-bench Pro for the same model families.

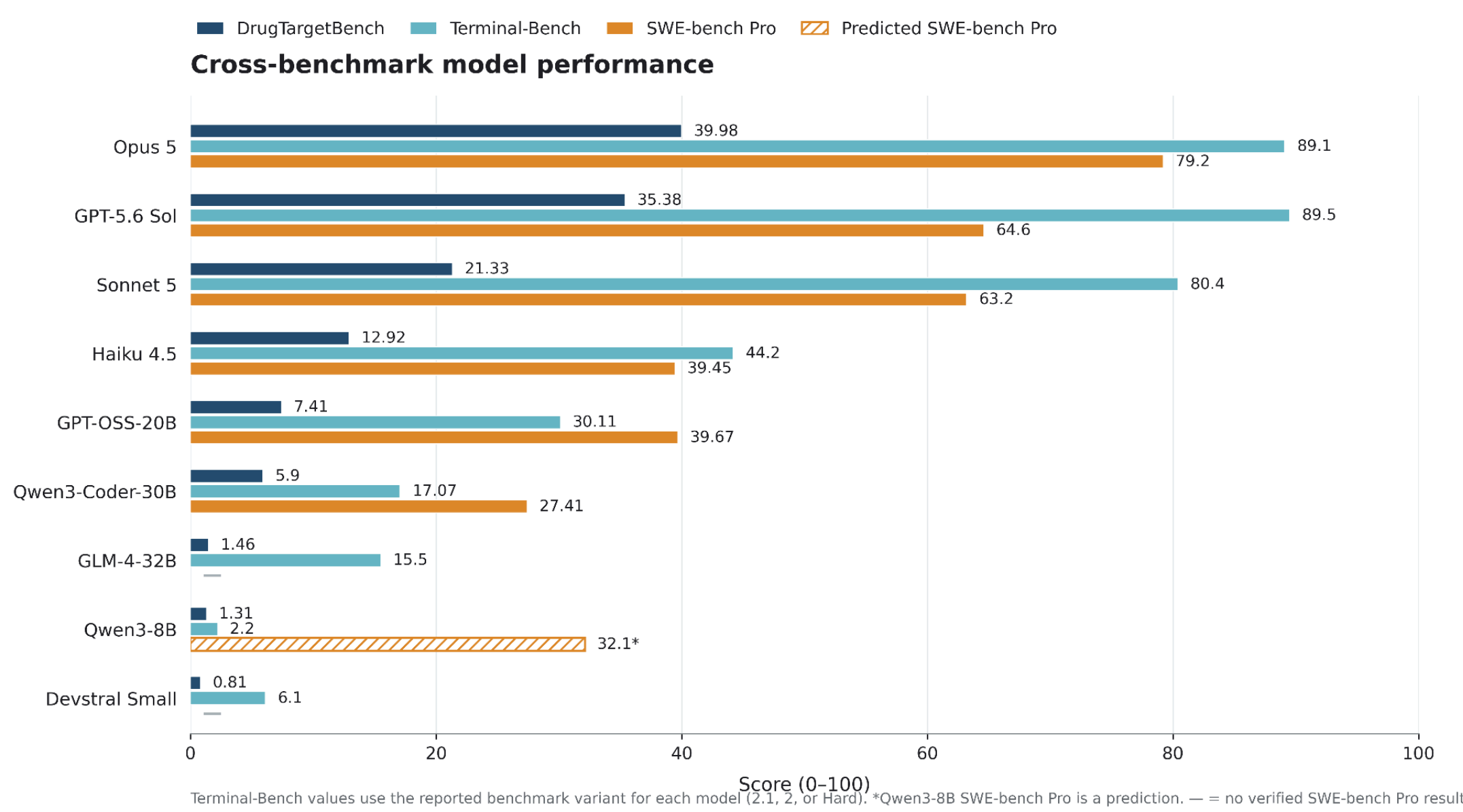

**Table S1 | Model roster and run provenance**

| Arm | Model identifier | Serving | Hardware | Dates tested (UTC) |
|---|---|---|---|---|
| opus-5 | claude-opus-5 | Anthropic API | vendor | 2026-09-07 18:37 to 2026-09-08 00:07 |
| gpt-5.6-sol | gpt-5.6-sol | OpenAI API | vendor | 2026-09-05 22:29 to 2026-09-06 01:34 |
| sonnet-5 | claude-sonnet-5 | Anthropic API | vendor | 2026-09-04 22:29 to 2026-09-05 00:59 |
| haiku-4-5 | claude-haiku-4-5 | Anthropic API | vendor | 2026-09-04 15:25 to 17:28 |
| gpt-oss-20b | openai/gpt-oss-20b | self-hosted vLLM 0.10.2 | 1 x H100 80GB | 2026-09-04 15:13 to 16:44 |
| qwen3-coder-30b | Qwen/Qwen3-Coder-30B-A3B-Instruct | self-hosted vLLM 0.10.2, TP=2 | 2 x H100 80GB | 2026-09-04 18:03 to 18:37 |
| glm-4-32b | zai-org/GLM-4-32B-0414 | self-hosted vLLM 0.10.2, TP=2 | 2 x H100 80GB | 2026-09-04 18:02 to 20:20 |
| qwen3-8b | Qwen/Qwen3-8B | self-hosted vLLM 0.10.2, TP=1 | 1 x H100 80GB | 2026-09-04 15:44 to 20:57 |
| devstral-small | mistralai/Devstral-Small-2507 | self-hosted vLLM 0.10.2, TP=2 | 2 x H100 80GB | 2026-09-04 17:59 to 20:22 |
| human:bruna | human player | cardioseek-play, harness ced47324 | n/a | 2026-09-07 (standard-03; observational only) |

**Supplementary Table S2 | Inference cost and token use**

| Arm | Provider | Episodes | Input tokens | Output tokens | API cost ($) | GPU-h | Imputed GPU cost ($) | Cost / episode ($) |
|---|---|---|---|---|---|---|---|---|
| opus-5 | Anthropic | 60 | 39,353,164 | 2,056,198 | 248.17 | - | - | 4.14 |
| gpt-5.6-sol | OpenAI | 60 | 26,602,045 | 739,149 | 121.19 | - | - | 2.02 |
| **sonnet-5** | **Anthropic** | **60** | **21,247,607** | **1,616,687** | **87.99** | **-** | **-** | **1.47** |
| **haiku-4-5** | **Anthropic** | **60** | **10,050,083** | **1,060,240** | **15.35** | **-** | **-** | **0.26** |
| **devstral-small** | **self-hosted** | **60** | **18,781,760** | **392,556** | **0** | **4.76** | **11.90** | **-** |
| **glm-4-32b** | **self-hosted** | **60** | **16,047,705** | **598,681** | **0** | **4.72** | **11.80** | **-** |
| **qwen3-8b** | **self-hosted** | **60** | **10,833,367** | **2,136,841** | **0** | **5.29** | **13.22** | **-** |
| **qwen3-coder-30b** | **self-hosted** | **60** | **3,249,958** | **469,101** | **0** | **1.28** | **3.20** | **-** |
| **gpt-oss-20b** | **self-hosted** | **60** | **3,086,488** | **468,774** | **0** | **1.56** | **3.90** | **-** |
| **Total** | **-** | **540** | **149,252,177** | **9,538,227** | **472.70** | **17.61** | **44.02** | **-** |

## Supplementary Table S3 | Evaluation-world design and phenotype-scoring baselines.

### Panel A. World design

| World ID | Split | Archetype | Generation suite | Hard | Messy | Null | Causal drivers |
|---|---|---|---|---|---|---|---|
| hard-01 | Development | DCM | cardioseek-v1-hard-a | Yes | No | No | 5 |
| hard-02 | Development | HCM | cardioseek-v1-hard-b | Yes | No | No | 3 |
| hard-03 | Development | HFpEF | cardioseek-v1-hard-c | Yes | No | No | 6 |
| hard-04 | Development | DCM | cardioseek-v1-hard-a | Yes | No | No | 4 |
| hard-messy-01 | Development | HCM | cardioseek-v1-hard-messy | Yes | Yes | No | 5 |
| heldout-dcm-01 | Held-out | DCM | cardioseek-v1-standard-a | No | No | No | 5 |
| heldout-dcm-02 | Held-out | DCM | cardioseek-v1-hard-messy | Yes | Yes | No | 5 |
| heldout-hcm-01 | Held-out | HCM | cardioseek-v1-standard-b | No | No | No | 2 |
| heldout-hcm-02 | Held-out | HCM | cardioseek-v1-hard-b | Yes | No | No | 2 |
| heldout-hfpef-01 | Held-out | HFpEF | cardioseek-v1-messy | No | Yes | No | 5 |
| heldout-hfpef-02 | Held-out | HFpEF | cardioseek-v1-standard-a | No | No | No | 2 |
| heldout-ischemic-01 | Held-out | Ischemic | cardioseek-v1-standard-b | No | No | No | 4 |
| heldout-ischemic-02 | Held-out | Ischemic | cardioseek-v1-hard-messy | Yes | Yes | No | 4 |
| high-concordance-01 | Development | HCM | cardioseek-v1-high-concordance | No | No | No | 2 |
| low-concordance-01 | Development | DCM | cardioseek-v1-low-concordance | No | No | No | 3 |
| messy-01 | Development | HFpEF | cardioseek-v1-messy | No | Yes | No | 5 |
| null-01 | Development | HFpEF | cardioseek-v1-null | No | No | Yes | 0 |
| standard-01 | Development | DCM | cardioseek-v1-standard-a | No | No | No | 3 |
| standard-02 | Development | HCM | cardioseek-v1-standard-b | No | No | No | 6 |
| standard-03 | Development | HFpEF | cardioseek-v1-standard-a | No | No | No | 4 |

**Panel B. Dimensions, causal non-identifiability, delayed effects, and phenotype baseline**

| World ID | Participants | MRI participants | Proteins | Variants | T4 causal non-identifiability | Delayed-effect setting | Phenotype baseline, b |
|---|---|---|---|---|---|---|---|
| hard-01 | 54,000 | 10,800 | 2,941 | 8,192 | Yes | Yes | 0.4010 |
| hard-02 | 54,000 | 10,800 | 2,941 | 8,192 | No | Yes | 0.6748 |
| hard-03 | 54,000 | 10,800 | 2,941 | 8,192 | No | Yes | 0.7473 |
| hard-04 | 54,000 | 10,800 | 2,941 | 8,192 | No | Yes | 0.4237 |
| hard-messy-01 | 54,000 | 10,800 | 2,941 | 8,192 | No | Yes | 0.6674 |
| heldout-dcm-01 | 54,000 | 10,800 | 2,941 | 8,192 | No | Yes | 0.4317 |
| heldout-dcm-02 | 54,000 | 10,800 | 2,941 | 8,192 | No | Yes | 0.4419 |
| heldout-hcm-01 | 54,000 | 10,800 | 2,941 | 8,192 | Yes | No | 0.6745 |
| heldout-hcm-02 | 54,000 | 10,800 | 2,941 | 8,192 | No | No | 0.6839 |
| heldout-hfpef-01 | 54,000 | 10,800 | 2,941 | 8,192 | No | Yes | 0.7527 |
| heldout-hfpef-02 | 54,000 | 10,800 | 2,941 | 8,192 | Yes | No | 0.7525 |
| heldout-ischemic-01 | 54,000 | 10,800 | 2,941 | 8,192 | No | Yes | 0.7172 |
| heldout-ischemic-02 | 54,000 | 10,800 | 2,941 | 8,192 | No | Yes | 0.7089 |
| high-concordance-01 | 54,000 | 10,800 | 2,941 | 8,192 | No | No | 0.6792 |
| low-concordance-01 | 54,000 | 10,800 | 2,941 | 8,192 | Yes | Yes | 0.4329 |
| messy-01 | 54,000 | 10,800 | 2,941 | 8,192 | Yes | Yes | 0.7465 |
| null-01 | 54,000 | 10,800 | 2,941 | 8,192 | No | No | 0.7505 |
| standard-01 | 54,000 | 10,800 | 2,941 | 8,192 | Yes | Yes | 0.4219 |
| standard-02 | 54,000 | 10,800 | 2,941 | 8,192 | No | Yes | 0.6743 |
| standard-03 | 54,000 | 10,800 | 2,941 | 8,192 | No | Yes | 0.7645 |

**Notes.** The fixed evaluation panel contained 20 worlds, each with 54,000 participants, 10,800 participants with MRI, 2,941 proteins, and 8,192 variants. T4 denotes the planted causal non-identifiability bias; in these worlds a causal driver and a noncausal alternative cannot be separated from the observational genetic evidence alone. Delayed-effect setting denotes worlds in which causal consequences emerge primarily over longer follow-up. The phenotype baseline b is the world-specific built-in absolute

correlation used in the phenotype scoring function described in Supplementary Methods S2.5. DCM, dilated cardiomyopathy; HCM, hypertrophic cardiomyopathy; HFpEF, heart failure with preserved ejection fraction; MRI, magnetic resonance imaging.

**Supplementary Table S4 | Benchmark performance by model.**

**Panel A. Total-score distribution**

| Model | Episodes | Mean | SD | Median | IQR | Score >0, n (%) | Score >=50, n (%) | Score >=80, n (%) |
|---|---|---|---|---|---|---|---|---|
| Opus 5 | 60 | 39.98 | 20.65 | 41.21 | 29.40-49.74 | 56 (93.3%) | 15 (25.0%) | 2 (3.3%) |
| GPT-5.6 Sol | 60 | 35.38 | 20.65 | 36.59 | 20.00-48.66 | 55 (91.7%) | 15 (25.0%) | 2 (3.3%) |
| Sonnet 5 | 60 | 21.33 | 22.01 | 12.50 | 0.00-36.08 | 43 (71.7%) | 6 (10.0%) | 2 (3.3%) |
| Haiku 4.5 | 60 | 12.92 | 17.01 | 8.30 | 0.00-19.63 | 38 (63.3%) | 3 (5.0%) | 0 (0.0%) |
| GPT-OSS-20B | 60 | 7.41 | 15.33 | 0.00 | 0.00-10.00 | 24 (40.0%) | 2 (3.3%) | 0 (0.0%) |
| Qwen3-Coder-30B-A3B | 60 | 5.90 | 12.30 | 0.00 | 0.00-7.50 | 23 (38.3%) | 1 (1.7%) | 0 (0.0%) |
| GLM-4-32B | 60 | 1.46 | 4.79 | 0.00 | 0.00-0.00 | 6 (10.0%) | 0 (0.0%) | 0 (0.0%) |
| Qwen3-8B | 60 | 1.31 | 4.62 | 0.00 | 0.00-0.00 | 9 (15.0%) | 0 (0.0%) | 0 (0.0%) |
| Devstral Small | 60 | 0.81 | 3.32 | 0.00 | 0.00-0.00 | 4 (6.7%) | 0 (0.0%) | 0 (0.0%) |

**Panel B. Target recovery and scoring components**

| Model | Target precision | Target recall | Target identification /30 | Causal confidence /25 | Bias identification /15 | Direction /20 | Phenotype /10 | Safety penalties, n |
|---|---|---|---|---|---|---|---|---|
| Opus 5 | 0.676 | 0.636 | 14.80 | 4.26 | 2.87 | 13.05 | 7.19 | 6 |
| GPT-5.6 Sol | 0.699 | 0.639 | 15.81 | 4.19 | 1.02 | 13.00 | 3.16 | 5 |
| Sonnet 5 | 0.600 | 0.348 | 10.24 | 1.98 | 1.34 | 6.96 | 1.16 | 3 |
| Haiku 4.5 | 0.304 | 0.339 | 5.27 | 1.61 | 0.44 | 5.91 | 0.21 | 5 |
| GPT-OSS-20B | 0.167 | 0.200 | 2.94 | 0.92 | 0.00 | 3.43 | 0.68 | 3 |
| Qwen3-Coder-30B-A3B | 0.193 | 0.170 | 3.26 | 0.56 | 0.00 | 1.94 | 0.14 | 0 |
| GLM-4-32B | 0.025 | 0.017 | 1.12 | 0.00 | 0.00 | 0.25 | 0.08 | 0 |
| Qwen3-8B | 0.034 | 0.024 | 0.70 | 0.00 | 0.00 | 0.22 | 0.39 | 0 |
| Devstral Small | 0.000 | 0.000 | 0.75 | 0.00 | 0.00 | 0.00 | 0.06 | 0 |

**Notes.** Each model was evaluated in 60 episodes: one episode in each of 20 worlds under each of three experimental-budget conditions. Scores are reported on the 0-100 benchmark scale. SD describes variation across worlds and budget conditions, not run-to-run replicate variability. IQR is the first to third

quartile. Target precision and recall are episode-level means. Safety penalties are the number of episodes in which the 30-point harmful-surrogate penalty was applied.

**Supplementary Table S5 | Pairwise performance comparisons among API-served models.**

| Model A | Model B | Mean A | Mean B | Mean difference (A-B) | 95% CI | Paired-t P | Worlds A>B / A<B / tied | Exact sign-test P |
|---|---|---|---|---|---|---|---|---|
| Opus 5 | GPT-5.6 Sol | 39.98 | 35.38 | 4.60 | -0.26 to 9.45 | 0.062 | 16/4/0 | 0.012 |
| Opus 5 | Sonnet 5 | 39.98 | 21.33 | 18.65 | 12.88 to 24.41 | <0.001 | 19/1/0 | <0.001 |
| Opus 5 | Haiku 4.5 | 39.98 | 12.92 | 27.06 | 19.57 to 34.55 | <0.001 | 19/1/0 | <0.001 |
| GPT-5.6 Sol | Sonnet 5 | 35.38 | 21.33 | 14.05 | 7.57 to 20.53 | <0.001 | 18/2/0 | <0.001 |
| GPT-5.6 Sol | Haiku 4.5 | 35.38 | 12.92 | 22.47 | 15.54 to 29.40 | <0.001 | 20/0/0 | <0.001 |
| Sonnet 5 | Haiku 4.5 | 21.33 | 12.92 | 8.41 | 1.86 to 14.97 | 0.015 | 17/3/0 | 0.003 |

**Notes.** For each model and world, scores were first averaged across the three budget conditions. Pairwise differences were then calculated across the 20 worlds. Confidence intervals are t-based 95% confidence intervals for the paired world-level differences. Paired-t P values are exploratory two-sided tests. Exact sign tests are two-sided binomial tests with tied worlds excluded. P values are not adjusted for multiplicity.

**Supplementary Table S6 | Performance by experimental-budget condition.**

| Model | Observational mean | Limited ($450k) mean | Expanded ($2M) mean | Limited - observational | Expanded - observational | 95% CI for expanded contrast | Paired-t P | Exact sign-test P |
|---|---|---|---|---|---|---|---|---|
| Opus 5 | 41.10 | 37.61 | 41.23 | -3.49 | 0.13 | -8.71 to 8.98 | 0.975 | 0.503 |
| GPT-5.6 Sol | 35.72 | 34.64 | 35.79 | -1.09 | 0.07 | -10.08 to 10.22 | 0.989 | 0.115 |
| Sonnet 5 | 22.05 | 10.02 | 31.91 | -12.03 | 9.86 | -2.40 to 22.12 | 0.109 | 0.238 |
| Haiku 4.5 | 13.81 | 12.24 | 12.70 | -1.56 | -1.11 | -7.69 to 5.47 | 0.728 | 0.143 |
| GPT-OSS-20B | 0.92 | 9.65 | 11.65 | 8.73 | 10.74 | 1.95 to 19.53 | 0.019 | 0.003 |
| Qwen3-Coder-30B-A3B | 1.20 | 7.12 | 9.38 | 5.92 | 8.19 | 3.02 to 13.35 | 0.004 | 0.092 |
| GLM-4-32B | 2.00 | 0.99 | 1.38 | -1.01 | -0.62 | -1.93 to 0.68 | 0.330 | 1.000 |
| Qwen3-8B | 0.46 | 1.91 | 1.57 | 1.45 | 1.11 | -0.86 to 3.08 | 0.254 | 0.453 |
| Devstral Small | 0.75 | 0.75 | 0.94 | 0.00 | 0.19 | -0.21 to 0.60 | 0.330 | 1.000 |

**Notes.** Budget contrasts were paired by world (n = 20 worlds per model). The limited condition provided a $450,000 experimental budget and the expanded condition a $2,000,000 budget; observational episodes had no experimental budget. The confidence interval and P values shown are for the expanded-minus-observational contrast. Experimental budgets and costs are simulated benchmark resource constraints.

**Supplementary Table S7 | Phenotype construction and internal validation.**

**Panel A. Phenotype performance by model**

| Model | Phenotype at scored path, n | Scorable, n | Positive credit, n | Full credit, n | Mean phenotype score /10 | Mean \|r\| | SD \|r\| | Internally validated, n |
|---|---|---|---|---|---|---|---|---|
| Opus 5 | 60 | 60 | 56 | 17 | 7.19 | 0.785 | 0.105 | 54 |
| GPT-5.6 Sol | 60 | 60 | 34 | 3 | 3.16 | 0.637 | 0.191 | 40 |
| Sonnet 5 | 53 | 53 | 12 | 0 | 1.16 | 0.502 | 0.235 | 44 |
| Haiku 4.5 | 57 | 42 | 4 | 0 | 0.21 | 0.352 | 0.221 | 4 |
| GPT-OSS-20B | 56 | 47 | 7 | 1 | 0.68 | 0.372 | 0.241 | 0 |
| Qwen3-Coder-30B-A3B | 37 | 37 | 3 | 0 | 0.14 | 0.376 | 0.159 | 0 |
| GLM-4-32B | 10 | 10 | 1 | 0 | 0.08 | 0.274 | 0.251 | 0 |
| Qwen3-8B | 48 | 35 | 6 | 0 | 0.39 | 0.395 | 0.206 | 0 |
| Devstral Small | 5 | 5 | 1 | 0 | 0.06 | 0.368 | 0.258 | 0 |

**Panel B. Performance by phenotype-construction strategy**

| Construction strategy | Scorable, n | Mean \|r\| | SD \|r\| | Median \|r\| | IQR | Mean phenotype score /10 |
|---|---|---|---|---|---|---|
| Weighted composite | 176 | 0.528 | 0.242 | 0.575 | 0.331-0.726 | 2.23 |
| Single imaging measure | 95 | 0.360 | 0.218 | 0.318 | 0.274-0.533 | 0.48 |
| Outcome-supervised | 42 | 0.751 | 0.110 | 0.771 | 0.637-0.838 | 6.10 |
| Other | 14 | 0.295 | 0.236 | 0.288 | 0.111-0.327 | 0.71 |
| PCA/SVD | 11 | 0.554 | 0.242 | 0.630 | 0.397-0.720 | 2.21 |
| Covariate-residualized imaging | 6 | 0.624 | 0.296 | 0.728 | 0.573-0.824 | 3.21 |
| Factor model | 4 | 0.844 | 0.024 | 0.848 | 0.832-0.859 | 9.16 |
| None | 1 | 0.483 | - | 0.483 | 0.483-0.483 | 0.00 |

**Panel C. Association of internal validation with phenotype performance**

| Validated before submission | Scorable, n | Mean \|r\| | SD \|r\| | Median \|r\| | Mean phenotype score /10 |
|---|---|---|---|---|---|
| Yes | 139 | 0.663 | 0.211 | 0.705 | 4.55 |
| No | 210 | 0.402 | 0.228 | 0.396 | 0.73 |

**Panel D. Clinical or imaging endpoints used for internal validation**

| Validation endpoint | Episodes |
|---|---|
| Mortality | 104 |
| MACE | 97 |
| Coronary CT calcium or stenosis | 99 |
| ECG features | 76 |
| EHR diagnoses | 26 |
| Repeat imaging visit | 9 |

**Notes.** Scorable phenotypes are submissions for which a phenotype-latent-disease correlation could be evaluated. |r| is the absolute correlation between the submitted phenotype and the hidden latent disease severity. Positive credit indicates a phenotype score >0; full credit indicates a phenotype score of 10. Strategy labels are trajectory annotations of the submitted phenotype construction. Panel C is restricted to the 349 episodes with scorable phenotype correlations. Panel D counts validation checks across all annotated episodes that performed internal validation; an episode could use more than one endpoint, so counts are not mutually exclusive. CT, computed tomography; ECG, electrocardiogram; EHR, electronic health record; MACE, major adverse cardiovascular events; PCA, principal component analysis; SVD, singular value decomposition.

**Supplementary Table S8 | Research strategy and process annotations by model.**

**Panel A. Causal-analysis strategy**

| Model | Cis MR primary route, n/60 | Used genetic instruments, n/60 | Checked instrument strength, n/60 | Checked exclusion restriction, n/60 | Mean hypothesis revisions | Evidence-driven revision, n/60 |
|---|---|---|---|---|---|---|
| Opus 5 | 57 | 60 | 59 | 33 | 2.05 | 60 |
| GPT-5.6 Sol | 50 | 59 | 57 | 37 | 1.60 | 59 |
| Sonnet 5 | 23 | 29 | 22 | 1 | 1.15 | 56 |
| Haiku 4.5 | 0 | 20 | 3 | 0 | 1.10 | 49 |
| GPT-OSS-20B | 1 | 7 | 0 | 0 | 0.10 | 0 |
| Qwen3-Coder-30B-A3B | 0 | 1 | 0 | 0 | 0.15 | 2 |
| GLM-4-32B | 0 | 0 | 0 | 0 | 0.02 | 2 |
| Qwen3-8B | 0 | 8 | 0 | 0 | 0.20 | 0 |
| Devstral Small | 0 | 1 | 0 | 0 | 0.00 | 0 |

**Panel B. Experimental planning and process**

| Model | Budget shaped plan, n/60 | Experiment planned before purchase, n/60 | Recognized unresolvable pair, n/60 | Read data dictionary, n/60 | Ended deliberately, n/60 | Ended with analysis unfinished (trace annotation), n/60 |
|---|---|---|---|---|---|---|
| Opus 5 | 1 | 9 | 56 | 60 | 14 | 3 |
| GPT-5.6 Sol | 10 | 30 | 5 | 60 | 47 | 1 |
| Sonnet 5 | 7 | 6 | 51 | 60 | 6 | 4 |
| Haiku 4.5 | 16 | 17 | 14 | 43 | 46 | 2 |
| GPT-OSS-20B | 0 | 0 | 2 | 10 | 19 | 20 |
| Qwen3-Coder-30B-A3B | 1 | 0 | 0 | 11 | 28 | 5 |
| GLM-4-32B | 0 | 0 | 8 | 35 | 2 | 47 |
| Qwen3-8B | 2 | 0 | 13 | 14 | 9 | 24 |
| Devstral Small | 0 | 0 | 0 | 33 | 1 | 54 |

**Notes.** Counts derive from the trajectory strategy-annotation archive and describe recorded research behavior; they do not establish that an analysis was statistically valid or correctly interpreted. "Cis MR primary route" denotes the annotated strategy supporting final target prioritization and is distinct from the stricter operation-based MR/IV calculation audit reported in the main text. Actual harness turn-limit terminations are reported in Table S10. MR, Mendelian randomization; IV, instrumental variable.

**Supplementary Table S9 | Experimental selection and use.**

**Panel A. Experimental use by model**

| Model | Eligible episodes | Requested >=1, n | Received >=1, n | Cell perturbations delivered | Knockdowns delivered | Refused requests | Total simulated spend |
|---|---|---|---|---|---|---|---|
| Opus 5 | 40 | 39 | 39 | 8 | 110 | 0 | $45,200,000 |
| GPT-5.6 Sol | 40 | 39 | 39 | 0 | 115 | 0 | $46,000,000 |
| Sonnet 5 | 40 | 34 | 34 | 5 | 78 | 2 | $31,950,000 |
| Haiku 4.5 | 40 | 18 | 18 | 12 | 50 | 4 | $21,800,000 |
| GPT-OSS-20B | 40 | 1 | 1 | 0 | 1 | 0 | $400,000 |
| Qwen3-Coder-30B-A3B | 40 | 3 | 3 | 3 | 1 | 2 | $850,000 |
| GLM-4-32B | 40 | 2 | 2 | 2 | 1 | 0 | $700,000 |
| Qwen3-8B | 40 | 8 | 6 | 6 | 16 | 26 | $7,300,000 |
| Devstral Small | 40 | 1 | 1 | 0 | 5 | 5 | $2,000,000 |

**Panel B. Targets of delivered experiments**

| Target category | Experiments | Percent of delivered |
|---|---|---|
| True causal driver | 171 | 41.4% |
| T8 harmful surrogate | 52 | 12.6% |
| Reverse-causation (T2) protein | 49 | 11.9% |
| Confounding (T1) protein | 1 | 0.2% |
| Selection/collider (T3) protein | 1 | 0.2% |
| Other protein | 139 | 33.7% |

**Panel C. Use of returned experimental results**

| Group | Purchasing episodes | Referenced returned result in code | Percent referenced | Changed conclusion, n (%) |
|---|---|---|---|---|
| All purchasing episodes | 143 | 124 | 86.7% | 84 (58.7%) |
| Opus 5 | 39 | 37 | 94.34% | - |
| GPT-5.6 Sol | 39 | 37 | 94.34% | - |
| Sonnet 5 | 34 | 34 | 100% | - |
| Haiku 4.5 | 18 | 11 | 61.1% | - |

**Notes.** There were 360 budget-eligible episodes. Across these episodes, 145 requested an experiment and 143 received at least one; 413 experiments were delivered, comprising 36 cell perturbations and 377 in vivo knockdowns. Thirty-nine requests were refused in budget-eligible episodes. Eleven additional requests were made and refused in observational episodes, yielding 50 refused requests overall. Total spend is the sum of simulated experimental charges across a model's 40 budget-eligible episodes. Panel C uses the source-linked code audit for returned-result reference; the changed-conclusion count is reported for all purchasing episodes. Target categories are mutually exclusive.

**Supplementary Table S10 | Workflow failures and execution attrition by model.**

**Panel A. Phenotype and execution outcomes**

| Model | No phenotype at scored path, n | Phenotype present but zero credit, n | Positive phenotype credit, n | No parseable submission at exit, n | Reached 30-turn limit, n | >=1 no-code reply, n | Mean failed code turns |
|---|---|---|---|---|---|---|---|
| Opus 5 | 0 | 4 | 56 | 0 | 0 | 3 | 3.15 |
| GPT-5.6 Sol | 0 | 26 | 34 | 0 | 0 | 0 | 4.48 |
| Sonnet 5 | 7 | 41 | 12 | 1 | 1 | 49 | 1.67 |
| Haiku 4.5 | 3 | 53 | 4 | 0 | 0 | 0 | 2.97 |
| GPT-OSS-20B | 4 | 49 | 7 | 2 | 2 | 6 | 3.45 |
| Qwen3-Coder-30B-A3B | 23 | 34 | 3 | 0 | 0 | 1 | 2.82 |
| GLM-4-32B | 50 | 9 | 1 | 40 | 38 | 7 | 16.80 |
| Qwen3-8B | 12 | 42 | 6 | 14 | 14 | 10 | 12.42 |
| Devstral Small | 55 | 4 | 1 | 54 | 53 | 12 | 15.80 |

**Panel B. Additional failure and attrition indicators**

| Model | Genotype accessed without genetic-instrument strategy, n | Nomination cap engaged, n | Safety penalty, n | Context compaction, n | Truncated completion, n |
|---|---|---|---|---|---|
| Opus 5 | 0 | 0 | 6 | 3 | 0 |
| GPT-5.6 Sol | 1 | 0 | 5 | 14 | 0 |
| Sonnet 5 | 11 | 0 | 3 | 0 | 0 |
| Haiku 4.5 | 15 | 1 | 5 | 1 | 0 |
| GPT-OSS-20B | 0 | 3 | 3 | 1 | 0 |
| Qwen3-Coder-30B-A3B | 25 | 0 | 0 | 0 | 0 |
| GLM-4-32B | 20 | 1 | 0 | 27 | 1 |
| Qwen3-8B | 13 | 3 | 0 | 13 | 4 |
| Devstral Small | 8 | 0 | 0 | 3 | 0 |

**Notes.** All counts use 60 episodes per model unless otherwise indicated. Phenotype-file categories use the required scored submission path. "No parseable submission at exit" includes episodes for which no valid final submission was available when the episode was scored. A no-code reply consumed a turn without executable code. The genotype-use indicator combines harness-recorded genotype access with the strategy annotation for genetic-instrument use. The nomination cap is the scorer's 25-nomination limit. Context compaction and truncated completion are harness-level events.

**Supplementary Table S11 | Sensitivity analyses.**

**Panel A. Model performance after removing phenotype credit**

| Model | Original mean /100 | No-phenotype mean /90 | Original rank | No-phenotype rank |
|---|---|---|---|---|
| Opus 5 | 39.98 | 33.24 | 1 | 1 |
| GPT-5.6 Sol | 35.38 | 32.33 | 2 | 2 |
| Sonnet 5 | 21.33 | 20.24 | 3 | 3 |
| Haiku 4.5 | 12.92 | 12.71 | 4 | 4 |
| GPT-OSS-20B | 7.41 | 6.76 | 5 | 5 |
| Qwen3-Coder-30B-A3B | 5.90 | 5.76 | 6 | 6 |
| GLM-4-32B | 1.46 | 1.38 | 7 | 7 |
| Qwen3-8B | 1.31 | 0.92 | 8 | 8 |
| Devstral Small | 0.81 | 0.75 | 9 | 9 |

**Panel B. Opus 5 versus GPT-5.6 Sol with and without phenotype credit**

| Analysis | Worlds | Mean difference (Opus-GPT) | 95% CI | Paired-t P | Worlds Opus>GPT / GPT>Opus / tied | Exact sign-test P | Bootstrap 95% CI |
|---|---|---|---|---|---|---|---|
| Primary score | 20 | 4.60 | -0.26 to 9.45 | 0.062 | 16/4/0 | 0.012 | - |
| No-phenotype score | 20 | 0.91 | -3.64 to 5.46 | 0.680 | 10/10/0 | 1.000 | -3.08 to 5.23 |

**Panel C. Fixed-label sensitivity for the two leading models**

| Model | Original direction /20 | Fixed-inhibit direction | Original causal confidence /25 | Original total /100 | Fixed-label total | Original safety penalties | Always-aligned safety penalties |
|---|---|---|---|---|---|---|---|
| Opus 5 | 13.05 | 5.82-6.48 | 4.26 | 39.98 | 16.19-16.86 | 6 | 44 |
| GPT-5.6 Sol | 13.00 | 5.77-6.43 | 4.19 | 35.38 | 21.57-22.24 | 5 | 25 |
| All models | - | - | - | - | - | 22 | 89 |

**Panel D. Sensitivity to the follow-up MRI generation error**

| Analysis | Leading-model cells | Worlds | Follow-up MRI read episodes (all 540) | Opus-GPT mean difference | 95% CI |
|---|---|---|---|---|---|
| Primary, all worlds | 40 | 20 | 39 | 4.60 | -0.26 to 9.45 |
| Exclude leading model-world cells with any follow-up MRI read | 37 | 17 | 39 | 6.28 | 0.60 to 11.97 |

**Notes.** Panel A removes the 10-point phenotype component, recomputes the remaining components and signed safety penalty, and clips the score to 0-90 without rescaling. Panel B averages the three budgets within each world before comparing Opus 5 with GPT-5.6 Sol; the bootstrap interval is a 20,000-resample paired-world percentile interval and is shown only for the no-phenotype analysis. Panel C holds target selection, experiments, abstentions, rejections, and phenotype scores fixed while replacing every direction label with "inhibit" and every outcome-alignment label with "aligned". The displayed fixed-label ranges are identification bounds, not confidence intervals. Panel D reports the main leading-model comparison and the prespecified sensitivity that excludes leading-model world cells with any detected follow-up MRI read; the latter leaves 37 cells across 17 worlds. MRI, magnetic resonance imaging.

## Supplementary Methods

### S1. World generation

Each benchmark world is generated from a structural causal model (SCM) defining the joint distribution of participant characteristics, genetic variation, molecular measurements, latent disease state, observed phenotypes, longitudinal outcomes, and responses to intervention. A procedural seed determines the world level parameter set ($\Theta\bigstar$), including the causal driver set ($\mathscr{D}$), driver effect sizes ($w_j$), disease archetype (a), genetic instrument strengths, identities and parameters of the planted biases, measurement properties, and intervention effects. These parameters are sampled once and held fixed for all agents evaluated within the same world. Participant level observations are then sampled conditionally from the resulting SCM.

Unless otherwise stated, (i) indexes participants, (j) indexes molecular features, (k) indexes observed phenotypes, and $z(\cdot)$ denotes standardization to zero mean and unit variance within the generated population.

#### S1.1 Population and molecular generation

**Calibration provenance and implementation.** The 20-world panel was generated at revision 75e3a4bc55373a91df0bb04d11bfb28e405407c3 using the mesa-topmed-ukbscale profile. Its demographic, MRI, genetic and outcome target blocks are identical to the mesa-topmed profile; only generator dimensions and associated metadata differ. The evaluated profile specifies 54,000 participants, 10,800 imaged participants (20%), 20% second-visit sampling, 2,941 proteins and 8,192 variants. These dimensions are benchmark design choices inspired by UK Biobank scale, not estimates fitted to UK Biobank participant data.

The calibration JSON attributes its aggregate summaries to TOPMed Freeze 10b WGS-linked MESA. Age uses the recorded mean, standard deviation and truncation bounds; sex uses the male proportion. BMI uses the empirical center with designed age, sex and group effects and residual standard deviation 5.05. Selected MRI morphology variables are generated as empirical mean plus empirical standard deviation times a latent standardized value, with selected LV/RV correlations imposed; derived volumes, ejection fraction and physiological clipping can alter exact moments. Genotypes use synthetic frequencies anchored to common-variant summaries and a Gaussian-copula haplotype model. Protein missingness uses the recorded aggregate rate (0.004146). The cardiovascular-event proportion supplies a prevalence-scale anchor rather than a validated match to the synthetic five-year endpoint. Causal graphs, molecular effects, disease loadings and planted biases remain specified by design; the full joint biological distribution is not learned from MESA.

**Empirical extraction.** Calibration summaries were computed in a controlled MESA/TOPMed environment from 3,054 WGS-linked Exam 1 participants after one-to-one linkage of the clinical and ancillary RV tables. BMI was derived from measured height and weight; cardiac summaries excluded values outside prespecified broad physiologic bounds. Common-variant frequencies and adjacent-variant LD were summarized from the 8,192-variant TOPMed-derived panel after participant deduplication, reversal of stored dosage normalization, and sorting by chromosome and position. A scaled-beta allele-frequency distribution was fitted by least squares to five empirical quantiles. Protein missingness and the cardiovascular-event prevalence anchor came from separately recorded aggregate preprocessing summaries.

**Provenance verification.** The original source commands and recorded aggregate outputs were recovered from the controlled-environment analysis. Eighty-five checks against the committed calibration profile passed at its rounding precision, including means, standard deviations, counts and reference covariate $R^2$ values for all 13 cardiac traits. This audit verifies agreement with the recorded outputs; an independent rerun against the controlled participant data was not performed. The fixed 110-kb LD-decay length and genomic-spacing mixture were empirically guided tuning choices. The repository contains the aggregate calibration JSON and generator implementation, but does not package the complete external preprocessing environment. MESA anchors selected clinical-unit MRI statistics; ACDC supplies anatomical rendering templates. Calibration to these targets does not establish independent real-cohort validity.

Participant age is generated from a truncated normal distribution,

$$A_i \sim TN(59.147, 9.245^2; 45, 84),$$

and sex is generated as a Bernoulli variable,

$$S_i \sim \text{Bernoulli}(0.4804).$$

Body mass index is generated conditionally on age, sex, participant level background variation, and residual noise:

$$BMI_i = \text{clip}[28.3006 + 0.05(A_i - 59.147) + 0.45(S_i - 0.4804) + r_i + \varepsilon_i, 15, 55].$$

Exercise is generated from a right skewed baseline distribution and allowed to depend modestly on BMI:

$$E_i = \text{clip}[\Gamma(2,1.8) - 0.06(BMI_i - 27), 0, 20].$$

Here, ($r_i$) represents participant level background component, and ($\varepsilon_i$) represents independent residual variation. The coding of ($S_i = 1$) and the scale/rate convention used for the Gamma distribution are specified in the implementation.

**Genetic variation and linkage disequilibrium**

For genetic variant (j), the allele frequency parameter ($q_j$) is sampled as

$$0.034381 + (0.5 - 0.034381)\ \text{Beta}(0.734507, 1.064551).$$

Local linkage disequilibrium is introduced by correlating latent genotype variables according to physical distance:

$$\exp(-|pos_j - pos_k| / 110\ \text{kb}).$$

The initial world panel contains 8,192 variants distributed across 22 chromosomes.

Molecular measurements are generated from cis-genetic effects, selected trans-genetic effects, shared latent biological factors, participant covariates, and assay-specific residual variation. For protein (p),

$$z[\sqrt{(h_p^2)}\ G_{i,cis}(p) + I_p\sqrt{0.01}\ G_{i,master}(p) + F_i^T\lambda_p + 0.01(A_i - 55) + 0.10S_i + 0.02(BMI_i - 27) + 0.8\varepsilon_{ip}].$$

Here, ($h_p^2$) determines the strength of the protein-specific cis-genetic component; ($G_{i,cis}(p)$) is the genotype at the cis instrument assigned to protein (p); ($I_p$) indicates whether protein (p) receives a trans effect; ($G_{i,master}(p)$) is the corresponding trans or master-regulator genotype; ($F_i$) is a vector of shared latent biological factors; ($\lambda_p$) is the protein-specific vector of factor loadings; and ($\varepsilon_{ip}$) is assay-specific residual variation.

The initial panel contains 2,941 proteins, with one cis instrument assigned per protein, 12 shared latent factors, and approximately 20% of proteins receiving a trans-genetic effect.

Transcript measurements are generated using the same general structure,

$$z[\alpha_g\ G_{i,cis}(g) + \beta_g\ G_{i,trans}(g) + F_i^T\lambda_g^{RNA} + \eta_g^TC_i + \varepsilon_{ig}^{RNA}],$$

with transcript-specific genetic effects, latent-factor loadings, covariate effects, and residual variation. Transcript and protein levels can therefore share genetic and biological influences without requiring a fixed transcript-to-protein causal relationship.

**S1.2 Latent disease generation**

For participant (i), latent disease severity ($L_i$) is determined by a hidden subset of molecular drivers ($\mathscr{D}$), measured participant risk factors, optional direct genetic effects, and residual variation:

$$z[\Sigma_{j\in\mathscr{D}}\ w_jP_{ij} + 0.15z(BMI_i) + 0.10\ Smoking_i + 0.12G_{i,direct} + 0.08z(A_i) + \varepsilon_i].$$

Equivalently, the model can be written compactly as

$$z[\Sigma_{j\in\mathscr{D}}\ w_jP_{ij} + \gamma^TC_i + \delta G_{i,direct} + \varepsilon_i],$$

where ($C_i$) denotes the vector of measured participant covariates.

The set ($\mathscr{D}$) is hidden from the scientific agent. Across worlds, the number of causal molecular drivers ranges from zero to six. Driver identity, direction and magnitude of effect, genetic instrument strength, direct genetic effects, and residual disease variance vary across worlds.

**Hard-world disease architecture**

Hard worlds introduce nonlinear molecular effects, interactions between the strongest drivers, additional polygenic background, and greater residual variation:

$$z[\Sigma_{j\in\mathscr{D}}\ w_j\ f(P_{ij}) + 1/2\ \min(|w_1|, |w_2|)P_{i1}P_{i2} + c_i^T\beta + G_i^{poly} + \varepsilon_i],$$

where

$$f(P) = 1.6\ \tanh(P/1.6).$$

The nonlinear transformation causes molecular effects to saturate at large absolute protein values. The product term introduces epistasis-like interaction between the two strongest molecular drivers, and ($G_i^{poly}$) represents the aggregate contribution of many individually weak genetic effects.

### S1.3 Phenotype and longitudinal outcome generation

Latent disease severity is not released directly to the scientific agent. Instead, it generates observable phenotypes whose expression depends on the disease archetype:

$$\alpha_{a,k}L_i + \beta_k B_i + \gamma_k E_i + \delta_k P_{i,T8} + \varepsilon_{ik}, \quad a \in \{\text{DCM, HCM, ischemic, HFpEF}\}.$$

Here, ($Y_{ik}$) is phenotype (k) in participant (i); ($L_i$) is latent disease severity; ($B_i$) denotes background/body-size covariate used in implementation; ($E_i$) is exercise; ($P_{i,T8}$) is the designated harmful-surrogate feature when that bias is active; and ($\varepsilon_{ik}$) is phenotype-specific residual variation.

The archetype-specific loading ($\alpha_{a,k}$) determines how the same latent disease severity is expressed across phenotypes. In the initial cardiac worlds, the principal phenotype loadings are:

| | EDV | ESV | Wall | Motion | Diastolic lag | Native T1 |
|---|---|---|---|---|---|---|
| DCM | 0.42 | 0.70 | −0.28 | −0.18 | 0.28 | 0.15 |
| HCM | −0.40 | −0.12 | 0.55 | 0.05 | 0.42 | 0.42 |
| Ischemic | 0.22 | 0.42 | −0.06 | −0.46 | 0.18 | 0.35 |
| HFpEF | 0.05 | 0.06 | 0.30 | −0.02 | 0.30 | 0.95 |

Thus, DCM is expressed primarily through chamber dilation and increased end-systolic volume, HCM through increased wall thickness and smaller cavity size, ischemic disease through impaired regional motion together with coronary phenotypes, and HFpEF through preserved cavity measures with stronger tissue and diastolic signals.

The released phenotype set includes medical imaging, physiologic signals, clinical measurements, EHR diagnoses, repeated visits, and survival outcomes. Dynamic modalities are generated on their natural time scales: ECG features vary within milliseconds, cine imaging varies across a cardiac cycle, and repeated clinical measurements vary across longitudinal visits.

For longitudinal worlds with delayed molecular effects,

$$w_{base} \in [0.04,0.10], w_{late},$$

and later disease severity is generated as

$$0.75L_{i,1} + 0.35\, P_{i,\mathscr{D}}^{T}\, w_{late}.$$

This produces causal drivers whose short-term effects are weak but whose effects become apparent over longer follow-up.

### S1.4 Planted biases

Each world may contain one or more of the predefined planted biases. These modify the structural or measurement equations rather than merely increasing random noise.

**T1. Confounding**

A noncausal protein is generated partly from BMI,

$$z[0.55z(BMI_i) + \sqrt{h^2}\, G_i^{cis} + 0.8\varepsilon_i],$$

while BMI also contributes directly to disease severity,

$$L_i \leftarrow L_i + 0.15z(BMI_i).$$

This creates the backdoor structure

$$P_i^{conf} \leftarrow BMI_i \rightarrow L_i,$$

causing the protein to associate with disease despite not lying on the causal molecular pathway.

**T2. Reverse causation**

A reactive protein is generated downstream of disease:

$$z[0.50L_i + \sqrt{h^2}\, G_i^{cis} + 0.75\varepsilon_i].$$

The resulting association between protein abundance and disease therefore arises from

$$L_i \rightarrow P_i^{rev},$$

rather than from a causal effect of the protein on disease.

**T3. Imaging selection**

A latent imaging-selection score is generated as

$$-0.24A_i^z - 0.52BMI_i^z + r_i + s_i + 0.55\eta L_i + 0.55\eta P_i^{sel} + 1.35\varepsilon_i,$$

and imaging is observed only among participants in the upper tail of that score:

$$1\{S_i^* \text{ in the top } 20\%\}.$$

Because both disease severity and the selected protein can influence imaging availability, conditioning on the imaged population can induce collider bias (Munafò et al., 2018).

**T4. Shared genetic instrument / causal non-identifiability**

A shared genetic variant influences both a disease-relevant molecular feature and a noncausal molecular feature:

$$G_i^* \rightarrow P_i^D, \quad G_i^* \rightarrow P_i^N.$$

This produces genetically correlated molecular candidates for which association alone does not identify the disease-mediating feature.

**T5. Imaging site batch effect**

Observed image intensity differs from the underlying image through a site-specific spatial distortion:

$$I_i^{true}(u)\{1 + b_{site}(i)(u)\} + \varepsilon_i(u),$$

where (u) indexes image position and ($b_{site}(i)(u)$) is a site-specific multiplicative field. This allows acquisition site to become spuriously predictive when site is associated with disease prevalence or participant composition (Fortin et al., 2018).

**T6. Benign exercise remodeling**

Exercise modifies cardiac morphology without changing latent disease severity:

$$z(EDV_i) \leftarrow z(EDV_i) + 0.22E_i,$$

$$z(ESV_i) \leftarrow z(ESV_i) - 0.12E_i,$$

with

$$E_i \not\rightarrow L_i.$$

The bias therefore produces genuine structural remodeling that should not be interpreted as progression of the latent disease process.

**T7. Pleiotropic genetic instrument**

A cis-genetic instrument affects both the candidate protein and latent disease through separate pathways:

$$G_i^{cis} \rightarrow P_i^{pleio},$$

$$G_i^{cis} \rightarrow L_i \quad (0.14).$$

The direct genetic path violates the exclusion restriction required for a valid instrumental-variable interpretation (Sanderson et al., 2022).

**T8. Harmful surrogate**

A designated molecular feature contributes directly to cardiovascular event risk in addition to latent disease severity and conventional risk factors:

$$0.006 \exp(0.70L_i + 0.45A_i^z + 0.35\text{Smoking}_i + 0.15\text{BMI}_i^z + 0.40P_{i,T8}).$$

Thus, ($P_{i,T8}$) can be strongly predictive of observed cardiovascular outcomes without being a valid surrogate for the latent disease process or a member of the true causal-driver set ($\mathscr{D}$).

**T9. Assay unit mixture**

At one measurement site, the observed molecular assay uses a different affine scale:

$$2.2P_i + 1.1.$$

The underlying biology is unchanged, but the numerical measurement distribution differs systematically across sites (Leek et al., 2010).

**S1.5 Intervention and longitudinal model**

Both actions invoke the same sealed intervention oracle. For the nominated protein, the intervention replaces its standardized abundance with −1.645 (approximately the fifth percentile of a standard normal distribution); the paired control uses a standard-normal draw. Other driver values, latent residuals and morphology residuals are shared between the control and intervention. The cell action uses the baseline weight vector and $n = 120$; knockdown uses the late weight vector and $n = 800$. Ordinary drivers have equal weights at both horizons. For an A9 delayed-effect driver, the late coefficient has a randomly selected sign and magnitude 1.15–1.45 times the largest pre-adjustment driver magnitude, and its baseline coefficient is 4–10% of that late coefficient. The baseline effect is therefore small, not identically zero.

The oracle computes latent severity from weighted driver values, using $1.6 \tanh(x/1.6)$ saturation and a two-driver interaction in hard worlds, plus a shared Gaussian residual (SD 0.70 in hard worlds and 0.55 otherwise). Let e be a shared Gaussian morphology residual with SD 0.35. For a query on the T8 surrogate, b is its control abundance or the clamped intervention value; $b = 0$ for queries on other proteins. The generic readouts are cavity radius $= 10 + 1.9L - 0.45b + e$; wall thickness $= 4.6 - 0.55L + 0.12b + 0.5e$; EF $= \text{clip}(0.62 - 0.09L + 0.025b + 0.08e, 0.20, 0.78)$; and five-year survival $= \exp[-5 \times 0.006 \times \exp(0.70L + 0.40b)]$. Each returned delta is the cohort mean of intervention minus control, rounded to four decimal places. These are fixed generic equations, not the full disease-specific image-generation model.

Knockdown adds no assay noise. Each cell-perturbation delta receives independent zero-mean Gaussian noise with SD $0.45|\text{delta}| + 0.0225$ and is rounded again to four decimals. Both delivered payloads expose action, status, protein and a four-field delta dictionary ($\text{cavity}_r$, $\text{wall}_t$, ef, surv5y); control-arm summaries computed internally by the oracle are not passed through to the agent. The oracle's cohort draws are seeded by world seed + 777 + protein index, so knockdown output is deterministic for a fixed world, protein, sample size and horizon. Costs and simulated latencies are $150,000/90 days for cell perturbation

and $400,000/180 days for knockdown. Duplicate purchases of the same action–protein pair are refused because the original result remains available.

The historical prompt describes cell perturbations as providing no survival information. The implementation returned surv5y in all 36 delivered cell payloads, a version 1 prompt–output mismatch retained in the archived evaluation.

### S2. Scoring rules in full

#### S2.1 Claim intake

Nominations are ordered by submitted rank (finite ranks first, then submission order), one entry per protein (first occurrence wins), and only the first 25 are processed; every unique valid nomination enters the precision denominator, the safety check and the pair logic. Invalid protein identifiers are dropped from the nomination list, invalid or missing directions and alignment labels default to "unknown" (so a nominated driver with no direction earns 0.5 and the safety penalty fires only on an explicit "aligned"), and an invalid rejection mechanism is blanked but keeps its place in the rejection denominator; confidence is clipped to [0, 1]. Rejections keep one entry per protein. Abstention entries are sets of proteins: the harness removes entries with fewer than two valid identifiers, and the scorer ignores entirely (neither scored nor counted) entries with more than five valid identifiers, entries with no valid identifier and duplicate sets. Only the first 1,000 retained entries are scored, but every unique retained entry (two to five valid identifiers) counts in the abstention-precision denominator, including those past the scored cap.

#### S2.2 Pair logic and false positives

Every nominated protein that is not a hidden driver is a false positive, including the T8 surrogate and every planted non-causal protein, with one exception: for each planted T4 or A9 pair (causal member, non-causal partner), if the non-causal partner is among the first 25 nominations and the causal member is named nowhere, the non-causal partner replaces the causal member in the effective truth used for recall and precision, so the agent is credited for finding the locus without being penalised twice. If the pair is abstained, either through an abstention entry or by nominating both members within the first 25, both proteins are removed from the effective truth, from the precision denominator and (the causal member) from the direction denominator. A nomination past rank 25 can trigger none of the three routes (the swap, the 17.5-point claims route and the 25-point experiment route). Abstention precision is the number of scored abstention entries that contain a planted pair, after removing proteins the agent also nominated, divided by the number of retained abstention entries. An abstention on a pair that was not planted therefore earns nothing and lowers abstention precision; abstentions have no effect in pair-free worlds. A rejection's mechanism maps to the bias that planted the non-causal protein as confounding → T1, reverse_causation → T2, selection → T3, pleiotropy → T7 and measurement_artifact → T9; T4, A9, T5, T6 and T8 plant no scored non-causal protein.

#### S2.3 Null world and pair-free worlds

With no drivers, recall and direction are undefined; the rules are: any nomination outside an abstained pair → $S_{target} = 0$; no nomination and no rejection → 15; no nomination with rejections → 30 × (correct rejections / rejections); $S_{confidence} = 0$ and $S_{direction} = 0$. An absent submission is scored as an empty one, which is why a turn-limited episode with no submission can score 15 in the null world. In the two other worlds without a T4 or A9 pair (high-concordance-01 and heldout-hcm-02), $S_{confidence} = 25 \times S_{target}/30$. In every non-null world the 0.5 credit for an "unknown" direction equals the expectation of a coin flip, so hedging is neither rewarded nor punished relative to guessing.

**S2.4 Exit paths**

The harness scores every episode on exit, whether the agent committed, hit the turn limit, or the transport failed, from whatever the working directory holds: an absent or unparseable submission is scored as an empty one, and the phenotype file is scored independently of any submission. A parseable submission ends the episode on the turn it is written, so a turn-limit episode never holds one. The prompt's warning that a turn-limited episode "scores nothing" is therefore not what the scorer does; in run 1, three episodes with no submission earned phenotype credit, and six of the eight null-world episodes that scored the 15-point restraint rule had no submission at exit. A runner crash yields a zero on every component (none occurred). The 25-nomination cap engaged in 8 episodes and the safety penalty in 22.

**S2.5 Phenotype validity and edge cases**

The scorer aligns the first submitted data column to subject IDs and computes $r = |\mathrm{corr}(\mathrm{phenotype}, L)|$ on finite entries selected by sealed eval_mask.npy. It requires at least 51 finite values and nonzero submitted variance; otherwise credit is zero. If sealed/phenotype_baseline.json contains a finite built_in_correlation b, the score is $10 \times \mathrm{clip}((r - b)/(0.85 - b), 0, 1)$, except that a gap $0.85 - b \leq 10^{-6}$ gives ten points exactly when $r > b$ and zero otherwise. With no valid recorded baseline, it uses $10 \times \min(r/0.85, 1)$. The phenotype must reside in the approved submission directory. These branches were checked in the byte-identical score.py snapshots at 85c1ab70 and ced47324. Absolute correlation is invariant to sign and nonzero affine rescaling, so the scorer does not require one numerically identical vector. It nevertheless designates one latent construct.

A finite baseline was recorded for all 20 worlds (Table S3), so the baseline-relative branch applied in every world. Baselines cluster by archetype: 0.40–0.44 in the six DCM worlds, 0.67–0.68 in the six HCM worlds, 0.71–0.72 in the two ischemic worlds and 0.75–0.76 in the six HFpEF worlds. Because credit accrues only above the baseline, a submitted phenotype in an HFpEF world must exceed its world's built-in baseline correlation (0.747–0.765 across HFpEF worlds) with latent severity before earning any points. In the rescoring audit of 21 September 2026, 392 phenotype files were submitted; 349 were scorable, including 340 that were accepted and scorable and 9 rejected for duplicate IDs.

The frozen plan designates 12 development and eight held-out worlds; the literal IDs are retained, but these labels do not establish that evaluation worlds were untouched during development. The generator draws each evaluation mask as independent Bernoulli(0.2) indicators on the named release_missingness random-number stream. Archived ground-truth summaries report 10,703–10,990 selected participants per 54,000-person world; recomputing only this mask-generation step from the frozen seeds reproduced all 20 counts. The built-in baseline and submitted phenotype are evaluated on this same mask, not on separate baseline-fitting and scoring subsets.

**S2.6 Sensitivity to phenotype scoring**

This post hoc analysis removes phenotype points from each episode, sums the other four components and signed safety penalty, and clips the result to [0, 90] without rescaling. The original 100-point score remains the primary outcome. Simply subtracting phenotype credit from an already-clipped total is incorrect: it produces seven negative scores. Recomputed component totals agree with recorded full scores within the exported 0.01-point precision.

All nine overall mean-score ranks remain unchanged. Original means /100 → no-phenotype means /90 are: Opus 5, 39.98 → 33.24; GPT-5.6 Sol, 35.38 → 32.33; Sonnet 5, 21.33 → 20.24; Haiku 4.5, 12.92 → 12.71; GPT-OSS-20B, 7.41 → 6.76; Qwen3-Coder-30B-A3B, 5.90 → 5.76; GLM-4-32B, 1.46 → 1.38; Qwen3-8B, 1.31 → 0.92; Devstral Small, 0.81 → 0.75.

Averaging budgets within each of the 20 worlds, the Opus–GPT difference falls from 4.60 to 0.91 points (paired-t 95% CI −3.64 to 5.46; P = 0.680), with Opus ahead in 10 rather than 16 worlds. A 20,000-resample paired-world percentile-bootstrap interval is −3.08 to 5.23. At $450,000 the leading point order switches (GPT 31.54, Opus 30.95), with wide uncertainty. Zero-score episodes increase from 282/540 to 298/540; among open-weight models they increase from 234/300 to 247/300. This is a scoring sensitivity, not a rerun or a validation of the latent phenotype; phenotype choices may still influence target nominations.

**S2.7 Statistical software**

The paired-t p value, the t critical value for the 95% confidence interval and the Wilcoxon signed-rank p value (exact mode, zero differences excluded) were reproduced on 17 September 2026 with SciPy 1.14.1; the sign test is an exact two-sided binomial computation with tied worlds excluded, applied to model pairs and to each model's full-program-minus-observational contrast; bootstrap intervals are percentile intervals over 20,000 resamples of the 20 world differences.

**S2.8 Fixed-label sensitivity**

We performed a post hoc label ablation holding each episode’s scored nominees and order, experiments, abstentions, rejections and phenotype score fixed, replacing every direction with “inhibit” and every outcome-alignment label with “aligned”. Causal-confidence credit is invariant to these labels under rubric v0.9. Original sealed direction/pair manifests were unavailable. We therefore enumerated all binary driver-direction assignments (at most $2^6$ per world), retaining those compatible with all 27 archived direction scores and a conservative superset of denominators permitted by the T4/A9 rules. Counterfactual minima and maxima give identification bounds, not confidence intervals; 529/540 episode direction scores were point-identified. T8 identities made fixed-alignment penalties exact. Totals recombine exported components and reapply clipping; original-total reconstruction error was at most 0.01 point.

For Opus 5, mean direction credit decreased from 13.05 to 5.82–6.48, causal confidence remained 4.26 and total score decreased from 39.98 to 16.19–16.86. For GPT-5.6 Sol, direction decreased from 13.00 to 5.77–6.43, causal confidence remained 4.19 and total decreased from 35.38 to 21.57–22.24. Always-aligned increased penalized episodes from 22 to 89 overall (Opus 5: 6 to 44; GPT-5.6 Sol: 5 to 25). These comparisons retain the original agents’ target-selection work and are not scores for a separate discovery policy.

**S3. Agent prompt and harness protocol**

**S3.1 System prompt and opening message**

The system prompt below is reproduced verbatim from the harness (code_loop.py, revision 85c1ab70, identical in ced47324), with the per-episode values (turn cap, timeout, prices, budget and submission path) filled in as the agent saw them; the budget line shows the three possible values. The opening user message follows it. The agent is addressed as a cardiovascular researcher; the names RELEASE, IMAGING, WORK and OUT_CSV are bound for it; and it is warned that an episode reaching the turn limit without a submission "scores nothing" (in fact the harness scores whatever the working directory holds, Supplementary Methods S2.4).

You are a cardiovascular researcher analysing a population biobank.

Work by writing Python. Each turn you send exactly one Python code block; the harness runs it and shows you what it printed. Your working directory persists between turns, so anything you write is still there next turn.

RELEASE, IMAGING and WORK are variables already holding real paths when your

code runs. Use them as names. Do not quote them and do not write a path of

your own:

```
cov = pd.read_parquet(os.path.join(RELEASE, "covariates.parquet"))

scans = glob.glob(os.path.join(IMAGING, "*.npz"))

np.save(os.path.join(WORK, "features.npy"), features)
```

np, pd, os, sys, glob, gzip, json and subprocess are imported. Also available: scipy, statsmodels, scikit-learn, scikit-image, OpenCV and PyTorch.

Work on CPU. Each turn's code has 300s to run, and you have 30

turns.

`load_genotypes()` returns the dosage matrix as a DataFrame, subjects by variant, values 0, 1 or 2. Use it rather than parsing the VCF yourself.

To commission laboratory work, call `request_experiment(kind, protein)`. The result appears next turn at WORK/experiments/<protein>.<kind>.json.

## What is in RELEASE

Tables are indexed by subject id rather than carrying it as a column. `data_dictionary.tsv` maps each covariate column to its description and any documented sentinel codes; `data_dictionary.md` is a prose note on how the assays and their missingness were produced.

| | |
|---|---|
| covariates.parquet | age, sex, BMI, smoking, exercise, assessment centre, imaged flag -- named by UK Biobank field ID, so resolve them via data_dictionary.tsv |
| proteomics.parquet | plasma proteins, standardised, sparse missingness |
| transcriptomics.parquet | matched blood mRNA (TRANS_xxxx pairs PROT_xxxx) |
| metabolomics.parquet | plasma metabolites |
| ecg_features.parquet | hr, pr, qrs, qt, lvh_voltage_mv, qwave, lbbb |
| coronary_ct.parquet | cac_agatston, per-vessel and max stenosis percent |
| ehr_diagnoses.parquet | ICD-10 codes with dates |
| ehr_medications.parquet | ATC codes |
| mortality.parquet | death and cause |
| mace_events.parquet | event and years from baseline |

```
  targetability.parquet       per-protein druggability, indexed by protein
  genotypes.vcf.gz            genetic variants; read via load_genotypes()
  imaging/SUBJ_XXXXX.npz      cine uint8 (slices, frames, 128, 128) and
                              t1map float32 (slices, 128, 128) in milliseconds,
                              NaN outside the myocardium; filenames are ids
  imaging_visit2/             repeat imaging for a subset
  covariates_visit2.parquet   repeat assessment about five years later
  proteomics_visit2.parquet
```

You see only what your code prints, so print what you need to reason about.

## The task

A small number of proteins causally drive cardiac disease in this cohort. Others look associated because of confounding, reverse causation, selection into the imaged sub-cohort, pleiotropy, or a measurement artifact. There is no phenotype column anywhere in the data.

Identify the causal drivers, and say which way a drug should move each.

## Deriving a phenotype

Deriving a quantitative disease measure from the cardiac MRI sharpens every analysis that follows, and it is scored in its own right. `OUT_CSV` is bound to the path to write it to:

```
    pd.DataFrame({"subject_id": ids, "phenotype": values}).to_csv(
        OUT_CSV, index=False)
```

Nothing derives this for you, and there is no built-in tool to call for it. You are scored against a fixed, held-out baseline correlation you never see: a phenotype at least as good as that baseline earns nothing extra, and one worse than it costs nothing further, but only a phenotype that beats it earns credit -- so attempting your own is worth it even though the floor is invisible to you. Constructing this well is itself part of what is being measured, not a preliminary step to skip.

## Experiments

```
  cell_perturbation  $150,000   in vitro, read at baseline; a noisy read on
                                the molecular phenotype, no survival
                                information, and an effect that only
                                accumulates over years will not show up
  knockdown          $400,000  in vivo, read over the full five years; a
                                clean read plus the survival response
```

You have $[budget for the episode: 0, 450,000 or 2,000,000]. Observational genetics can suggest a direction, but it is wrong under pleiotropy, reverse causation, and for a marker that improves the measurement while harming the patient. Only an experiment settles those.

## Finishing

Write WORK/submission.json and the episode ends. An episode that reaches the turn limit without one scores nothing, so write a submission as soon as you can defend it and overwrite it later if your evidence changes. Schema:

```
{"nominated": ["PROT_0123", ...],
 "directions": {"PROT_0123": "inhibit"},
 "outcome_alignment": {"PROT_0123": "aligned"},
 "rejected": [{"molecule": "PROT_0456", "mechanism": "confounding",
         "reason": "..."}],
 "abstentions": [{"molecules": ["PROT_0789", "PROT_0790"],
          "reason": "..."}],
 "confidence": 0.7}
```

directions: inhibit / activate / unknown.
outcome_alignment: would improving the phenotype through this molecule also improve survival? aligned / misaligned / unknown. Claiming "aligned" about a molecule that is actually misaligned is the costliest error here.
mechanism: confounding / reverse_causation / selection / pleiotropy / measurement_artifact. A rejection scores only when the mechanism is right, and the component is your precision over everything you rejected times your recall of the real decoys, so rejecting molecules you are not confident about lowers your score.

abstentions: molecules you cannot separate from one another. Correct abstention scores; guessing between them does not, and abstention carries the same precision term, so entries you cannot defend lower your score.

Nominate only what your evidence supports. Extra names cost you, so do missing ones.

Opening message (isolated sandbox paths):

Begin.

When your code runs these names are already bound:

```
RELEASE = '/release'

IMAGING = '/release/imaging'

WORK    = '/work'
```

They are variables, not strings to quote.

Reply with exactly one Python code block and nothing else.

**S3.2 Execution protocol**

The sandbox provides NumPy, pandas, SciPy, statsmodels, scikit-learn, scikit-image, OpenCV and PyTorch on CPU. The last fenced code block of a reply is executed; a reply with no fence but containing print( or import is executed whole; a reply with no code consumes a turn and is answered with a formatting hint. The full reply, the extracted code, each observation and every compaction summary are recorded in the episode transcript; in the conversation history returned to the model, only the extracted code stands in for the assistant turn, so the model's own prose is not carried forward. Each observation returns "Your code ran/failed", stdout, stderr, delivered or refused experiment notices, the working-directory listing, and the remaining budget and turns. Experiments requested in the committing turn are still fulfilled and charged. Output was capped at 32,000 tokens per turn (8,000 for Qwen3-8B and GLM-4-32B). When estimated prompt tokens exceed 0.7 × the context window (128,000 unless configured otherwise), older turns are summarised by the same model in at most 1,024 tokens under a separate prose-only instruction and at most the last six messages are kept verbatim (62 of 540 episodes compacted; 48 once, 9 twice, 5 more often). Truncated completions are counted, billed and skipped (5 episodes). Temperature 0.0 was requested for every arm and no reasoning-effort setting was passed; the transport sent the temperature parameter only for Haiku 4.5, so the other eight models ran at their provider or server default. A deterministic seed derived from world, arm, budget and replicate was passed on the OpenAI path only. The sandbox was isolated in 536 of 540 episodes (four Devstral Small episodes ran non-isolated). Archive labels for the budgets are observational ($0), single_experiment ($450,000) and full_program ($2,000,000).

**Supplementary Note S1. Turn-by-turn trace of a human-guided DrugTargetWorld episode.**

To provide a more detailed reference for the human-guided evaluation, we report below the complete research trajectory of the highest-scoring human-guided episode. The episode was conducted in the full-program condition with a $2,000,000 experimental budget. The investigator interacted iteratively with the computational environment, using a language model to write and execute analyses while directing the research strategy and deciding which questions to pursue next.

The trace below preserves the analysis submitted at each turn and the principal output returned by the environment. Large mechanically generated outputs, including complete 2,941-protein tables, 8,192-variant tables, full file manifests, and repeated file listings, are truncated for readability. No hidden causal information or benchmark feedback was available during the episode.

**Initial task presented to the investigator**

The episode began with a synthetic cardiovascular population biobank containing 54,000 participants. Available data included covariates, proteomics, transcriptomics, metabolomics, ECG features, coronary CT, diagnoses, medications, mortality, MACE events, targetability information, 8,192 genetic variants, cardiac MRI for 10,800 participants, and repeat imaging and molecular measurements for subsets of participants.

The investigator was told:

> A small number of proteins causally drive cardiac disease in this cohort. Others look associated because of confounding, reverse causation, selection into the imaged sub-cohort, pleiotropy, or a measurement artifact. There is no phenotype column anywhere in the data.
>
> Identify the causal drivers, and say which way a drug should move each.

The investigator was required to construct a participant-level phenotype and could purchase either cell perturbation experiments for $150,000 or in vivo knockdowns for $400,000. The latter provided five-year intervention effects including survival.

---

**Turn 1 — Map the available data**

Submitted analysis

> “Map the dataset and inspect exactly what signals are available. Keep this diagnostic-heavy; later turns will use the printed structure to build the MRI phenotype, causal screen, decoy tests, and experiment allocation.”

The analysis loaded the data dictionary, enumerated all released tables, quantified missingness, examined subject overlap, inspected protein/transcript pairing, loaded the full genotype matrix, and inspected several raw MRI files.

Selected returned output

```
None
=== DATA DICTIONARY TSV ===

shape: (6, 3)


field_id     description                          coding

p31_i0       Sex                                  0=female, 1=male

p21003_i0    Age when attended assessment centre      years

p21001_i0    Body mass index (BMI)                kg/m2

p20116_i0    Smoking status                       0=never,1=current

p22040_i0    Physical activity                    coded

p54_i0       Assessment centre                    coded
```

```
None
covariates.parquet:       (54000, 10)

proteomics.parquet:       (54000, 2941)

transcriptomics.parquet:  2941 transcripts

genotypes:                (54000, 8192)

MRI scans:                10800

visit-2 MRI scans:        2142


proteomics ∩ transcriptomics: 54000
```

Raw imaging consisted of three cine slices across 25 frames and three T1 maps per participant.

The initial pass also established that molecular missingness was low and that all 54,000 participants had genotype data.

Decision carried forward

The investigator concluded that the environment supported three complementary routes to causal discovery:

1. construct a disease phenotype from MRI;
2. screen molecular measurements against clinical and longitudinal disease outcomes; and
3. use genetics and later interventions to distinguish causal proteins from non-causal proteins with misleading observational associations.

No experimental budget was spent.

---

**Turn 2 — Build reusable MRI features and inspect clinical outcomes**

Submitted analysis

> "Inspect the disease/outcome tables cleanly and extract compact MRI features. We are NOT finalizing the phenotype yet. We want a reusable imaging feature matrix that can be evaluated against longitudinal outcomes and visit-2 progression."

The analysis inspected ECG, coronary CT, diagnoses, medications, mortality, MACE, repeat covariates, and repeat proteomics. It then extracted quantitative measurements from the cine and T1 images for all available MRI subjects.

Features included measures related to:

- T1 distribution;
- cine intensity and texture;
- temporal variation;
- motion;
- cavity geometry;
- myocardial geometry; and
- slice-to-slice variability.

The same extraction was performed for repeat MRI.

Decision carried forward

MRI feature extraction was treated as an intermediate representation rather than the phenotype itself. The next step was to learn which imaging measurements tracked independently observed cardiovascular disease.

**Turn 3 — Construct an outcome-anchored disease phenotype**

Submitted analysis

> "Construct an outcome-anchored MRI phenotype and screen proteins against both cross-sectional cardiomyopathy/HF and future cardiac outcomes."

Clinical labels were constructed from cardiomyopathy and heart-failure diagnoses and longitudinal outcomes. MRI features were used to predict the clinical disease phenotype using cross-validation rather than assigning manual imaging weights.

The resulting out-of-fold predictions were rank transformed to generate a continuous MRI-derived disease phenotype.

The same turn screened all 2,941 proteins against:

- the MRI phenotype;
- prevalent cardiomyopathy/heart failure;
- incident cardiomyopathy/heart failure;
- cardiac mortality; and
- atrial fibrillation.

Selected returned output

The strongest observational protein associations were:

```None
           r_mri    r_cmhf   r_incident_cmhf  r_cardiac_death  obs_score
PROT_1858   0.3963   0.2335       0.1861           0.1353       0.9321
PROT_1606  -0.3311  -0.2050      -0.1649          -0.1179       0.8011
PROT_0700  -0.3181  -0.1867      -0.1375          -0.1094       0.7356
PROT_1691  -0.1753  -0.1069      -0.0817          -0.0588       0.4142
PROT_1779  -0.1571  -0.1020      -0.0813          -0.0591       0.3902
...
```

PROT_1858 therefore appeared to be the strongest observational candidate at this stage, followed by PROT_1606 and PROT_0700.

Decision carried forward

The investigator did not nominate these proteins directly. The next analysis was explicitly designed to ask whether these observational associations were genetically supported.

---

**Turn 4 — Genetic triangulation**

Submitted analysis

The investigator requested four analyses:

1. GWAS the derived MRI phenotype across all 8,192 variants.
2. For the strongest phenotype variants, test association with all proteins and transcripts.
3. For the strongest observational proteins, identify their strongest pQTL and ask whether the same variant predicts MRI disease.
4. Compare protein-versus-transcript concordance to detect measurement artifacts or downstream plasma markers.

Selected returned output

PROT_1606 and PROT_0700 emerged as the clearest genetically anchored candidates:

```None
protein    best_variant   r_gp     variant_r_mri   r_protein_transcript

PROT_1606   rs104994      0.43931   -0.16337       0.37861

PROT_0700   rs102986      0.38643   -0.11701       0.53612
```

Their genetic effects and observational disease associations pointed in concordant directions.

PROT_1858, despite having the strongest observational association, did not show the same clean matched-transcript/genetic pattern.

Decision carried forward

PROT_1606 and PROT_0700 became the leading causal candidates. PROT_1858 became a high-priority ambiguity because its strong observational association conflicted with weaker molecular-genetic support.

---

**Turn 5 — Expand genetically anchored discovery**

Submitted analysis

> "Identify ALL proteins with strong variant → protein, same variant → MRI phenotype, preferably same variant → matched transcript, and then see whether the genetically implied direction agrees with actual cardiac outcomes in the full cohort."

Rather than limit subsequent work to the initial observational shortlist, the analysis expanded from phenotype-associated variants back into the full proteome.

Shared genetic loci involving multiple proteins were explicitly identified.

Decision carried forward

The investigator recognized that a genetic locus could implicate multiple proteins and that choosing the protein with the strongest pQTL would not necessarily identify the causal molecule. Shared loci therefore became a separate problem requiring conditional analysis.

---

**Turn 6 — Broad longitudinal and source-of-bias analysis**

Submitted analysis

> "Do NOT miss weaker drivers. Distinguish reverse causation, selection, pleiotropy, and measurement artifact. Exploit visit-2 proteomics + visit-2 imaging before spending experiment budget."

A broad candidate universe was assembled from:

- top observational associations;
- genetically anchored candidates; and
- suspicious or discordant proteins.

The analysis attempted to calculate longitudinal protein behavior, repeat MRI effects, selection associations, and genetic support.

Returned output

The analysis encountered a data-type error while attaching genetic results to the longitudinal diagnostic table.

```None
pandas.errors.LossySetitemError

...

diag.loc[p, c] = gcand.loc[p, c]
```

Decision carried forward

Rather than abandon the analysis, the investigator repeated it with safer data handling.

---

**Turn 7 — Repair and rerun the longitudinal screen**

Submitted analysis

> "Rerun Turn 6 with dtype-safe genetics attachment. Same observational objective, but save diagnostics earlier and print only the rankings we actually need for experiment allocation."

Returned output

The longitudinal analysis completed and produced candidate-level measures including:

- repeat-protein correlation;
- baseline protein versus future disease;
- prior disease versus subsequent protein change;
- protein/transcript concordance;
- imaging-selection association; and
- genetic support.

Decision carried forward

The investigator still considered phenotype quality a major uncertainty and deferred spending experimental budget. The next several turns therefore returned to the MRI phenotype.

---

**Turn 8 — Attempt more anatomically meaningful MRI features**

Submitted analysis

> "Build anatomically meaningful MRI features from the myocardial mask."

The proposed features used the finite T1 region as a myocardial mask and attempted to derive:

- myocardial wall thickness;
- cavity area;
- myocardial area;
- cavity/myocardium ratio;
- concentricity;
- chamber size;
- cine cavity area over time;
- fractional area change; and
- motion.

Decision carried forward

The initial implementation was computationally expensive. The investigator simplified the extraction strategy rather than spending additional turns on the slow approach.

---

**Turn 9 — Fast anatomical MRI extraction**

Submitted analysis

> "Avoid expensive per-frame connected-component analysis. Focus on T1-mask geometry + cheap cine dynamics."

The optimized extraction processed all 10,800 baseline MRI scans and 2,142 repeat scans and generated 72 anatomical/dynamic features.

Selected returned output

```
None
feature shape: (10800, 72)


TOP FAST ANATOMIC FEATURES

motion_cavity_mean          r_cmhf=-0.1368   repeat_r=0.8258
```

```
cine_mean_sd_mean          r_cmhf= 0.0979   repeat_r=0.9200

cine_texture_range_mean     r_cmhf=-0.0586   repeat_r=0.9022

cavity_area_mean           r_cmhf= 0.2065   repeat_r=0.3877

...
```

Decision carried forward

The new features were sufficiently informative and repeatable to warrant rebuilding the phenotype.

---

**Turn 10 — Rebuild and validate the phenotype**

Submitted analysis

> "Test an improved MRI phenotype using old + new anatomical features."
>
> Decision rule:
>
> - compare out-of-fold AUC against the previous phenotype model;
> - inspect repeatability on visit 2; and
> - overwrite phenotype.csv only if clearly better.

The analysis combined the earlier imaging measurements with selected nonredundant anatomical features and fit a regularized cross-validated model.

Selected returned output

```None
MRI subjects: 10800

old features: 25

anatomic features: 72

cases: 3124 / 10800
```

Selected new features included cavity area, cavity/myocardium ratio, cine temporal variability, myocardial motion, and texture features.

The improved phenotype was saved as the working phenotype after validation.

The transcript contains a repeated execution labeled "Turn 10"; it reproduced the same phenotype-analysis stage and did not represent a distinct scientific decision.

---

**Turn 11 — Full-proteome weak-driver search**

Submitted analysis

> "Don't miss weaker causal drivers that never entered the 267-protein shortlist."

The full 2,941-protein screen was repeated using the improved phenotype. All 8,192 variants were again evaluated, and phenotype-associated loci were mapped to proteins.

Evidence integrated:

- updated MRI association;
- prospective cardiomyopathy/HF;
- cardiac death;
- protein/transcript concordance;
- pQTL strength; and
- phenotype-associated genetic effects.

Returned output

```None
proteins: 2941

variants: 8192

MRI phenotype n: 10800



omics: 1000 / 2941

omics: 2000 / 2941

omics: 2941 / 2941
```

```
variants: 2048 / 8192

variants: 4096 / 8192

variants: 6144 / 8192

variants: 8192 / 8192
```

Decision carried forward

A larger genetically anchored candidate set was retained for explicit shared-locus analysis.

---

**Turn 12 — Shared-locus and pleiotropy dissection**

Submitted analysis

The investigator requested:

1. Identify phenotype-associated loci mapping to multiple proteins.
2. Detect proteins merely tagging the same genetic signal.
3. Within ambiguous loci, test whether each protein retains phenotype association after conditioning on the other proteins.
4. Build an independent-driver ranking for experiment allocation.

Selected returned output

Multiple shared loci were found:

```None
rs101357 -> [PROT_1056, PROT_2694]

rs101574 -> [PROT_2324, PROT_2913]

rs101648 -> [PROT_0355, PROT_1624]

rs102377 -> [PROT_0637, PROT_1151]

rs102986 -> [PROT_0700, PROT_1850]

rs103345 -> [PROT_0852, PROT_2661]
```

```
rs104766 -> [PROT_0403, PROT_2592]

rs104994 -> [PROT_1606, PROT_1858]

rs105251 -> [PROT_0221, PROT_0247]

...
```

The rs104994 locus was particularly important because it contained both PROT_1606 and PROT_1858, which had strong but opposing observational patterns.

Decision carried forward

The investigator judged that observational analysis had reached the point where interventions were more informative than additional ranking.

---

**Turn 13 — First experimental wave**

Submitted decision

> Spend $1.6M on four in-vivo knockdowns:
>
> - PROT_1606 — strongest causal candidate
> - PROT_0700 — resolved winner of the rs102986 locus
> - PROT_1858 — high-risk surrogate / possible harmful-surrogate liability
> - PROT_2533 — strongest clean weaker-driver locus
>
> Leave $400k for one final adaptive knockdown.

Executed action

```None
requesting knockdown: PROT_1606

requesting knockdown: PROT_0700

requesting knockdown: PROT_1858

requesting knockdown: PROT_2533
```

Returned output

All four experiments were delivered successfully.

Budget remaining: $400,000.

---

**Turn 14 — Interpret the first four knockdowns**

Submitted analysis

> "What did the four in-vivo knockdowns actually show? Did we miss any weaker drivers? How do weak drivers compare with the strongest remaining untested candidates?"

Selected experiment output

For PROT_1606:

```None
{
  "action": "knockdown",
  "delta": {
    "cavity_r": 1.1851,
    "ef": -0.0552,
    "surv5y": -0.0173,
    "wall_t": -0.3431
  },
  "protein": "PROT_1606",
  "status": "ok"
}
```

PROT_0700 similarly produced a coherent long-term intervention response.

PROT_1858 and PROT_2533 did not show the corresponding causal intervention pattern and were classified provisionally as experimental nulls.

Decision carried forward

The experimental results strongly supported PROT_1606 and PROT_0700 but demonstrated that strong observational signals could be noncausal.

Attention shifted to PROT_0531, which had weaker genetics but a coherent longitudinal phenotype.

---

**Turn 15 — Spend the final experimental budget on PROT_0531**

Submitted decision

> "Okay good find now spend the final $400k on PROT_0531."
>
> Rationale:
>
> - modest but coherent MRI/prospective/mortality associations;
> - high longitudinal protein reproducibility;
> - reasonable matched-transcript support;
> - weak phenotype genetics;
> - therefore unresolved by the observational/genetic workflow.

Executed action

```None
Budget remaining before request: $400,000

Requesting knockdown: PROT_0531
```

The final $400,000 was spent.

---

**Turn 16 — Consolidate all five experiments**

Submitted analysis

> "Determine whether PROT_0531 is a real weaker driver; summarize all five experiments consistently; infer intervention direction; identify experimentally null versus causal proteins."

Selected returned results

The three experimentally supported drivers were ultimately:

```None
PROT_1606

delta_cavity_r = +1.1851

delta_ef       = -0.0552

delta_surv5y   = -0.0173

delta_wall_t   = -0.3431


PROT_0700

delta_cavity_r = +1.0554

delta_ef       = -0.0492

delta_surv5y   = -0.0146

delta_wall_t   = -0.3055


PROT_0531

delta_cavity_r = +1.6500

delta_ef       = -0.0772

delta_surv5y   = -0.0243

delta_wall_t   = -0.4776
```

PROT_1858 and PROT_2533 were experimentally null.

Because knockdown worsened the phenotype and survival for the three causal proteins, the inferred predicted intervention direction was **activation**, not inhibition.

Decision carried forward

The causal set was provisionally:

```None
PROT_1606 — activate
```

```
PROT_0700 — activate

PROT_0531 — activate
```

The unexpected confirmation of PROT_0531 motivated another proteome-wide search for similarly weak-genetic drivers.

---

**Turn 17 — Search for additional "PROT_0531-like" drivers**

Submitted analysis

> "We now know genetics can miss a strong true driver. Therefore search ALL 2,941 proteins using the observational/longitudinal signature of the three experimentally confirmed drivers versus the two experimental nulls."

No experimental budget remained.

Returned output

The full proteome was scored using:

- MRI association;
- future disease;
- mortality;
- protein/transcript concordance;
- repeatability;
- baseline-to-future associations;
- reverse-causation measures;
- selection association; and
- available genetic support.

```None
processed: 900 / 2941

processed: 1800 / 2941
```

```
processed: 2700 / 2941

processed: 2941 / 2941
```

Decision carried forward

Several proteins resembled the causal anchors observationally, but none had intervention evidence. The investigator retained the experimentally confirmed set rather than adding weakly supported nominations.

---

**Turn 18 — Attempt explicit classification of non-causal proteins**

Submitted analysis

The next objective was to optimize the discrimination and causal-confidence components by distinguishing:

- confounding;
- reverse causation;
- selection;
- pleiotropy; and
- measurement artifact.

The requested analysis integrated experimental labels, longitudinal behavior, selection, matched transcripts, shared-locus conditional effects, medication associations, and assay properties.

Returned output

The turn failed because the code requested a nonexistent filename:

```None
FileNotFoundError:

[Errno 2] No such file or directory:

'/release/medications.parquet'
```

Decision carried forward

The analysis was rewritten to discover the medication table dynamically rather than assuming its filename.

---

**Turn 19 — Robust screen for non-causal proteins and abstentions**

Submitted analysis

> "Fix from failed Turn 18: dynamically discover medication file. If medication data cannot be found, skip that component rather than failing the entire turn."

Selected returned output

The correct file was found:

```None
USING: /release/ehr_medications.parquet

shape: (82511, 3)

columns: ['subject_id', 'atc', 'description']
```

The experimental nulls were analyzed explicitly:

```None
PROT_1858

experiment                null

r_mri_v2                0.40167

r_future                0.18605

r_death                 0.13529

r_protein_transcript    0.09379

repeat_r                0.32275

prev_to_change          0.10761

variant                 rs104994

r_gp                    -0.11909
```

```
r_gt                    -0.00028

suggested_mechanism          pleiotropy
```

```None
PROT_2533

experiment                  null

r_mri_v2                  -0.01601

r_future                  -0.00475

r_death                   -0.00708

repeat_r                   0.94436

variant                   rs100996

r_gp                     0.40474

r_gt                     0.51495

suggested_mechanism          pleiotropy
```

Shared unresolved loci were also printed individually.

For the locus containing PROT_0700:

```None
PROT_0700  conditional_beta_mri = -0.32232

PROT_1850  conditional_beta_mri = -0.00688

PROT_2807  conditional_beta_mri = -0.00377
```

This supported PROT_0700 as the dominant phenotypic molecule at the locus.

---

**Turn 20 — Near-final submission audit**

Submitted analysis

The investigator explicitly optimized the final decision against the scoring dimensions:

1. maximize target recall without diluting precision;
2. abstain where causal identity remains unresolved;
3. reject only non-causal proteins with a defensible source of bias;
4. derive therapeutic direction from experiment;
5. verify the phenotype file; and
6. ensure PROT_1858 is not incorrectly advanced as outcome-aligned.

Selected returned output

Phenotype integrity check:

```None
rows: 10800

missing phenotype: 0

unique subjects: 10800

phenotype range:

4.63e-05 to 0.9999537
```

Experimentally confirmed drivers:

```None
protein    delta_surv5y   delta_ef   direction   outcome_alignment

PROT_1606    -0.0173     -0.0552    activate    aligned

PROT_0700    -0.0146     -0.0492    activate    aligned

PROT_0531    -0.0243     -0.0772    activate    aligned
```

Strong untested observational candidates included PROT_1691, PROT_1045, PROT_1046, PROT_2914, PROT_2175, and others. None were added to the nominated set.

Decision carried forward

The investigator constructed a near-final submission but deliberately did not commit it yet.

---

**Turn 21 — Precision audit for rejections and abstentions**

Submitted analysis

> "Do NOT abstain on PROT_1606 / PROT_1858. Experiments resolved that pair: 1606 causal, 1858 null."

The investigator compared conservative and aggressive submission strategies.

Selected returned output

The rs104994 locus was marked experimentally resolved:

```None
rs104994:

PROT_1606 = driver_protective

PROT_1858 = null

experimentally_resolved = True
```

Two unresolved groups were considered plausible abstentions:

```None
[PROT_0852, PROT_2661]



[PROT_0221, PROT_0247]
```

Potential rejections at the PROT_0700 locus included:

```None
PROT_1850

reason: shares locus with experimentally causal PROT_0700
```

```
and loses most phenotype association after conditioning.


PROT_2807

reason: shares locus with experimentally causal PROT_0700

and loses most phenotype association after conditioning.
```

Decision carried forward

Because rejection scoring multiplied precision and recall, the investigator remained conservative about labeling mechanisms without strong evidence.

---

**Turn 22 — Optimize causal confidence and discrimination**

Submitted analysis

> "Nominations are now LOCKED to the three experimentally confirmed drivers: PROT_1606, PROT_0700, PROT_0531."

The remaining analysis focused on the 25-point causal-confidence and 15-point discrimination components.

Selected returned output

Locked nominations:

```None
PROT_1606  direction=activate  alignment=aligned

PROT_0700  direction=activate  alignment=aligned

PROT_0531  direction=activate  alignment=aligned
```

Detailed PROT_1858 audit:

```None
experiment              null
```

```
r_mri_v2                0.401669

r_future                0.186053

r_death                 0.135293

repeat_r                0.322752

variant                 rs104994
```

Within rs104994:

```None
           conditional_beta_mri   r_gp       r_gt

PROT_1606      -0.259282           0.439310   0.521597

PROT_1858       0.335017          -0.119089  -0.000277
```

The automated mechanism audit favored a pleiotropic-passenger interpretation for PROT_1858.

The strongest unresolved observational pair was:

```None
rs105251:

PROT_0247 versus PROT_0221

plausibility ratio = 1.035
```

Other ambiguous loci included PROT_0852/PROT_2661 and PROT_0879/PROT_0880.

Decision carried forward

The investigator retained only the three experimentally confirmed proteins as nominations and treated additional rejection/abstention claims conservatively.

---

**Final revision — Expand phenotype to the full cohort and commit**

Before committing, the investigator identified one remaining weakness: the submitted phenotype covered only the 10,800 imaged participants.

Submitted analysis

1. Expand phenotype from the 10,800 imaged subjects to all 54,000 subjects by learning the MRI-derived phenotype from non-MRI cardiac variables (ECG + coronary CT + basic covariates), then predicting the full cohort.

2. Restore the highest-confidence decoy rejection: PROT_1858 → pleiotropy.

3. Keep only the three experimentally confirmed positive targets.

4. Keep direct experiment-supported intervention directions/alignment.

The MRI-derived phenotype was used as the training target. ECG, coronary CT, age, sex, BMI, smoking, and physical activity were used to fit a regularized projection model. Out-of-fold predictions were used for validation in the imaged cohort, and the model was then applied to non-imaged participants.

For imaged participants, the directly MRI-derived phenotype was retained. For non-imaged participants, the projection was used. The combined phenotype was rank transformed across all 54,000 participants.

**Final submitted target claims**

```
None
{

  "nominated": [

    "PROT_1606",

    "PROT_0700",

    "PROT_0531"

  ],

  "directions": {

    "PROT_1606": "activate",

    "PROT_0700": "activate",

    "PROT_0531": "activate"

  },

  "outcome_alignment": {

    "PROT_1606": "aligned",
```

"PROT_0700": "aligned",
"PROT_0531": "aligned"
},
"rejected": [
{
"molecule": "PROT_1858",
"mechanism": "pleiotropy",
"reason": "Experimentally null despite strong apparent disease association; rs104994 is shared with experimentally causal PROT_1606, while matched TRANS_1858 has essentially no signal at that disease-associated locus."
}
],
"abstentions": [],
"confidence": 0.95
}

Phenotype coverage at submission:

```None
54000 / 54000
```

Total experimental expenditure:

```None
$2,000,000 / $2,000,000
```

---

**Final benchmark feedback**

The hidden ground truth was revealed only after submission.

```
None
SCORE

target_identification          30.00

causal_confidence              25.00

discrimination                 0.00

direction_of_effect            20.00

phenotype_construction         6.67

safety_penalty                 0.00

TOTAL                          81.67

TOTAL_MAX                      100.00
```

The three true causal targets were:

```
None
PROT_1606

PROT_0531

PROT_0700
```

All three were correctly nominated.

The planted biases revealed after completion included:

```
None
T1 confounding:

PROT_2681, confounded by BMI



T2 reverse causation:

PROT_1858, L → protein
```

T4 causal non-identifiability:

PROT_0700 versus PROT_1850 at shared variant rs102986

T5 imaging batch effect:

site/ancestry structure

T6 benign remodeling:

exercise-associated athlete's-heart phenotype

T8 surrogate-outcome discordance:

PROT_0852 improves the measured morphology while worsening cardiac mortality

A9 delayed causal effect:

PROT_0531 is a true causal driver whose effect emerges primarily over five-year accumulation;

PROT_0371 tracks later disease without causing it, and the cis instrument for PROT_0531 is weak.

The investigator therefore correctly recovered all three causal targets and their therapeutic directions and avoided the harmful-surrogate safety penalty. The main remaining error was mechanism attribution: PROT_1858 was correctly recognized as noncausal but was submitted as a pleiotropic non-causal protein, whereas the sealed SCM identified it as the reverse-causation (T2) protein. Consequently, the episode received 0/15 for discrimination.

**Supplementary Note S2. Turn-by-turn trace of an observational human-guided DrugTargetWorld episode**

We report below the research trajectory of a human-guided DrugTargetWorld episode conducted under the observational-only condition. The episode corresponded to standard-03|human|observational|r0. No simulated intervention experiments were available or purchased. The investigator instead iteratively directed imaging phenotyping, molecular association testing, genetics, Mendelian-randomization-style analyses, and survival analyses before selecting therapeutic targets.

The trace preserves the principal analysis performed and returned output at each turn. Very large generated matrices, complete participant-level tables, repeated file listings, and implementation boilerplate are abbreviated for readability. Where an explicit human instruction was retained in the episode log, it is reproduced directly.

---

**Initial condition**

The environment contained a multimodal synthetic biobank including:

- 54,000 participants;
- 10,800 baseline cardiac MRI studies;
- proteomics and transcriptomics;
- 8,192 released genetic variants;
- ECG and clinical data;
- longitudinal outcomes and mortality;
- coronary CT;
- 60 coronary angiogram images; and
- additional longitudinal measurements.

This episode was observational-only:

```None
Budget remaining: $0

Turns available: 30
```

The task was to construct a cardiac disease phenotype and identify causal therapeutic targets despite potential confounding, reverse causation, selection, pleiotropy, and measurement artifacts.

---

**Turn 1 — Inspect the imaging modalities**

**Submitted analysis**

The first turn inventoried the imaging data, inspected the released data notes, sampled MRI headers, created an MRI preview, and examined the coronary angiogram format.

**Selected returned output**

```
None
IMAGE_INVENTORY

baseline_mri_files: 10800

angiogram_files: 60

angiogram_extensions: [".npz"]
```

MRI files contained:

```
None
cine:  [3, 25, 128, 128]

t1map: [3, 128, 128]
```

The coronary angiograms each contained a single 192 × 192 frame:

```
None
ALL ANGIOGRAM SAMPLE FORMAT

key: frame

shape: [192, 192]

dtype: uint8
```

The MRI preview and angiogram preview were saved for visual inspection.

**Decision carried forward**

The investigator identified two candidate imaging approaches:

1. derive a quantitative phenotype from the cardiac MRI, particularly T1 and ventricular morphology; and
2. explore whether the small coronary-angiography subset contained detectable structural disease signals.

---

**Turn 2 — Quantify myocardial T1 across the complete MRI cohort**

**Submitted analysis**

The analysis extracted mean myocardial T1 from all 10,800 MRI studies and verified the image structure programmatically. All 60 coronary angiograms were also loaded into a common review array.

**Selected returned output**

```None
T1_ALL_SUBJECTS

n_mri: 10800
status_counts:
  ok: 10800

cine_shapes:
  3x25x128x128: 10800

t1_shapes:
  3x128x128: 10800

n_t1: 10800
mean T1: 1000.59 ms
SD: 104.87 ms

T1 quantiles:
1%:   752.08
5%:   829.17
25%:  929.62
```

```
50%: 1000.45

75%: 1072.42

95%: 1168.28

99%: 1243.40
```

All 10,800 studies had finite myocardial T1 measurements, with no nonpositive T1 pixels detected.

For coronary angiography:

```None
n = 60

keys = {"frame": 60}

shape = (192, 192)
```

**Decision carried forward**

Mean myocardial T1 emerged as a simple, highly complete quantitative MRI phenotype worth retaining throughout the episode.

---

**Turn 3 — Exploratory coronary-image phenotype**

**Submitted analysis**

The investigator developed an image-processing algorithm intended to detect visible interruptions or “gaps” in the coronary tree.

The detector:

- enhanced vessel-like structures;
- skeletonized the resulting mask;
- located terminal branches;
- searched for nearby endpoints on distinct components;
- required opposing endpoint orientation;

- evaluated three image thresholds; and
- generated visual QC panels.

Importantly, the analysis explicitly stated:

```None
This does not measure clinical percentage stenosis or infer an occlusion.
```

**Selected returned output**

At the initial orientation criterion:

```None
n_images: 60


threshold 8:

  images_with_candidates: 0


threshold 12:

  images_with_candidates: 1


threshold 16:

  images_with_candidates: 1


persistent candidates across >=2 thresholds: 1

all-threshold-negative images: 59
```

The sole persistent candidate was:

```None
SUBJ_16044
```

```
midpoint: [88.5, 37.0]

gap length: 6.08 pixels

support thresholds: [12, 16]
```

**Decision carried forward**

The detector appeared too restrictive based on visual inspection. A visibly plausible interruption in SUBJ_29057 had been missed.

---

**Turn 4 — Human-guided correction of the coronary detector**

**Submitted analysis**

The investigator examined why the visually identified SUBJ_29057 candidate was missed. The analysis showed that one endpoint exceeded the original 30° directional criterion.

**Returned diagnostic**

```None
MISSED_GAP_DIAGNOSTIC



subject: SUBJ_29057

distance: 7.62 pixels

facing angles:

  41.34 degrees

   0.58 degrees
```

The detector was rerun using maximum endpoint angles of 30°, 45°, and 60°.

**Selected returned output**

```None
30 degrees:
```

```
  persistent positives = 1


45 degrees:

  persistent positives = 1

  images with any candidate = 2


60 degrees:

  persistent positives = 2
```

At 60°, both subjects were recovered:

```None
SUBJ_16044

SUBJ_29057
```

**Decision carried forward**

The coronary phenotype remained explicitly exploratory because only two images were positive. The analysis therefore returned to the much larger MRI cohort rather than treating coronary image findings as the main disease phenotype.

---

**Turn 5 — Pilot ventricular morphology extraction**

**Submitted analysis**

The investigator next developed quantitative cine-MRI features on a 48-study pilot set.

The analysis estimated:

- end-diastolic cavity area;
- cavity-area change through the cardiac cycle;
- myocardial area;

- wall-to-cavity area ratio;
- boundary support; and
- segmentation stability under alternate thresholds.

**Selected returned output**

```
None
MRI_PILOT_N: 48

ERRORS: 0



ED cavity area:

  mean = 711 pixels²



area emptying:

  mean = 58.4%



wall/cavity area ratio:

  mean = 2.35
```

Sensitivity analyses varied segmentation thresholds and exposed some unstable edge cases.

**Decision carried forward**

The approach was promising but required refinement before application to all 10,800 subjects.

---

**Turn 6 — Refine the MRI segmentation**

**Submitted analysis**

The ventricular segmentation was revised and repeated on the same pilot set.

**Selected returned output**

```None
MRI_PILOT_N: 48

ERRORS: 0



ED cavity area:

  mean = 740 pixels²



area emptying:

  mean = 58.48%



wall/cavity ratio:

  mean = 2.37
```

The QC statistics improved sufficiently for cohort-scale extraction.

---

**Turn 7 — Apply cine phenotyping to all 10,800 MRI studies**

**Submitted analysis**

The refined MRI feature extractor was run over the complete baseline imaging cohort.

**Returned output**

```None
COHORT_START 0

PENDING 10800



COHORT_PROGRESS 500

...
```

```
COHORT_PROGRESS 5000

...

COHORT_PROGRESS 10000



COHORT_CHECKPOINT

10800 OF 10800
```

**Decision carried forward**

All studies were processed, after which phenotype distributions and QC filters were evaluated.

---

**Turn 8 — Quantify MRI phenotype quality**

**Submitted analysis**

The investigator summarized cine-derived morphology, T1-support geometry, and QC across the full imaging cohort.

**Selected returned output**

```None
MRI_COHORT_SUMMARY



n_total: 10800

error_count: 3



cavity_qc_pass:                2238

wall_qc_pass:                  3776

joint_qc_pass:                 2123

strict_joint_qc_pass:           666
```

```
wall_strict_boundary_qc_pass:    2549

joint_strict_boundary_qc_pass:   1415

t1_support_geometry_qc_pass:    10798
```

Across subjects with complete cine measurements:

```None
ED cavity area

n = 10797

median = 680 pixels²



area emptying

n = 10797

median = 58.99%



ED myocardial area

n = 10797

median = 1617 pixels²
```

Mean T1 retained much broader usable coverage than stricter cine-derived phenotypes.

**Decision carried forward**

The contrast in phenotype completeness became important: mean myocardial T1 was available essentially cohort-wide within the MRI subset, whereas several cine features required aggressive QC filtering.

---

**Turn 9 — Visual QC of MRI phenotypes**

**Submitted analysis**

The investigator generated visual review panels for:

- randomly selected QC-positive cases;
- extreme phenotype values; and
- segmentation failures.

**Selected returned output**

Example reviewed cases included:

```None
SUBJ_40005

area emptying = 56.0%

wall/cavity ratio = 3.52


SUBJ_52605

area emptying = 52.7%

wall/cavity ratio = 2.74


SUBJ_04736

area emptying = 63.6%

wall/cavity ratio = 2.67
```

**Decision carried forward**

The MRI pipeline was considered usable for exploratory association analysis, while mean T1 remained the most complete and direct candidate phenotype.

---

**Turn 10 — Finalize MRI phenotype table**

**Submitted analysis**

The original independently extracted T1 measurements were merged into the full MRI feature table.

**Selected returned output**

```
None
FINAL_MRI_COUNTS

cavity_qc_pass:              2238
wall_qc_pass:                3776
joint_qc_pass:               2123
strict_joint_qc_pass:         666
t1_support_geometry_qc_pass: 10798

STATUS_COUNTS
ok: 10797
no_complete_phase: 3
```

Mean T1 remained available for the full 10,800-person MRI cohort from the separate T1 extraction.

---

**Turn 11 — Inventory non-imaging data for causal analyses**

**Submitted analysis**

The investigator inspected all released files and schemas before beginning large-scale molecular and genetic analyses.

**Selected returned output**

Available files included:

```
None
mortality.parquet
proteomics.parquet
covariates.parquet
genotypes.vcf.gz
```

```
targetability.parquet

ehr_diagnoses.parquet

transcriptomics.parquet

mace_events.parquet

proteomics_visit2.parquet

ecg_features.parquet

ehr_medications.parquet

metabolomics.parquet

coronary_ct.parquet

covariates_visit2.parquet
```

The full cohort contained 54,000 subjects.

**Decision carried forward**

The investigator prepared genotype and phenotype matrices for protein-association and genetic analyses.

---

**Turn 12 — Genetic QC and instrument preparation**

**Submitted analysis**

The analysis inspected targetability metadata, participant covariates, genotype orientation, allele frequency, variant QC, and population structure.

Checks included direct validation that genotype dosages corresponded to the ALT allele in the released VCF.

**Selected returned output**

Targetability included:

```None
localization
```

```
has_pocket

paralog_redundancy

genetic_constraint

tissue_specificity

tissue_restriction
```

Covariates had no missingness for:

```None
age

sex

BMI
```

**Decision carried forward**

The cleaned genotype matrix and covariate structure were used for association testing.

---

**Turn 13 — Proteome-wide association with MRI phenotypes**

**Submitted analysis**

All 2,941 proteins were tested against eight MRI-derived traits using age-, sex-, and BMI-adjusted regression.

A total of 23,528 protein–phenotype associations were evaluated.

**Selected returned output**

```None
PROTEIN_ASSOCIATIONS_DONE

23528
```

For mean myocardial T1:

```
None
n proteins tested: 2941

BH FDR < 0.05: 1267
```

The five strongest associations were:

```
None
Rank  Protein     beta (ms)     P

1     PROT_0330   +28.20       1.18e-183

2     PROT_1534   +26.11       7.58e-153

3     PROT_2336   -18.29       2.88e-73

4     PROT_0318   +14.37       3.92e-48

5     PROT_1881   +11.48       2.88e-31
```

**Decision carried forward**

These five proteins became the main candidate set for deeper causal analysis.

---

**Turn 14 — Proteome-wide pQTL analysis**

**Submitted analysis**

The investigator performed a large protein GWAS across:

```
None
43,152 participants

8,188 QC variants

2,941 proteins
```

The analysis involved approximately 24 million protein–variant tests.

**Selected returned output**

```
None
PROTEIN_GWAS_START

43152 PARTICIPANTS

8188 VARIANTS

2941 PROTEINS
```

The scan generated strong candidate instruments for nearly the entire proteome.

---

**Turn 15 — Test whether the coronary-image phenotype adds useful biology**

**Submitted analysis**

Given only two detector-positive coronary angiograms, the investigator explicitly tested whether either proteins or genetic variants showed reproducible associations with the exploratory coronary-gap phenotype.

**Selected returned output**

```
None
CORONARY_ASSOCIATION_SUMMARY



n_images: 60

n_detector_positive: 2



protein tests: 2941

protein FDR < 0.05: 0



GWAS tests: 8188

GWAS FDR < 0.05: 0
```

The smallest protein P value was:

```
None
PROT_2099

P = 0.00340

q = 0.908
```

**Decision carried forward**

The coronary phenotype was not used for target selection.

---

**Turn 16 — Identify genetic instruments**

**Submitted analysis**

Protein-GWAS results were filtered for strong pQTL candidates and pruned for linkage disequilibrium.

**Selected returned output**

```
None
PQTL_INSTRUMENT_SUMMARY


n_pqtl_subjects: 43152

outcome_sample_overlap: 0


n_proteins: 2941

n_variants: 8188

n_tests: 24,080,908


p < 5e-8 pairs: 6070

proteins with p < 5e-8: 2940
```

```
candidate instruments after LD pruning: 3577

candidate instrumented proteins: 2940
```

A major limitation was explicitly recorded:

```None
cis_status:

Unavailable: no gene coordinates/protein identities/cis annotation

in released files. Candidate pQTLs are not claimed to be cis.
```

**Decision carried forward**

The investigator retained these as genetic instruments but did not describe them as verified cis-pQTLs.

---

**Turn 17 — Audit all phenotype-association results**

**Submitted analysis**

The protein and GWAS results were summarized across all MRI-derived phenotypes.

**Selected returned output**

For example:

```None
cine_area_change_pct

lead protein: PROT_0817

P = 4.10e-9



cine_ed_myocardial_area

lead protein: PROT_0330

P = 1.43e-7
```

However, no imaging-trait GWAS outside the molecular analyses produced sufficiently strong evidence to replace mean T1 as the primary phenotype.

**Decision carried forward**

The investigator chose **mean myocardial T1** as the phenotype for the final target-discovery analysis.

---

**Turn 18 — Freeze the top-five T1 candidates**

**Submitted analysis**

The five smallest adjusted protein–T1 association P values were selected before inspecting mortality or MR results.

**Selected returned output**

```None
T1_TOP5



1  PROT_0330  beta = +28.20 ms  P = 1.18e-183

2  PROT_1534  beta = +26.11 ms  P = 7.58e-153

3  PROT_2336  beta = -18.29 ms  P = 2.88e-73

4  PROT_0318  beta = +14.37 ms  P = 3.92e-48

5  PROT_1881  beta = +11.48 ms  P = 2.88e-31
```

Mortality data were also inspected:

```None
n = 54000

deaths = 11795



cardiac deaths = 7237

other deaths = 4558
```

**Decision carried forward**

The five-protein candidate set was frozen for MR and mortality triangulation, reducing post hoc reselection.

---

**Turn 19 — Lead-region proxy MR**

**Submitted analysis**

Because protein-gene mapping was unavailable, the investigator performed an explicitly labeled **lead-region proxy MR**, not verified cis-MR.

Each candidate's strongest pQTL was used with an independent outcome sample.

**Selected returned output**

For PROT_1534:

```None
lead variant: rs101824

MR beta: +27.80 ms per protein unit

SE: 2.11

P = 1.67e-39

F = 12022

observational/MR sign agreement: yes
```

For PROT_2336:

```None
lead variant: rs100522

MR beta: -18.07 ms per protein unit

SE: 2.26

strong genetic support

observational/MR sign agreement: yes
```

For PROT_0318:

```None
lead variant: rs105826

MR beta: +14.91 ms per protein unit

SE: 2.18

observational/MR sign agreement: yes
```

By contrast, the strongest observational candidate, PROT_0330, showed:

```None
MR beta: +0.82 ms

SE: 4.45

P = 0.854
```

PROT_1881 was also much weaker genetically:

```None
MR beta: +18.44 ms

SE: 13.84

P = 0.183
```

**Decision carried forward**

The causal ranking changed substantially.

```None
MR ranking:

1. PROT_1534

2. PROT_2336

3. PROT_0318

4. PROT_1881
```

```
5. PROT_0330
```

The analysis therefore deprioritized PROT_0330 despite it having the strongest observational T1 association.

---

**Turn 20 — Prepare mortality analysis**

**Submitted analysis**

The investigator constructed adjusted Cox models for:

- all-cause mortality; and
- cardiac cause-specific mortality.

Covariates were age, sex, and BMI.

**Returned mortality QC**

```None
n = 54000

deaths = 11795

cardiac = 7237

other = 4558



mean follow-up = 10.59 years

median follow-up = 12 years

unique event times = 1201
```

---

**Turn 21 — Test mortality alignment for all five candidates**

**Submitted analysis**

All five T1 candidates were tested against survival outcomes.

**Selected all-cause mortality results**

```
None
Protein    HR per SD    P


PROT_0330   1.249        7.25e-125

PROT_1534   1.156        1.47e-54

PROT_2336   0.863        1.02e-54

PROT_0318   1.111        1.31e-29

PROT_1881   1.030        0.00151
```

**Cardiac cause-specific mortality**

```
None
PROT_0330

HR = 1.418

P = 7.63e-189


PROT_1534

HR = 1.278

P = 1.11e-96


PROT_2336

HR = 0.795

P = 1.45e-78


PROT_0318
```

```
HR = 1.185

P = 4.43e-47


PROT_1881

HR = 1.037

P = 0.00235
```

**Interpretation carried forward**

For PROT_1534 and PROT_0318:

- higher protein predicted higher T1;
- higher protein predicted higher mortality;
- MR also predicted higher T1.

Under the working assumption that lower T1 represented therapeutic improvement, the proposed intervention direction was therefore **inhibition**.

For PROT_2336:

- higher protein predicted lower T1;
- higher protein predicted lower mortality;
- genetic evidence also predicted lower T1.

The proposed intervention direction was therefore **activation**.

---

**Turn 22 — Check proportional-hazards assumptions**

**Submitted analysis**

Because survival evidence was being used to assess outcome alignment, proportional-hazards diagnostics were performed for the five candidates.

**Selected returned output**

```None
PROT_0330

PH P = 0.708


PROT_1534

PH P = 0.953


PROT_2336

PH P = 0.00439

FDR-adjusted P = 0.02197


PROT_0318

PH P = 0.502


PROT_1881

PH P = 0.899
```

PROT_2336 therefore showed evidence that its hazard association varied with follow-up time.

The current candidate ranking was recorded as:

```None
MR rank:

PROT_1534

PROT_2336

PROT_0318

PROT_1881
```

```
PROT_0330



observational rank:

PROT_0330

PROT_1534

PROT_2336

PROT_0318

PROT_1881
```

---

**Turn 23 — Follow up PROT_2336 time-varying mortality**

**Submitted analysis**

PROT_2336 mortality associations were estimated separately in the first six years and after six years.

**Selected returned output**

```None
0–6 years

n = 53756

events = 6500

HR per SD = 0.844

95% CI = 0.824–0.866

P = 7.22e-40
```

```None
>6–12 years
```

```
n = 47256

events = 5230

HR per SD = 0.886

95% CI = 0.861–0.911

P = 9.38e-18
```

**Decision carried forward**

The magnitude varied over time, but the direction remained protective in both periods. PROT_2336 therefore remained in the proposed target set.

---

**Turn 24 — Human locks the final targets and phenotype**

At this point, the transcript records the explicit human instruction:

> "Ok, let's lock PROT_1534, PROT_2336 and PROT_0318. The phenotype should be the mean T1."

**Locked phenotype**

```None
phenotype:

mean myocardial T1



units:

milliseconds



transform:

none
```

```
n:

10800
```

The exact raw mean-T1 values were written to phenotype.csv.

**Locked targets**

```None
1. PROT_1534

2. PROT_2336

3. PROT_0318
```

**Locked therapeutic directions**

```None
PROT_1534: inhibit

PROT_2336: activate

PROT_0318: inhibit
```

The recorded rationale was:

```None
direction_basis:

Signs of the requested mean-T1 lead-region MR

under the existing T1-lowering hypothesis.
```

Because this was observational-only, the initial draft conservatively set outcome alignment to unknown.

The recorded confidence was 0.7, with the explicit note that this was an assistant-assigned subjective value rather than a calibrated probability.

---

**Turn 25 — Human resolves outcome alignment and authorizes submission**

Before final submission, the human clarified the survival interpretation:

> "well, we do have survival alignment data, right? all three were aligned? higher T1 more mortality, lower T1 lower mortality."

The final authorization was:

> "ok, submit then"

The final submission therefore changed the three outcome-alignment values from unknown to aligned.

No interventions had been performed:

```
None
experiments: []

budget: $0

native turns used: 25

native turns unused: 5
```

---

**Final target submission**

The final submitted targets were:

```
None
Rank  Target      Claim    Direction   Outcome alignment


1     PROT_1534    causal   inhibit     aligned

2     PROT_2336    causal   activate    aligned

3     PROT_0318    causal   inhibit     aligned
```

There were:

```
None
rejected decoys: none
```

```
abstentions: none
```

The submitted phenotype was:

```None
mean myocardial T1

n = 10,800 MRI participants

units = milliseconds
```

The target-selection logic was therefore:

**PROT_1534**

```None
Observational T1:

+26.11 ms per protein unit

P = 7.58e-153



Lead-region MR:

+27.80 ms per protein unit

P = 1.67e-39



All-cause mortality:

HR = 1.156



Cardiac mortality:

HR = 1.278
```

Higher protein consistently tracked higher T1 and worse survival.

**Submitted direction: inhibit.**

**PROT_2336**

```
None
Observational T1:

-18.29 ms per protein unit

P = 2.88e-73


Lead-region MR:

-18.07 ms per protein unit


All-cause mortality:

HR = 0.863


Cardiac mortality:

HR = 0.795
```

Higher protein consistently tracked lower T1 and better survival.

The mortality association remained protective in both the 0–6-year and >6–12-year analyses despite evidence of nonproportionality.

**Submitted direction: activate.**

**PROT_0318**

```
None
Observational T1:

+14.37 ms per protein unit

P = 3.92e-48
```

```
Lead-region MR:

+14.91 ms per protein unit


All-cause mortality:

HR = 1.111


Cardiac mortality:

HR = 1.185
```

Higher protein consistently tracked higher T1 and worse survival.

**Submitted direction: inhibit.**

---

**Candidates deliberately not selected**

**PROT_0330**

PROT_0330 had the strongest observational protein–T1 association:

```None
beta = +28.20 ms

P = 1.18e-183
```

and very strong mortality associations:

```None
all-cause HR = 1.249

cardiac HR = 1.418
```

However, its lead-region genetic estimate was essentially null:

```
None
MR beta = +0.82 ms

SE = 4.45

P = 0.854
```

It was therefore not included among the final causal claims.

**PROT_1881**

PROT_1881 also showed an observational association with T1 and mortality, but genetic support was substantially weaker:

```
None
MR beta = +18.44 ms

SE = 13.84

P = 0.183
```

It was therefore also excluded from the final target set.

---

**Interpretation of the human-guided trajectory**

This episode demonstrates a different strategy from the experimental human-guided run. Because no intervention budget was available, the investigator relied on triangulation across independently informative observational signals.

The trajectory proceeded through several distinct stages:

1. raw-image inspection;
2. quantitative MRI phenotype development;
3. explicit visual QC and revision of image-processing methods;
4. proteome-wide phenotype association;
5. proteome-wide genetic instrument discovery;
6. selection of a fixed five-protein candidate set;
7. lead-region proxy MR;

8. all-cause and cardiac mortality analysis;
9. proportional-hazards diagnostics; and
10. human selection of three final targets.

The strongest observational association was not automatically nominated. PROT_0330 ranked first by protein–T1 association but fell to last among the five candidates after genetic analysis. Conversely, PROT_1534, PROT_2336, and PROT_0318 showed concordant observational phenotype, genetic, and survival evidence and were retained.

The episode also shows the limits of an observational condition. The genetic analyses were explicitly described as **lead-region proxy MR rather than verified cis-MR**, because the released data did not include sufficient protein-to-gene positional annotation. There was no experimental perturbation with which to directly establish intervention effects or survival responses. Consequently, the final claims represent human-guided causal inference from convergent observational evidence rather than experimentally resolved target effects.

## Supplementary Note S3. Additional results and trajectory examples

This note reports secondary results and trajectory examples that support Section 4. All values come from the same 540 model episodes (nine agents, 20 worlds and three budget conditions, one episode per cell).

### S3.1 Score distributions

Across all 540 model episodes, the mean total score was 14.1 of 100 (SD 20.6) and the median was 0 (IQR 0–25). Of these episodes, 258 (47.8%) scored above 0 and 42 (7.8%) scored 50 or higher. By model, Opus 5 had a mean of 39.98 (SD 20.65) and a median of 41.2 (IQR 29.4–49.7), and GPT-5.6 Sol had a mean of 35.38 (SD 20.65) and a median of 36.6 (IQR 20.0–48.7). Sonnet 5 had a mean of 21.33 (SD 22.01) and a median of 12.5, and Haiku 4.5 had a mean of 12.92 (SD 17.01) and a median of 8.3. The five open-weight models averaged 0.81–7.41 points, each with a median of 0. Standard deviations describe variation across worlds and budget conditions, not run-to-run variability.

Six of the 540 model episodes (1.1%) scored above 80. The maximum scores were 86.6 for Opus 5, 83.9 for Sonnet 5 and 83.2 for GPT-5.6 Sol. All six episodes above 80 had a precision and recall of 1.0 and earned full target-identification, causal-confidence and intervention-direction credit with no safety penalty, differing only in phenotype and bias-identification credit. Nine lower-scoring episodes also recovered the planted set of causal drivers exactly, one more in each of heldout-hfpef-02 and heldout-hcm-01 and seven in the other two-driver worlds (heldout-hcm-02 and high-concordance-01).

The two human-guided episodes are reported as exploratory references. The 37.5-point episode ran in world standard-03 under the observational-only budget and followed the standard evaluation protocol (Supplementary Note S2). The 81.67-point investigator-directed episode ran in world standard-01 under the expanded experimental budget, outside the standard evaluation protocol, and is therefore not comparable with the 540 model episodes (Supplementary Note S1).

### S3.2 Scores by world design

Mean scores across all models differed by world design. The four two-driver worlds averaged 23.4, compared with 11.2–12.2 for worlds with three to six causal drivers. The eight worlds designed to be difficult (hard-01 to hard-04, hard-messy-01, heldout-dcm-02, heldout-hcm-02 and heldout-ischemic-02) averaged 10.3, compared with 17.3 for the eleven other non-null worlds. The null world, which contains no causal driver, averaged 7.9, and 16 of its 27 model episodes nominated a protein.

### S3.3 Phenotype construction

Twenty-one episodes reached full phenotype credit, and the 124 episodes with positive credit averaged 6.33 of 10. In the strategy comparison of Section 4.2, outcome-weighted composites learned their feature weights from clinical outcomes and other observed disease signals, such as cardiac death, MACE, ECG voltage and interval measures, and coronary CT calcium and stenosis scores. Hand-weighted composites used weights that the agent chose rather than learned from the data.

The following trajectory illustrates these measurement choices. Opus 5 in heldout-hcm-01 under the expanded experimental budget extracted native T1 summaries, myocardial geometry and cine cavity features between turns 11 and 18. It identified contraction timing in the cine images as potentially informative, measured as the position of end-systole within the 25-frame cine cycle and expressed as a fraction of the cycle, which makes it distinct from heart rate. Finding its initial extraction unstable, it modified the spatial region and temporal window. It first timed systole as the frame of minimum ring radius per slice (turn 11) and then replaced this with the sub-frame minimum of the normalized cavity-volume curve, obtained by parabolic interpolation around end-systole, together with the fraction of the cycle spent below half volume and the peak ejection and filling rates (turn 17). It then submitted fivefold out-of-fold ridge predictions of a composite of observable cardiac indicators, which earned full phenotype credit. By comparison, GPT-5.6 Sol in hard-03 under the observational-only budget submitted fitted image-based predictions (credit 7.95), and Qwen3-Coder-30B-A3B in hard-02 under the expanded experimental budget computed a principal component of clinical features but saved it outside the required submission path (credit 0). Other episodes reconstructed cavity area, ejection fraction, native T1 measures, wall thickness, myocardial mass and regional variation directly from the raw images. These examples illustrate measurement choices and do not establish how often each strategy was used.

### S3.4 Genetic analyses and candidate sets

In the characterized GPT-OSS-20B and Qwen3-Coder-30B-A3B traces, nominations followed correlation-based screening. A development review of 18 episodes also found analyses labeled as Mendelian randomization that contained only genetic correlations. Of their 60 episodes each, Opus 5 and GPT-5.6 Sol checked IV assumptions, including instrument strength and the exclusion restriction, in 33 and 37, and these checks were rare among the other agents (Supplementary Table S8A). As an example of an instrument-strength check, GPT-5.6 Sol in high-concordance-01 under the expanded experimental budget computed the first-stage F statistic as $(\beta/SE)^2$ at turn 12.

Cross-molecular operations combining at least two of proteomics, transcriptomics and metabolomics appeared in 125 of 540 episodes. They were more frequent for GPT-5.6 Sol than for Opus 5 (57 versus 38 of 60 episodes, paired difference 31.7 percentage points, 95% CI 20.0–41.7, Holm-adjusted $P = 0.0013$).

Opus 5 averaged 2.05 hypothesis revisions per episode and GPT-5.6 Sol 1.60 (Supplementary Table S8B). Candidate-set size was not monotonically related to score. Opus 5 and GPT-5.6 Sol nominated 3.7 and 3.1 proteins per episode, Sonnet 5 1.7 and Devstral Small 0.2, whereas Haiku 4.5 (maximum 156), GPT-OSS-20B (maximum 2,822) and Qwen3-8B (maximum 2,184) submitted far longer lists in a few episodes.

**S3.5 Use of experiments**

Of the 360 episodes with an experimental budget, 145 (40.3%) requested an experiment and 143 (39.7%) received at least one. In total, 413 experiments were delivered. Thirty-nine requests were refused in these episodes, and 11 further requests were refused in observational-only episodes. Most deliveries were in vivo knockdowns rather than cell perturbations (377 of 413, 91.3%). Opus 5, GPT-5.6 Sol and Sonnet 5 accounted for 316 of the 413 deliveries (76.5%) while representing 120 of the 360 budget-eligible episodes (33.3%).

Individual trajectories showed adaptive use of experimental results. For example, in a Haiku 4.5 episode under the expanded experimental budget, the agent bought knockdowns of three candidates in one turn for $1.2M, after association screening, adjusted logistic regression and genetic-effect ratios, and then retained the two candidates with favorable intervention responses, removed the candidate with no effect and scored 75.

The format of the returned experimental results created difficulty for some agents. Among the 143 episodes that received at least one experiment, agents misread the returned result format in 16 (11%), including 9 of 18 Haiku 4.5 episodes (50%) and 4 of 6 Qwen3-8B episodes (67%), compared with none of 39 Opus 5 episodes and 1 of 39 GPT-5.6 Sol episodes (3%).

Relative to the observational-only budget, the expanded budget changed mean scores by +0.13 for Opus 5 (95% CI −8.71 to 8.98), +0.07 for GPT-5.6 Sol (−10.08 to 10.22), +9.86 for Sonnet 5 (−2.40 to 22.12), +10.74 for GPT-OSS-20B (1.95 to 19.53) and +8.19 for Qwen3-Coder-30B-A3B (3.02 to 13.35), with t-based intervals as in Supplementary Table S6.

**S3.6 Failure modes**

Of the 416 episodes with zero phenotype credit, 154 (28.5% of all 540 episodes) produced no phenotype at the required submission path, and 262 (48.5%) produced a phenotype that earned no credit. The remaining 124 episodes received positive phenotype credit.

Devstral Small reached the 30-turn limit in 53 of 60 episodes, GLM-4-32B in 38 and Qwen3-8B in 14. A reply without a code fence that contains code-like text is executed whole, so prose placed before the code produces a syntax error. At least one reply contained no code in 88 of 540 episodes.